\pdfoutput=1

\documentclass[11pt]{article}

\usepackage{acl}

\usepackage{microtype}
\usepackage{comment}
\usepackage[most]{tcolorbox}
\usepackage{fontawesome}
\usepackage{times}
\usepackage{enumitem}
\usepackage{multirow}
\usepackage{placeins}
\usepackage{subcaption}
\setlist{nosep}
\usepackage{soul}
\usepackage{multicol}

\usepackage{dblfloatfix}
\usepackage{titlesec}
\usepackage{adjustbox}

\title{LLMs as Meta-Reviewers' Assistants: A Case Study}

\author{\begin{minipage}{0.9\textwidth}
 \centering
 \bf Eftekhar Hossain{\textsuperscript{$\clubsuit$}}, Sanjeev Kumar Sinha{\textsuperscript{$\spadesuit$}}, Naman Bansal{\textsuperscript{$\spadesuit$}}, Alex Knipper{\textsuperscript{$\clubsuit$}},\\ \bf Souvika Sarkar{\textsuperscript{$\diamondsuit$}}, John Salvador{\textsuperscript{$\clubsuit$}}, Yash Mahajan{\textsuperscript{$\spadesuit$}}, Sri Guttikonda{\textsuperscript{$\spadesuit$}},
  \\ \bf Mousumi Akter{\textsuperscript{$\spadesuit$}}, Md. Mahadi Hassan{\textsuperscript{$\clubsuit$}},  Matthew Freestone{\textsuperscript{$\spadesuit$}}, \\ \bf Matthew C. Williams Jr.{\textsuperscript{$\spadesuit$}}, Dongji Feng{\textsuperscript{\faCny}}, Santu Karmaker{\textsuperscript{$\clubsuit$}}\end{minipage}\\\\
  {\textsuperscript{$\clubsuit$}}\textit{Bridge-AI Lab}, Department of CS, University of Central Florida\\
 \textsuperscript{{$\spadesuit$}}Department of CSSE, Auburn University,
    \textsuperscript{{$\diamondsuit$}}School of Computing, Wichita State University\\{\textsuperscript{\faCny}} Department of MCS, Gustavus Adolphus College\\
  \texttt{\{eftekhar, santu\}@ucf.edu}\\
}

\begin{document}

\maketitle

\begin{abstract}
One of the most important yet onerous tasks in the academic peer-reviewing process is composing meta-reviews, which involves assimilating diverse opinions from multiple expert peers, formulating one's self-judgment as a senior expert, and then summarizing all these perspectives into a concise holistic overview to make an overall recommendation. This process is time-consuming and can be compromised by human factors like fatigue, inconsistency, missing tiny details, etc. Given the latest major developments in Large Language Models (LLMs), it is very compelling to rigorously study whether LLMs can help meta-reviewers perform this important task better. In this paper, we perform a case study with three popular LLMs, i.e., GPT-3.5, LLaMA2, and PaLM2, to assist meta-reviewers in better comprehending multiple experts' perspectives by generating a controlled multi-perspective-summary (MPS) of their opinions. To achieve this, we prompt three LLMs with different types/levels of prompts based on the recently proposed TELeR taxonomy. Finally, we perform a detailed qualitative study of the MPSs generated by the LLMs and report our findings\footnote{Our dataset is available at \href{https://github.com/BridgeAI-Lab/LLM-as-Meta-Reviewer}{https://github.com/BridgeAI-Lab/LLM-as-Meta-Reviewer}}.
\end{abstract}

\section{Introduction}


Meta-reviewing is a critical step in the overall scientific peer-review process, which focuses on understanding the consensus of expert opinions on a scholarly work and making informed judgments on its scientific merit~\cite{fernandes2022between}. Meta-reviews are also helpful as they concisely summarize the strengths and weaknesses of the manuscript and highlight the scope for further improvement. Composing a meta-review from peer reviewers' comments includes the following steps:

\begin{enumerate}[leftmargin=*,itemsep=0ex,partopsep=0.3ex,parsep=0ex]

    \item Collect expert opinions on the paper from peer researchers in the same/similar domain.
    
    \item Analyze the comments and opinions from different reviewers and comprehend the common strengths, weaknesses, and suggestions for improvement emphasized by the peer reviewers and summarize them into a holistic overview.

    \item Present meta-reviewer's own judgment about the paper (resolve or arbitrate conflicts among reviewers) and make a final recommendation.
\end{enumerate}

The process of meta-reviewing is arduous and time-consuming~\cite{MetaWriter}. In addition, over the years, the number of research manuscript submissions has been rising exponentially~\cite{article}. As such, the volume of meta-reviewing tasks is also rising exponentially, making the job of meta-reviewers even more challenging and time-consuming. 
Since human beings are prone to fatigue, bias, and distraction~\cite{goldberg2023peer}, it is possible that this large workload may make meta-reviewers susceptible to inconsistency and missing details. To address this issue, we performed a case study using LLMs to generate multi-perspective summaries (MPSs)~\cite{SOFSAT,bansal-etal-2022-semantic} of reviewers' comments to see whether LLM-generated summaries can assist meta-reviewers in being more consistent and efficient in this important role. Although several works have proposed automated meta-review generation techniques~\cite{zeng2024,bhatia2020metagen,pradhan2021deep,shen2022mred}, our work is different in three aspects: 1) Instead of meta-review generation, we focus on the MPS task with a goal to assist the human meta-reviewer, 2) We study LLMs in combination with a taxonomy-based prompting technique (e.g., TELeR~\cite{santu2023teler}), which is currently understudied, 3) We examine the correlations between human and automatic (GPT-4) evaluations to explore the reliability of LLM-based assessments.

\medskip
\noindent \textbf{\underline{Case Study}:}  For our case study, we selected 40 research papers submitted to the ICLR Conference in recent years, along with peer-reviewer comments and a handcrafted meta-review written by a highly experienced researcher for each article. These manuscripts and the associated peer-review comments and meta-reviews are all publicly available on OpenReview.net.  Using this data set in combination with the three LLMs (GPT-3.5, PaLM2, LLaMA2) and the TELeR taxonomy, we performed an extensive qualitative analysis by engaging human evaluators to go through the meticulous process of reading and analyzing $40$ individual paper abstracts, $120$ $(40\times 3)$ associated peer-review narratives, $40$ corresponding human-written meta-reviews, and a total of $480$ distinct LLM-generated MPSs\footnote{Total comes from 3 LLMs $\times$ 4 prompt levels $\times$ 40 papers $\times$ 5 evaluation criteria $\times$ 2 types (precision/recall)}  and then collected comprehensive survey data, which includes $4800$ granular manuscript level judgments on five specific criteria and $90$ LLM level judgments\footnote{Ten annotators judge 3 LLMs over $3$ questions for MPS task (Total = $10\times 3 \times 3$)} from humans. Although there are several popular text generation evaluation metrics such as ROUGE, SEM-F1, BERTScore, etc.~\cite{sai2022survey}, we exclusively focus on meticulous qualitative human evaluation because previous studies have reported several limitations of automated evaluation metrics for the text generation task~\cite{akter-etal-2022-revisiting,DBLP:sofsat-sem-f1}. In addition to human judgments, we also conducted a GPT-4-based evaluation of MPSs and examined the correlation between human and automatic evaluations to explore the reliability of LLM-based assessments of MPSs along five specific criteria.


\medskip
\noindent \textbf{\underline{Findings}:} Our qualitative analysis reveals that regarding the quality of the generated MPSs, GPT-3.5 and PaLM2 performed comparably and were rated higher by humans than LLaMA2 in terms of manuscript-level judgments. Interestingly, PaLM2 generally yielded better recall scores, while GPT-3.5 yielded better precision scores. More surprisingly, GPT-3.5 was rated poorly by humans at LLM-level judgments, which demands further investigation. On the other hand, the automatic evaluation shows that the GPT-4 evaluation correlates poorly with human judgments when tasked with evaluating the meta-reviews generated by its predecessors (i.e., GPT-3.5). Our findings also highlight that LLM-based evaluations are unreliable in tasks that require aspect-aware summary evaluation, especially when dealing with narratives from multiple diverse perspectives.



\section{Related Work}
\label{sec:RelatedWork}


Over the past few years, several studies have been accomplished on meta-review generation. Some researchers have employed rule-based techniques in conjunction with deep neural network architectures \cite{bhatia2020metagen,pradhan2021deep}, while others proposed encoder-decoder-based methods \cite{kumar2021deep,li2023summarizing} to tackle the meta-review generation task. Recent developments with Large Language Models (LLMs), including but not limited to GPT-3.5 and GPT-4~\cite{brown2020language,ouyang2022training}, PaLM~\cite{thoppilan2022lamda}, LLaMA \cite{touvron2023llama},  Bloom~\cite{scao2022bloom}, GLaM~\cite{du2022glam}, have shown remarkable performance in traditional summarization tasks. Yet, their performance on complex, i.e., multi-constraint multi-document summarization tasks~\cite{DBLP:sofsat-sem-f1} like meta-review composition is still understudied.




Recently, \citet{du2024llms} examined the effectiveness of large language models (LLMs) in both reviewing and meta-reviewing research papers. Similarly, \citet{tyser2024ai} evaluated LLMs' ability to review papers and compared their quality with human-written peer reviews. In another line of work,  \citet{zeng2024} introduced a Checklist-guided Iterative Introspection (CGI2) approach using LLMs, which breaks down scientific opinion summarization into several stages. However, LLMs are quite sensitive to prompts that are fed to them~\cite{loya2023exploring,sclar2023quantifying}, and therefore, designing appropriate prompts with the right amount of detail has become more important than ever~\cite{liu2023pre,han2022ptr}. This issue worsens in practice as there is no standard taxonomy of prompts to follow for benchmarking LLMs on complex tasks that the community has a general consensus on yet. As such, the main motivation of this work, which makes our case study more compelling, stems from the following two gaps in the literature: 1) LLM's performance on multi-constraint multi-perspective summarization tasks (e.g., meta-review composition) is still understudied, 2) no study has yet followed a standard prompting taxonomy to systematically compare LLMs on such complex tasks.

Fortunately, very recently, \citet{santu2023teler} have proposed a general prompting taxonomy called TELeR, which can serve as a unified standard for comparing and benchmarking LLMs' performances on complex generation tasks (details in appendix). In this work, we leveraged this TELeR taxonomy for designing prompts in the context of a meta-review composition task. We reported the specific categories of prompts we experimented with across multiple LLMs. This enabled more meaningful comparisons among the three LLMs we studied and, thereby, helped to derive more accurate conclusions about their relative performances.



\section{Case Study} \label{sec:case-study}

This section describes the details of our case study, including the data set, LLMs, evaluation criteria, and prompting details.


\subsection{Data-set Details}

For evaluation, we carefully selected 40 research papers and their associated peer-review comments and meta-reviews from the collection of more than $13,800$ research manuscripts submitted to the ICLR conference during a span of four years, i.e., the year 2020 through 2023. While selecting papers for our evaluation, we followed some general principles, as mentioned below.

\begin{enumerate}
    
    \item Meta-review is detailed and substantially encompasses the core contributions, strengths, weaknesses, and scopes for improvements.
    
    \item Manuscript received at least 3 detailed reviews that comment on different aspects, e.g., core contributions, strengths, weaknesses, suggestions for improvement, and literature review quality.
    
    \item Preference was given to include more recent papers, especially those submitted during 2021-2023 dealing with diverse topics.
    
    \item The corpus also includes 10 rejected papers (25\%) to increase meta-review diversity.

\end{enumerate}

\subsection{Large Language Models}

As mentioned previously, our case study evaluates 3 recent popular LLMs - GPT-3.5~\cite{brown2020language}, PaLM2~\cite{chowdhery2023palm}, LLaMA2~\cite{touvron2023llama2}. Prompting of GPT-3.5 and LLaMA2 (13B) was performed through their API, but interaction with Google's PaLM2 was done manually through a web browser because PaLM2 does not provide an API. As Zero-shot learning techniques have become very popular in recent years~\cite{DBLP:conf/emnlp/SarkarFS23,DBLP:conf/ijcnlp/SarkarFS22}, we use all LLMs in the Zero-shot setting for prompting without further fine-tuning.

\subsection{Prompt Design using TELeR Taxonomy}

We prompted LLMs with different levels of prompts designed based on the recently proposed TELeR taxonomy (described in section~\ref{sec:background}). To be more specific, let us look at what the TELeR levels meant for our MPS task. In all the following examples, we assume that a prompt is a concatenation of directive text (instruction) followed by the peer review texts.


\begin{itemize}[leftmargin=*,itemsep=0.4ex,partopsep=0.2ex,parsep=0ex]
    \item \underline{Level 1}: This high-level directive asks the LLMs to simply generate a meta-review without providing any further specific requirements. 

    \item \underline{Level 2}: At level 2, directives include a paragraph with multiple sentences describing instructions/questions articulating multiple sub-tasks.

    \item \underline{Level 3}: Directives are a bulleted list in this level, where each bullet describes a particular sub-task. 

    \item \underline{Level 4}: This level is similar to level 3. The only difference is that the prompts at this level also ask LLMs to explain their own outputs. 
    
\end{itemize}

As part of the meta-review generation task, LLMs were asked to perform the following sub-tasks for prompts at level 2 and above. 

\begin{enumerate}[leftmargin=*,itemsep=0.4ex,partopsep=0.5ex,parsep=0ex]
    \item What is the summary of core contributions? Provide an answer with supporting evidence.
    
    \item Which common strengths are referred to in the reviews? A common strength is a strength that is mentioned in at least two reviews.

    \item What common weaknesses are described in the reviews? A common weakness is a weakness that is mentioned in at least two reviews.

    \item What suggestions for improvement have been provided by three reviews? A common suggestion for improvement is a suggestion that is mentioned in at least two reviews.
    
    \item Do the reviews mention about missing references? A list of missing references is optional.
\end{enumerate}

For further details, the exact prompts used in our case study are shown in the appendix (see Table \ref{tbl_prompt_design}).

\subsection{Evaluation Criteria and Process}


For qualitative evaluation, $10$ human annotators, all researchers in NLP and co-authors of this paper, were involved as volunteers. Each annotator was assigned to assess LLM-generated MPSs for $4$ research papers. For each paper, an annotator was provided with MPSs generated by $3$ LLMs using $4$ prompts at $4$ different TELeR levels. Each annotator evaluated the LLM-generated MPS in light of publicly available reviews posted by the peer-reviewers corresponding to a particular manuscript. It was not disclosed to them which LLM they were evaluating to preclude any psychological bias in the evaluation. All of them were briefed in detail about the evaluation task. Two separate evaluation forms were designed to collect their judgments: 

\smallskip
\begin{enumerate}[leftmargin=*,itemsep=0.5ex,partopsep=0.2ex,parsep=0ex]
    \item \textbf{Micro-Evaluation:} The first set of human evaluations captured their judgment on the quality of the MPSs generated by each LLM for each prompt. For a given prompt and a corresponding output from a particular LLM $x$, each human evaluator was asked to judge the generated MPS based on five different criteria as described below. For each criterion, the human evaluators were presented with an assertive statement about that criterion and then asked to provide their judgments by choosing their level of agreement with the given statement in terms of five different labels: 1) Strongly agree, 2) Agree, 3) Neutral, 4) Disagree, and 5) Strongly disagree. The five criteria and their associated assertive statements (against which humans rated their level of agreement) are presented below.

    \smallskip
    \begin{enumerate}[leftmargin=*,itemsep=0.2ex,partopsep=0.2ex,parsep=0ex]

        
        \item \underline{Core Contribution}: \textcolor{blue}{\textit{LLMx generated MPS captured core contributions very well}}.

        
        \item \underline{Common Strengths}: \textcolor{blue}{\textit{LLMx generated MPS captured common strengths very well}}.

        \item \underline{Common Weaknesses}: \textcolor{blue}{\textit{LLMx created MPS captured common weaknesses very well}}.

        
        \item \underline{Common Tips for Improvement}: \textcolor{blue}{\textit{MPS generated by LLM $x$ captured common suggestions for improvement very well}}.

        
        \item \underline{Literature Review}: \textcolor{blue}{\textit{MPS generated by LLM $x$ included the missing references mentioned in the reviews very well}}.
    \end{enumerate}

    
    


\smallskip
    \item \textbf{Macro-Evaluation:} The second set of evaluations was designed to capture human judgments regarding the overall performance of a particular LLM on the MPS task. We call it the ``Macro-Evaluation''. This form requires two types of responses - the first type has three assertive statements concerning LLM's performance in terms of adherence to instructions, ability to create useful MPSs, and matching against actual expert-written meta-reviews. Again, evaluators were asked to rate their level of agreement on a five-point scale - 1) Strongly agree, 2) Agree, 3) Neutral, 4) Disagree, and 5) Strongly disagree, with respect to the following 3 Macro-Evaluation statements.
    
    \smallskip
    \begin{enumerate}[leftmargin=*,itemsep=0.4ex,partopsep=0.2ex,parsep=0ex]
        \item \textcolor{blue}{\textit{``While generating MPSs, LLMx strictly followed prompt instructions''}}.
        \item \textcolor{blue}{\textit{``LLMx generated MPS is useful for preparing a meta-review''}}.
        \item \textcolor{blue}{\textit{``MPS generated by LLMx matches with the actual meta-review to a great extent''}}.
    \end{enumerate}
    \smallskip
The second type of feedback is an essay question - ``Kindly provide comments about LLMx's performance in generating MPSs." where $x$ represents a particular LLM.
    
\end{enumerate}

\section{Results}\label{Results}

\subsection{Results from Micro-Evaluation}
For the micro-evaluation experiments, we focused on answering the following Research Questions (RQs) related to different components of the LLM-generated MPS: 

``\textit{While generating a multi-perspective summary (MPS) based on the consensus among multiple reviewers' comments on a scholarly work, can LLMs properly capture-}''

\begin{enumerate}[leftmargin=*,itemsep=0.5ex,partopsep=1ex,parsep=0ex]
    \item{\textit{\textbf{RQ Mic-1}:  the core contributions of the  work?}}
    \item{\textit{\textbf{RQ Mic-2}: the major strengths of the work?}}
    \item{\textit{\textbf{RQ Mic-3}: the major weaknesses of the work?}}
    \item{\textit{\textbf{RQ Mic-4}: the avenues for improving the work?}}
    \item{\textit{\textbf{RQ Mic-5}: how well the manuscript conducted review of relevant literature?}}
\end{enumerate}



Below, we answer each Micro-Evaluation research question based on $4800$ $(40\times 3 \times 4\times 2\times 5)$ fine-grained human ratings related to $480$ $(40\times 3 \times 4)$ different LLM-generated meta-reviews.

    \begin{figure*}[!htb]
        \begin{subfigure}{0.5\textwidth}
            \includegraphics[width=\textwidth, height=3.2cm]{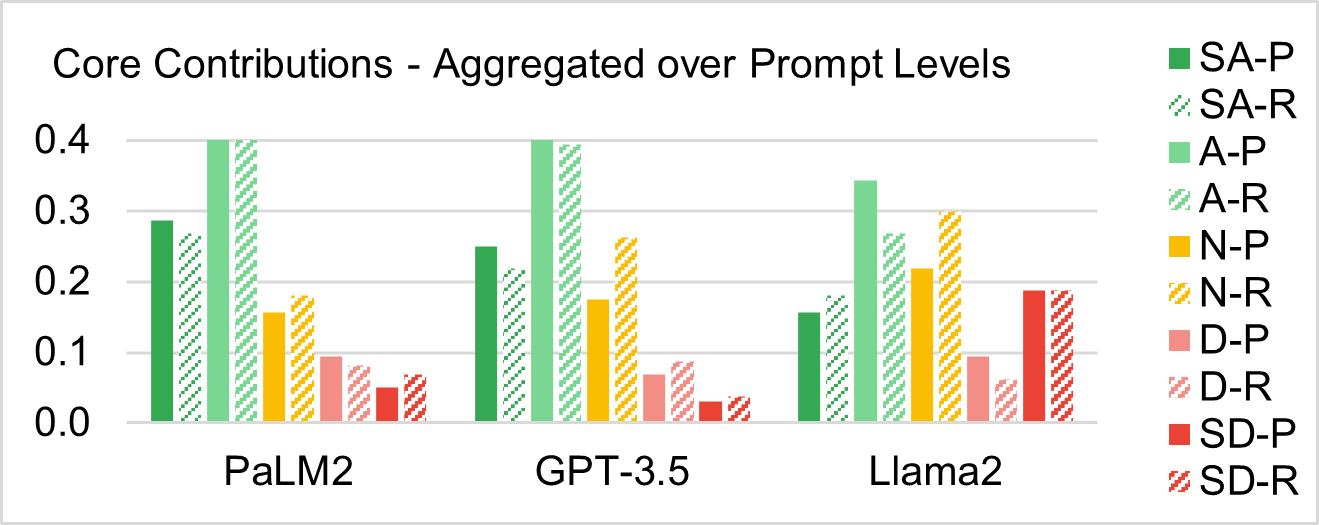}
            \centering            \captionsetup{justification=centering,margin=0cm}
            \caption{Core Contributions - Aggregated over all Prompt Levels}
            \label{subfig:cc-pr-p1-4-a}  
        \end{subfigure}
        \begin{subfigure}{0.5\textwidth}
            \includegraphics[width=\textwidth, height=3.2cm]{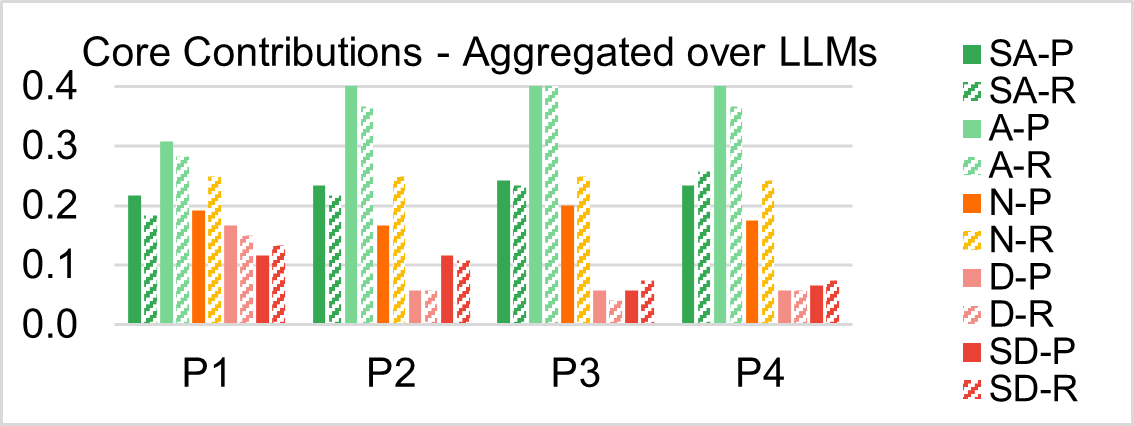}
            \centering            \captionsetup{justification=centering,margin=0cm}
            \caption{Core Contributions - Aggregation over LLMs}
            \label{subfig:cc-pr-p1-4-al} 
        \end{subfigure}
        \caption{Core Contributions Ratings - rated separately across different Prompt Levels and different LLMs. Here, SA: Strongly Agree, A: Agree, N: Neutral, D: Disagree, SD: Strongly Disagree, P: Precision, and R: Recall.}
        \label{fig:cc-ag-pr}
        
    \end{figure*}

\smallskip    
\begin{itemize}[leftmargin=*,itemsep=1ex,partopsep=0.5ex,parsep=0ex]
    \item{\textit{\textbf{RQ Mic-1 (Core Contributions):}}}  We answer this question by analyzing Figure~\ref{fig:cc-ag-pr} and Figure~\ref{fig:cc-pr} (in appendix), which depicts the comparative performances of three LLMs in terms of their (human-perceived) precision/recall distributions, where human-perceived precision and recall distributions are derived from their qualitative rating counts, i.e., \{SA=Strongly Agree, A=Agree, N=Neutral, D=Disagree, SD=Strongly Disagree\}, followed by a normalization term to convert them into a valid distribution (individual values ranging between [0-1]). To be more specific, while computing the precision/recall distributions corresponding to RQ-Mic-1, human annotators independently rated their agreement with the following two statements.

    \begin{itemize}[leftmargin=*,itemsep=0.5ex,partopsep=0.2ex,parsep=0ex]
        \item \textbf{\textit{RQ Mic-1-P: Core Contributions (Precision)}}: \textcolor{blue}{\textit{``While generating meta-review, LLM $x$ was precise in capturing the core contributions as highlighted by at least two reviewers.''}}
        \item \textbf{\textit{RQ Mic-1-R: Core Contributions (Recall)}}: \textcolor{blue}{\textit{``While generating meta-review, LLM $x$ indeed covered all the core contributions highlighted by at least two reviewers.''}}
    \end{itemize}
    \smallskip
    Individual Plots for each TELeR level (1-4) are provided in Figure~\ref{fig:cc-pr} of the appendix due to lack of space. Figure~\ref{fig:cc-pr} shows that PaLM2 was rated the highest for prompt level 1 (minimum details), while GPT-3.5 was rated the best in the case of levels 2 - 4 (more details) for both precision and recall. This suggests that GPT-3.5 could understand complex prompts better than PaLM2 and LLaMA2.



    Next, we looked at precision and recall distributions for each LLM by aggregating over all 4 Prompt Levels (Figure~\ref{subfig:cc-pr-p1-4-a}), which revealed that GPT-3.5 and PaLM2 were comparable, while LLaMA2 was often rated as inferior (more Disagree and Strong Disagree ratings). Finally, Figure~\ref{subfig:cc-pr-p1-4-al} shows the precision/recall distributions for each prompt level while aggregating human ratings over all three LLMs, indicating that Level 3 and Level 4 prompts were more effective for the meta-review generation task than Level 1 and Level 2, which is intuitive as meta-review generation is a complex task.

    \begin{figure}[!htb]
        \includegraphics[width=0.9\linewidth]{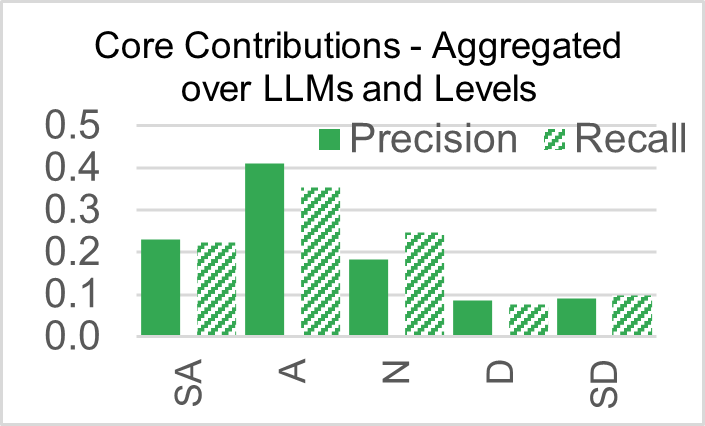}
        \centering            \captionsetup{justification=centering,margin=0cm}
        \caption{Core Contributions - Overall Ratings.}
        \label{fig:cc-pr-p1-4-all}   
        \vspace{-2mm}
    \end{figure}

    Finally, Figure~\ref{fig:cc-pr-p1-4-all} shows the precision/recall distributions aggregated over three LLMs and four prompt levels. This result suggests that humans generally voted in favor of LLMs more often for the meta-review generation task than otherwise. 

    \begin{figure*}[!htb]
        \begin{subfigure}{0.5\textwidth}
            \includegraphics[width=0.95\textwidth, height=3cm]{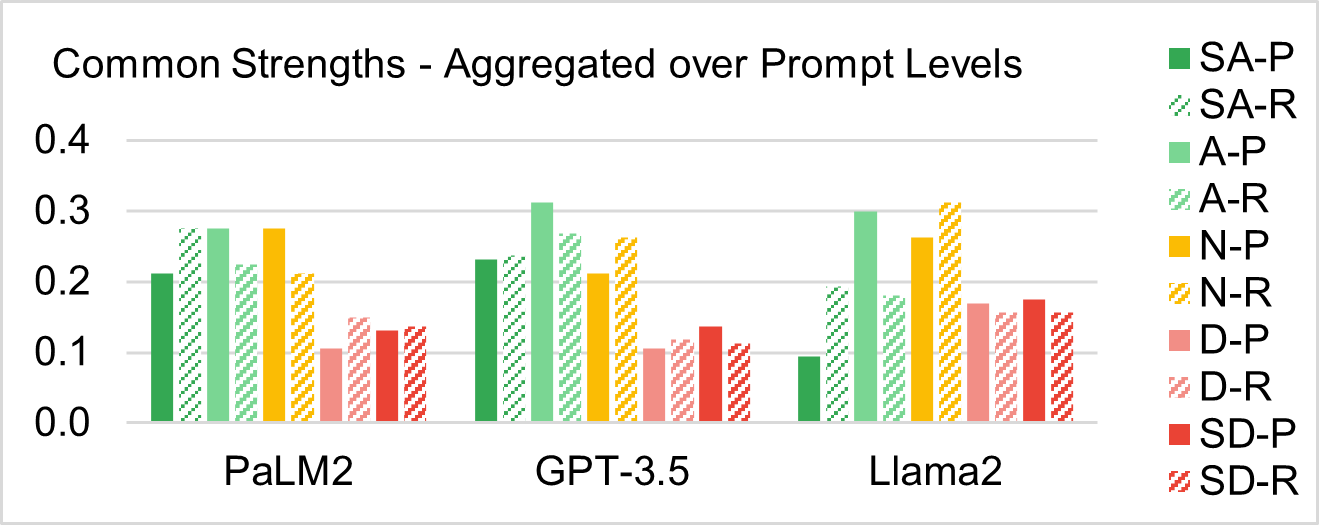}
            \centering            \captionsetup{justification=centering,margin=0cm}
            \caption{Common Strengths - Aggregated over all Prompt Levels}
            \label{subfig:cs-pr-p1-4-a}            
        \end{subfigure}
        \begin{subfigure}{0.5\textwidth}
            \includegraphics[width=0.95\textwidth, height=3cm]{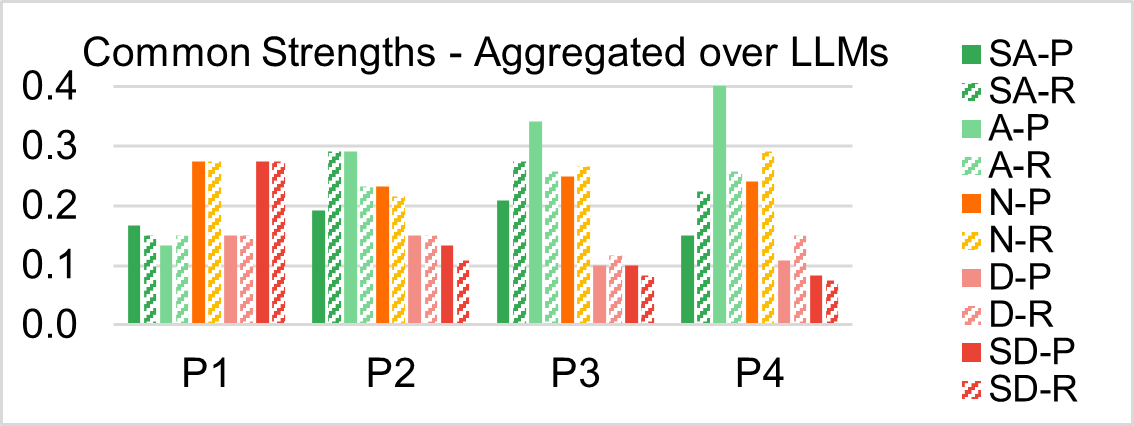}
            \centering            \captionsetup{justification=centering,margin=0cm}
            \caption{Common Strengths - Aggregation over LLMs}
            \label{subfig:cs-pr-p1-4-al}            
        \end{subfigure} 
        ~
        \begin{subfigure}{0.5\textwidth}
            \includegraphics[width=0.95\textwidth, height=3cm]{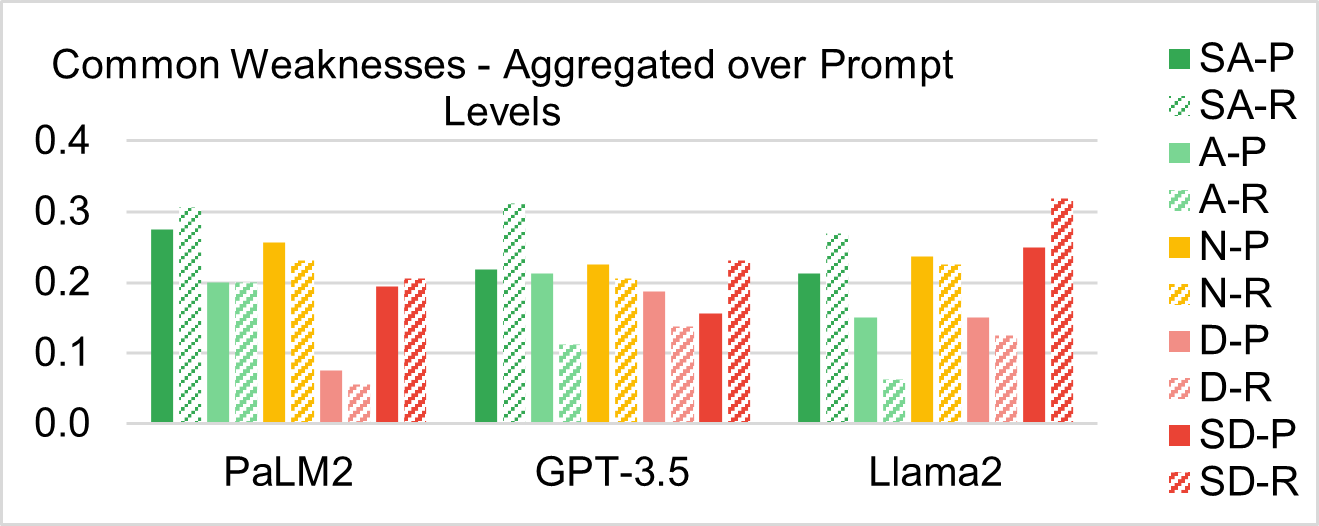}
            \centering            \captionsetup{justification=centering,margin=0cm}
            \caption{Common Weaknesses - Aggregated over all Prompt Levels}
            \label{subfig:cw-pr-p1-4-a}            
        \end{subfigure}
        \begin{subfigure}{0.5\textwidth}
            \includegraphics[width=0.95\textwidth, height=3.1cm]{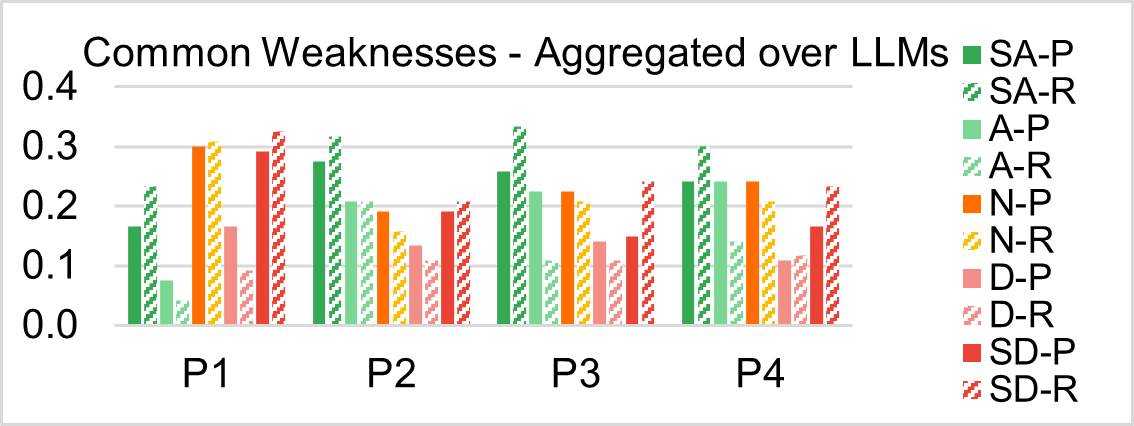}
            \centering            \captionsetup{justification=centering,margin=0cm}
            \caption{Common Weaknesses - Aggregation over LLMs}
            \label{subfig:cw-pr-p1-4-al}            
        \end{subfigure}
        ~
        \begin{subfigure}{0.5\textwidth}
            \includegraphics[width=0.95\textwidth, height=3cm]{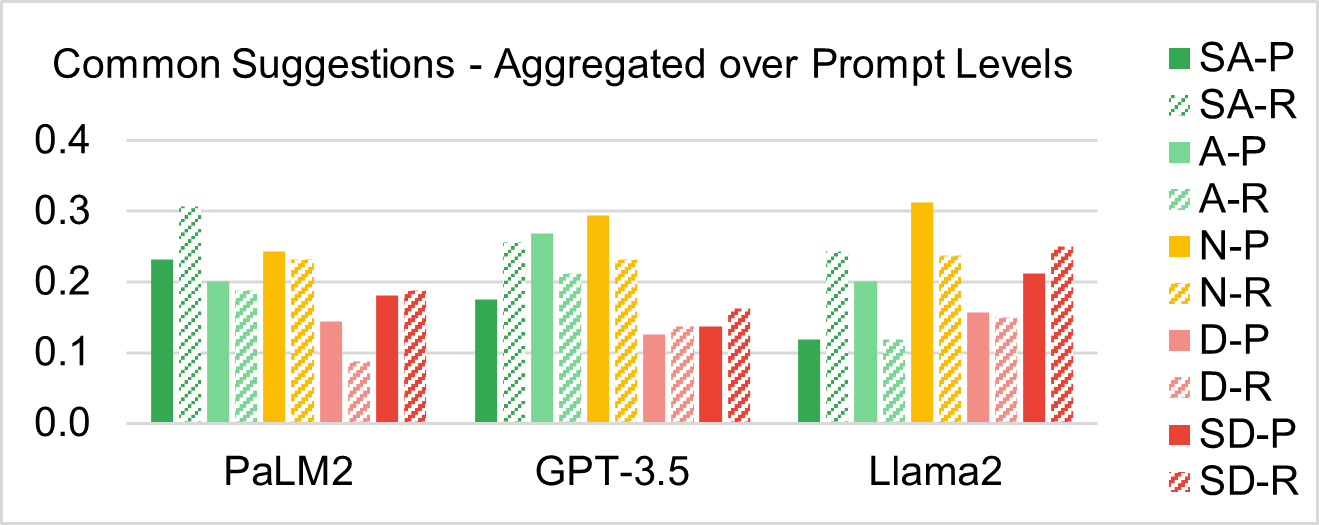}
            \centering            \captionsetup{justification=centering,margin=0cm}
            \caption{Common Suggestions - Aggregated over all Prompt Levels}
            \label{subfig:csu-pr-p1-4-a}            
        \end{subfigure}
        \begin{subfigure}{0.5\textwidth}
            \includegraphics[width=0.95\textwidth, height=3cm]{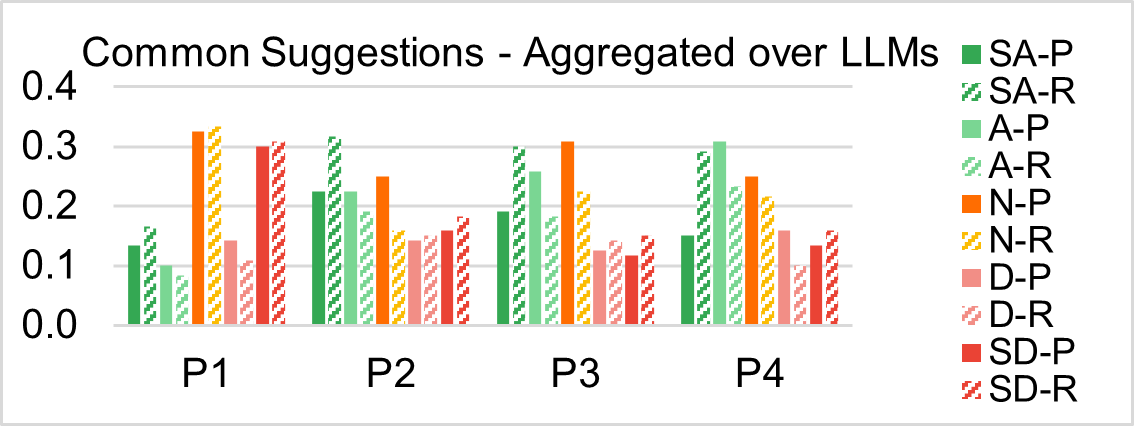}
            \centering            \captionsetup{justification=centering,margin=0cm}
            \caption{Common Suggestions - Aggregation over LLMs}
            \label{subfig:csu-pr-p1-4-al}            
        \end{subfigure}
        ~
        \begin{subfigure}{0.5\textwidth}
            \includegraphics[width=0.95\textwidth, height=3cm]{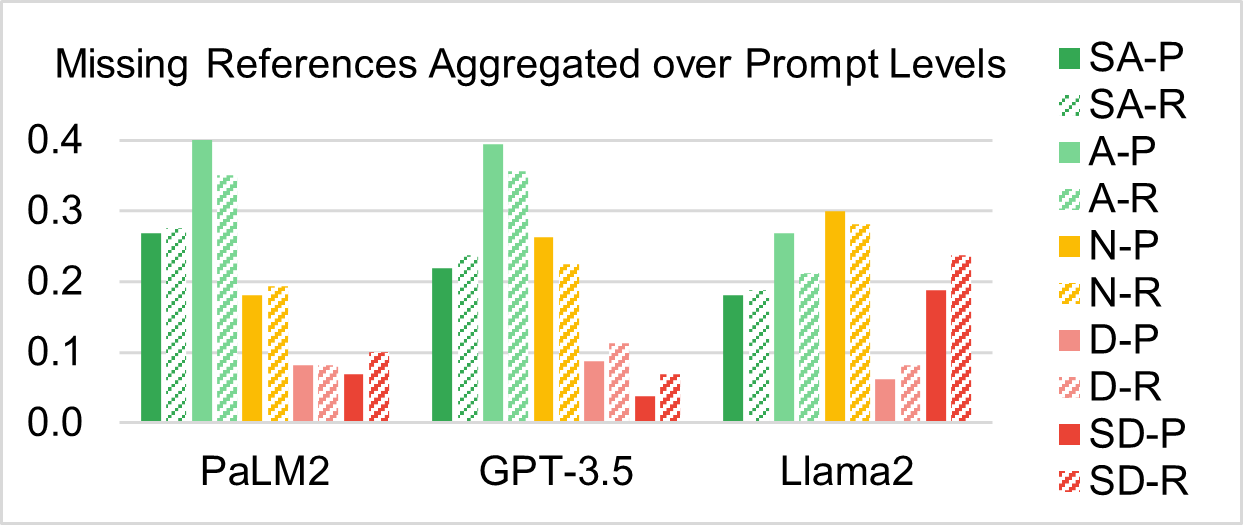}
            \centering            \captionsetup{justification=centering,margin=0cm}
            \caption{Literature Review - Aggregated over all Prompt Levels}
            \label{subfig:mr-pr-p1-4-a}            
        \end{subfigure}
        \begin{subfigure}{0.5\textwidth}
            \includegraphics[width=0.95\textwidth, height=3cm]{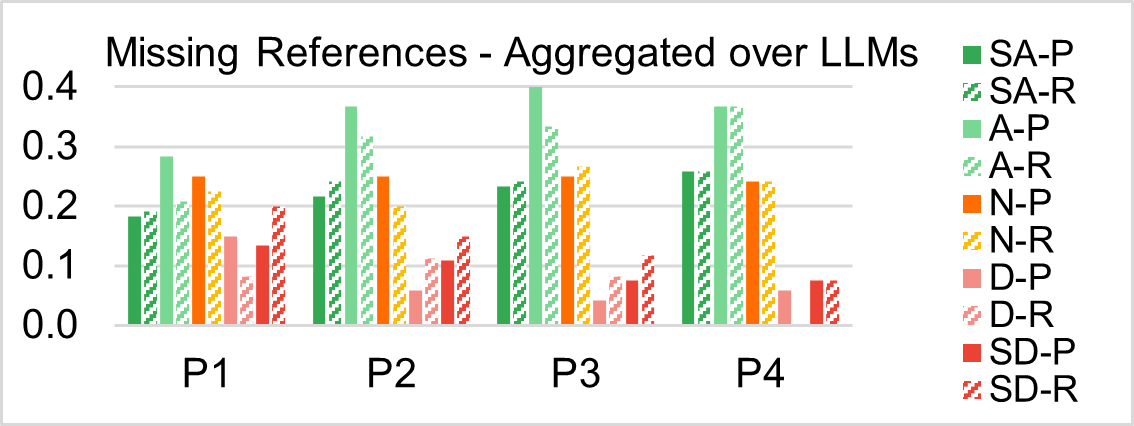}
            \centering            \captionsetup{justification=centering,margin=0cm}
            \caption{Literature Review - Aggregation over LLMs}
            \label{subfig:mr-pr-p1-4-al}       
        \end{subfigure}
        \caption{Ratings of Four Criteria (Common Strengths, Common Weaknesses, Common Suggestions, and Literature Review Quality) - Aggregated separately across different Prompt Levels and different LLMs.}
        \label{fig:overall-aggregate} 
        \vspace{1mm}
    \end{figure*}

\begin{figure*}[!htb]
        \begin{subfigure}{0.24\textwidth}
            \includegraphics[width=\textwidth]{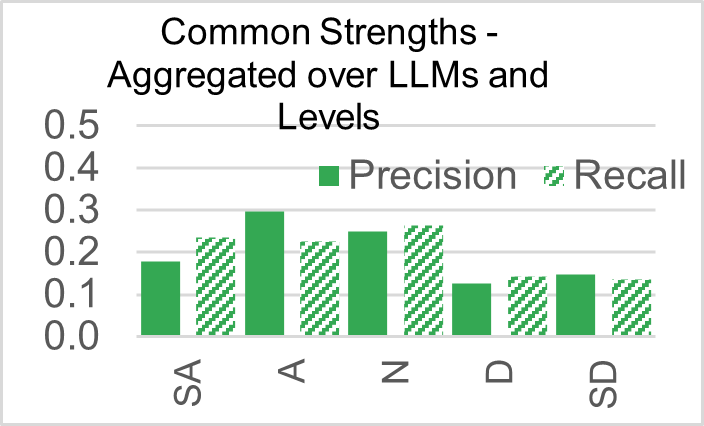}
            \centering            \captionsetup{justification=centering,margin=0cm}
            \caption{Common Strengths}
            \label{subfig:cs-pr-p1-4-all}            
        \end{subfigure}
        \begin{subfigure}{0.24\textwidth}
            \includegraphics[width=\textwidth]{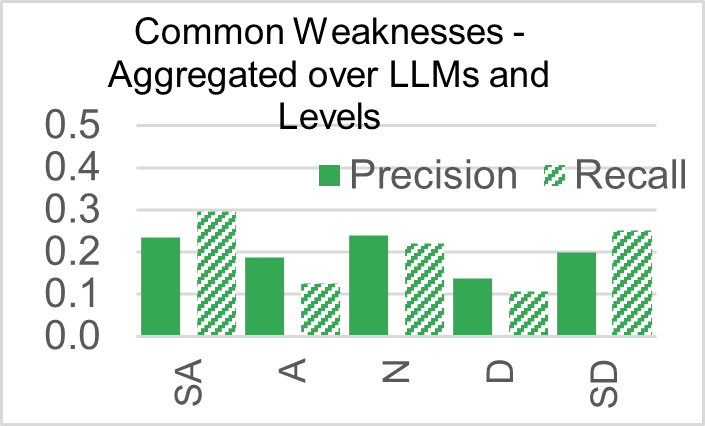}
            \centering            \captionsetup{justification=centering,margin=0cm}
            \caption{Common Weaknesses}
            \label{subfig:cw-pr-p1-4-all}
        \end{subfigure}
        \begin{subfigure}{0.24\textwidth}
            \includegraphics[width=\textwidth]{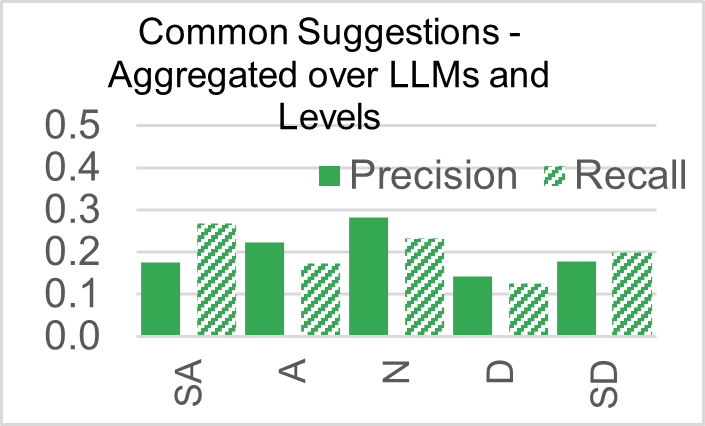}
            \centering            \captionsetup{justification=centering,margin=0cm}
            \caption{Common Suggestions}
            \label{subfig:csu-pr-p1-4-all}            
        \end{subfigure}
        \begin{subfigure}{0.24\textwidth}
            \includegraphics[width=\textwidth]{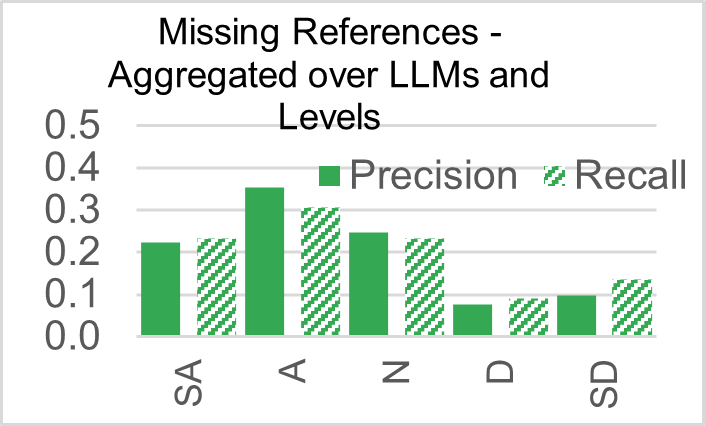}
            \centering            \captionsetup{justification=centering,margin=0cm}
            \caption{Literature Review}
            \label{subfig:mr-pr-p1-4-all} 
        \end{subfigure}
        \caption{Overall Rating aggregated over three LLMs and four Prompt Levels.}
        \label{fig:overall} 
        \vspace{-2mm}
\end{figure*}


    \item{\textit{\textbf{RQ Mic-2 (Common Strengths):}}} In the case of Common strengths, Figure~\ref{subfig:cs-pr-p1-4-a} reveals that GPT-3.5 and PaLM2 were comparable, while LLaMA2 was often rated as slightly inferior. On the other hand, Figure~\ref{subfig:cs-pr-p1-4-al} demonstrates that the performance of LLMs improves substantially from level 1 to level 2 prompts. There is further improvement in the precision of LLMs as we go to higher levels of prompts. But this is not true in the case of recall, where the performance keeps improving till level 3 and then goes down.

    
    \item{\textit{\textbf{RQ Mic-3 (Common Weaknesses):}}} We notice a similar comparative performance of LLMs (Figure~\ref{subfig:cw-pr-p1-4-a}) and Prompt Levels (Figure~\ref{subfig:cw-pr-p1-4-al}) in the case of Common weaknesses as we observed for Common Strengths. 
    
    
    \item{\textit{\textbf{RQ Mic-4 (Common Suggestions):}}} We again notice a similar comparative performance of LLMs (Figure~\ref{subfig:csu-pr-p1-4-a}) and Prompt Levels (Figure~\ref{subfig:csu-pr-p1-4-al}) in the case of Common suggestions as we observed for Common Strengths and Weaknesses.

    
    \item{\textit{\textbf{RQ Mic-5 (Literature Review):}}} We notice GPT-3.5 and PaLM2 were rated more favorably than LLaMA2 (Figure~\ref{subfig:mr-pr-p1-4-a}) by the human annotators for the Literature Review criterion. Further, as Figure~\ref{subfig:mr-pr-p1-4-al}) demonstrates, the performance of LLMs improves substantially from level 1 to level 2 prompts. However, we did not find appreciable improvement in performance by upgrading from level 2 to levels 3 and/or 4.

\end{itemize}

\noindent{\underline{\textbf{Summary of Micro-Evaluation}}:}  Figure~\ref{fig:overall} shows the precision/recall distributions aggregated over three LLMs and four prompt levels for the following aspects of LLM-generated meta-reviews: Common Strengths, Common Weaknesses, Common Suggestions and Literature Review. This result (in combination with Figure~\ref{fig:cc-pr-p1-4-all}) suggests that humans generally voted in favor of LLMs more often for the Core Contributions, Common Strengths, and Literature Review aspects. Common weaknesses and common suggestions were evaluated with mixed ratings, suggesting LLMs struggled more along these aspects. Finally, an overall precision and recall score was computed by assigning the following scores to ratings, i.e., Strongly Agree = 4, Agree = 3, Neutral = 2, Disagree = 1, and Strongly Disagree = 0, and, thereafter, adding the scores of 4 prompt styles and dividing the sum by the maximum possible score to get a normalized score in the range [0,1]. Table~\ref{tbl_case_study_part1} summarizes these scores, where we observe the following:

\begin{enumerate}[leftmargin=*,itemsep=0.5ex,partopsep=0.2ex,parsep=0ex]
    \item GPT-3.5 and PaLM2s were rated higher than LLaMA2 by humans in all five aspects.
    \item For the ``Core Contributions''  and ``Common suggestions'' aspects, PaLM2 and GPT-3.5 were comparable and better than LLaMA2.
    \item GPT-3.5 was rated the highest in the ``Common Strengths'' aspect.
    \item PaLM2 was rated the highest for ``Common weaknesses'' and ``Literature Review'' aspects.
    \item PaLM2 yielded better Recall scores in general (except for Common Strengths), while GPT-3.5 yielded better Precision scores in general (except in the cases of Common weaknesses and Literature Review).
\end{enumerate}

\begin{table}[!hbt]\small
\begin{adjustbox}{width=1\linewidth,center}

\begin{tabular}{c|c|c|c|c} 
 \hline
  \textbf{Aspect} & \textbf{P/R} & \textbf{PaLM2} & \textbf{GPT-3.5} & \textbf{Llama2} \\
 \hline\hline
 {Core}
  & P & 0.698 & \textbf{0.711} & 0.547 \\ 
  Contributions & R & \textbf{0.680} & 0.667 & 0.548 \\\cline{2-5}
  \hline
 {Common} & P & 0.583 & \textbf{0.598} & 0.492 \\ 
  Strengths & R & 0.588 & \textbf{0.600} & 0.525 \\\cline{2-5}
 \hline
 {Common} & P & \textbf{0.572} & 0.538 & 0.481 \\ 
  Weaknesses & R & \textbf{0.586} & 0.534 & 0.459 \\\cline{2-5}
 \hline
 {Common} & P & 0.539 & \textbf{0.555} & 0.464
 \\ 
  Suggestions & R & \textbf{0.584} & 0.566 & 0.489
 \\\cline{2-5}
 \hline
 {Literature} & P & \textbf{0.680} & 0.667 & 0.548
 \\ 
 Review & R & \textbf{0.655} & 0.645 & 0.508
 \\\cline{2-5}
 \hline
\end{tabular}
\end{adjustbox}
\vspace{-2mm}
\caption{\label{tbl_case_study_part1}Scores - Case Study Micro Evaluation.} 

\end{table}
\begin{figure*}[!htb]
        \begin{subfigure}{0.48\textwidth}
            \includegraphics[width = \textwidth]{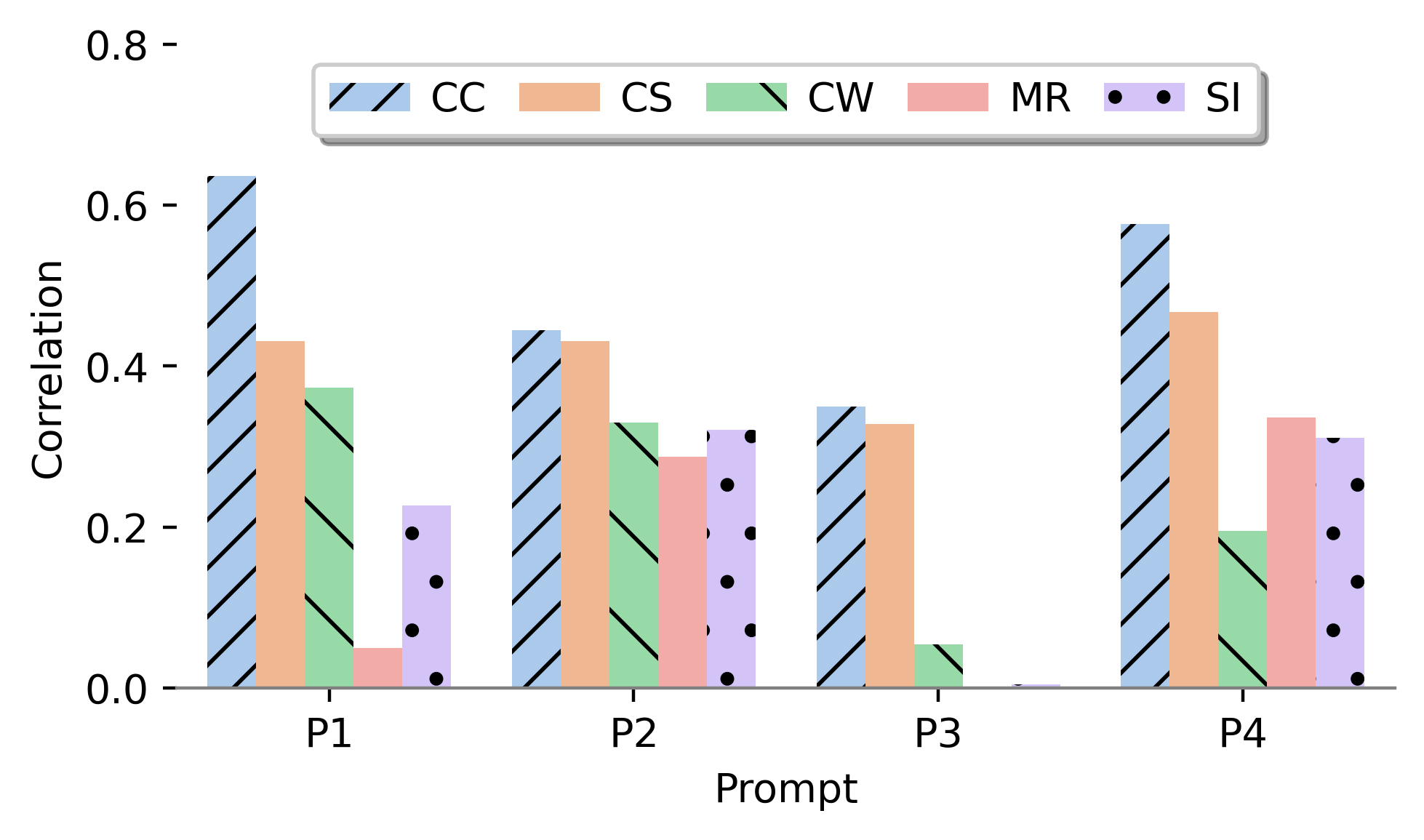}
            \centering            \captionsetup{justification=centering,margin=0cm}
            \caption{Correlation for PaLM2}
            \label{subfig:corr-palm2}            
        \end{subfigure}
        \begin{subfigure}{0.48\textwidth}
            \includegraphics[width = \textwidth]{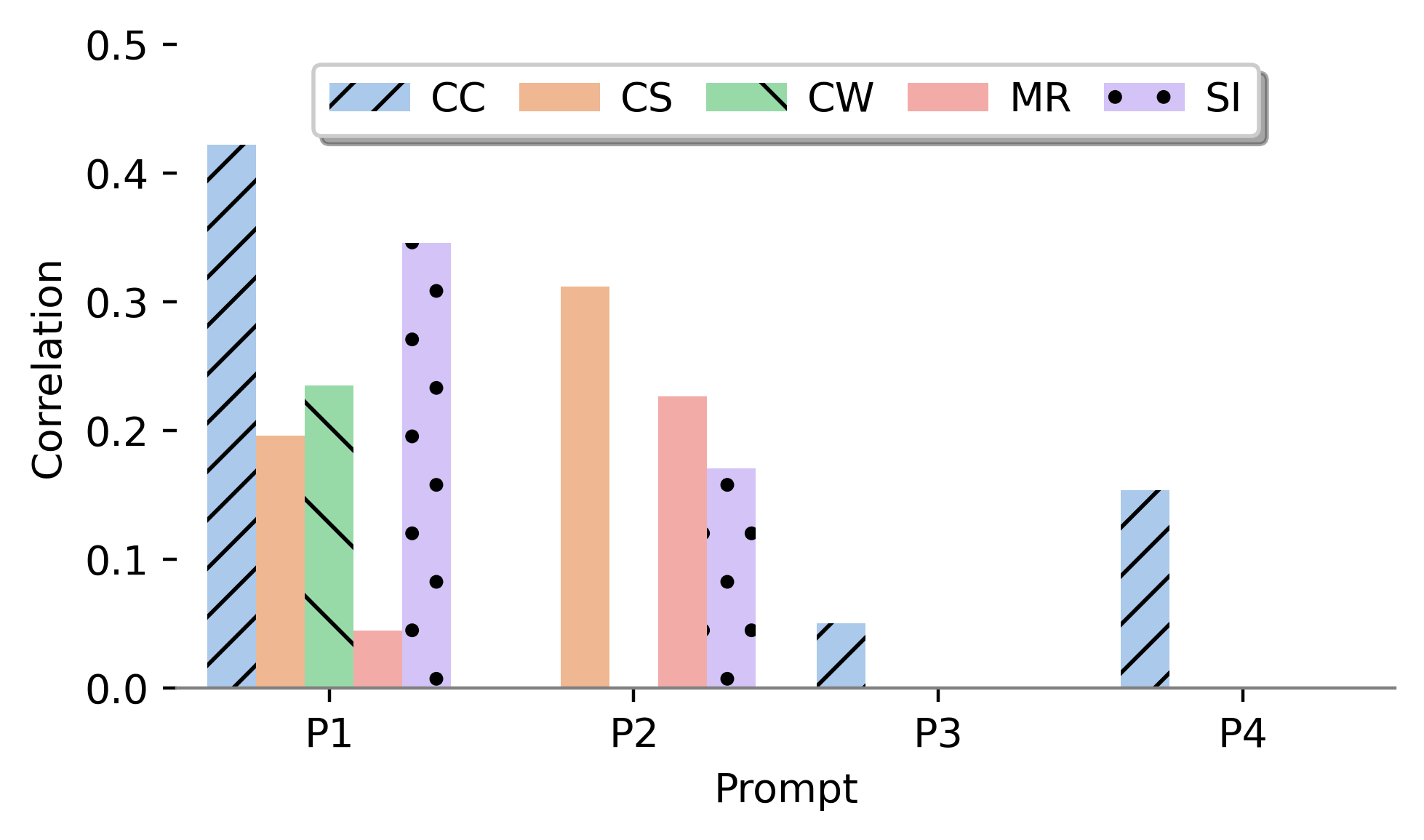}
            \centering            \captionsetup{justification=centering,margin=0cm}
            \caption{Correlation for GPT-3.5}
            \label{subfig:corr-gpt}            
        \end{subfigure}
        \begin{subfigure}{0.48\textwidth}
            \includegraphics[width = \textwidth]{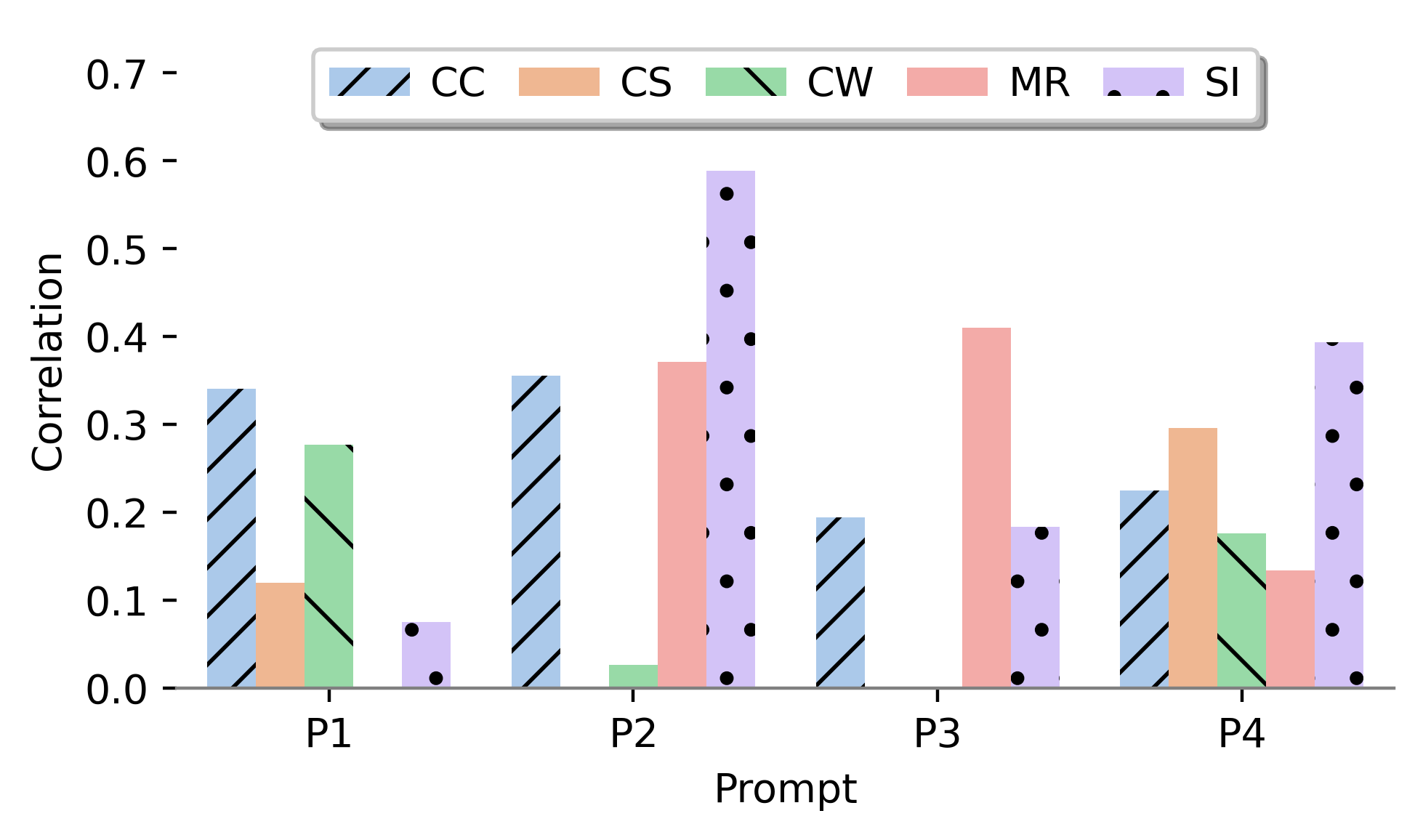}
            \centering            \captionsetup{justification=centering,margin=0cm}
            \caption{Correlation for LLaMA2}
            \label{subfig:corr-llama2}            
        \end{subfigure}
        \begin{subfigure}{0.48\textwidth}
            \includegraphics[width = \textwidth,height=5cm]{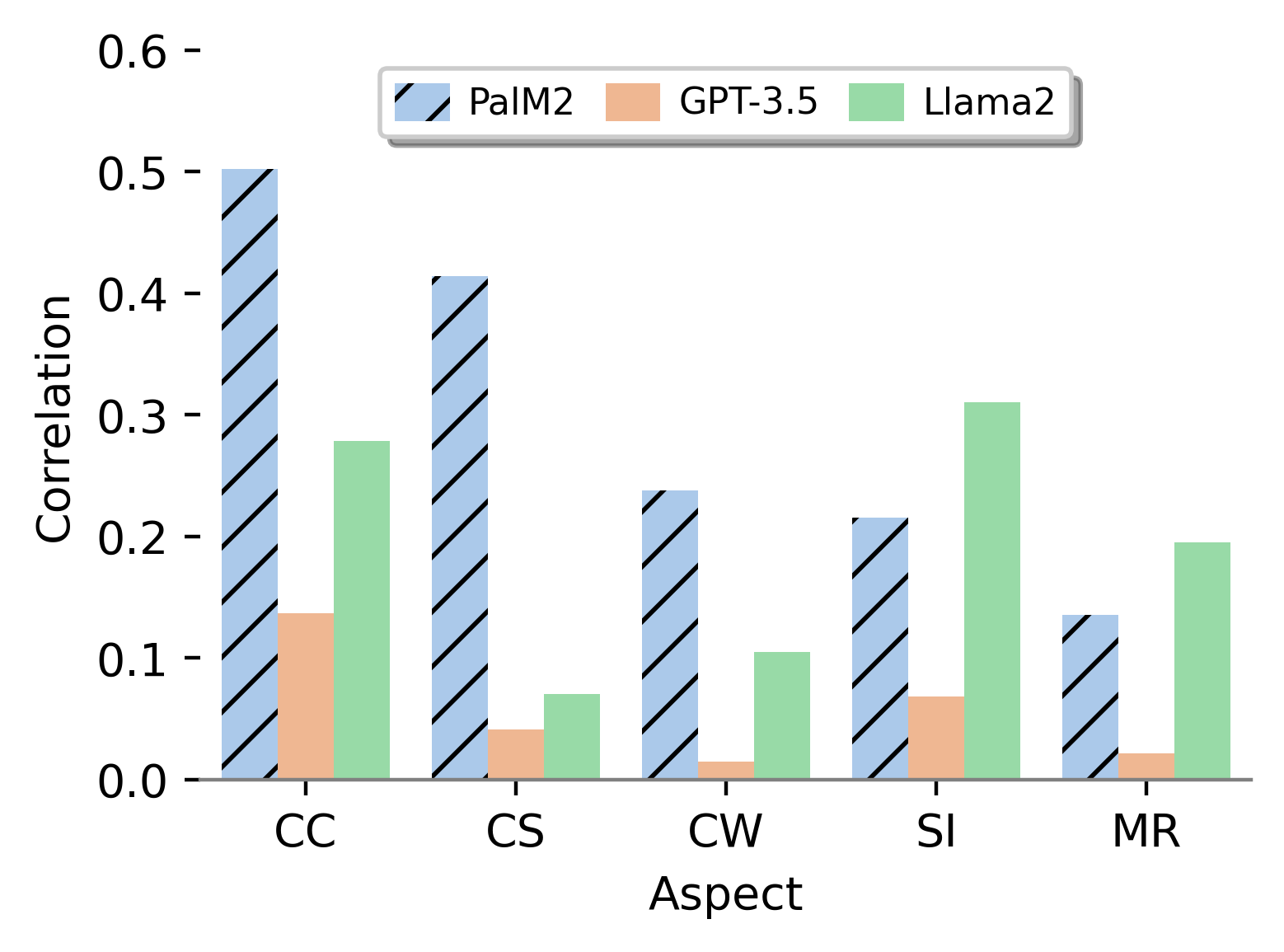}
            \centering            \captionsetup{justification=centering,margin=0cm}
            \caption{Correlation-Aggregated over Prompt Levels}
            \label{subfig:corr-all-p}            
        \end{subfigure}
        ~
        \caption{Pearson correlation between human and GPT-4 evaluations across different aspects of Micro Evaluation (Core Contribution-CC, Common Strengths-CS, Common Weaknesses-CW, Literature Review Quality or Missing References-MR, Suggestions for Improvement-SI). Here, Px indicates the prompt levels.}
        \label{fig:auto-eval} 
        \vspace{-2mm}
    \end{figure*}
        
\subsection{Results from Macro-Evaluation}

For the macro-evaluation experiments, we focused on the following broad research questions related to the overall performance of LLMs.

\begin{enumerate}[leftmargin=*,itemsep=0.5ex,partopsep=0.2ex,parsep=0ex]
    \item{\textit{\textbf{RQ Mac-1}: Can LLMs properly understand the complex requirements and sub-tasks of MPS?}}
    \item{\textit{\textbf{RQ Mac-2}: How useful were the meta-review generated by the LLMs?}}
    \item{\textit{\textbf{RQ Mac-3}: Do LLM-generated meta-reviews match human-expert-written meta-review?}}
\end{enumerate}

\begin{table}[!htb]
\begin{adjustbox}{width=\linewidth,center}

\begin{tabular}{r|c|c|c} 
 \hline
  \textbf{Aspect} & \textbf{PaLM2} & \textbf{GPT-3.5} & \textbf{LLaMA2} \\ [0.5ex] 
 \hline\hline
 Adherence to instructions & \textbf{0.65} & 0.53 & 0.55 \\ 
  \hline
 Meta-review usefulness & \textbf{0.65} & 0.53 & 0.63 \\  
 \hline
 Meta-review matching & \textbf{0.58} & 0.50 & 0.50 \\ 
 \hline
\end{tabular}
\end{adjustbox}
\vspace{-2mm}
\caption{\label{tbl_case_study_part2}Macro Evaluation Results}

\end{table}

\begin{figure*}[!htb]
        \begin{subfigure}{0.48\textwidth}
            \includegraphics[width = \textwidth]{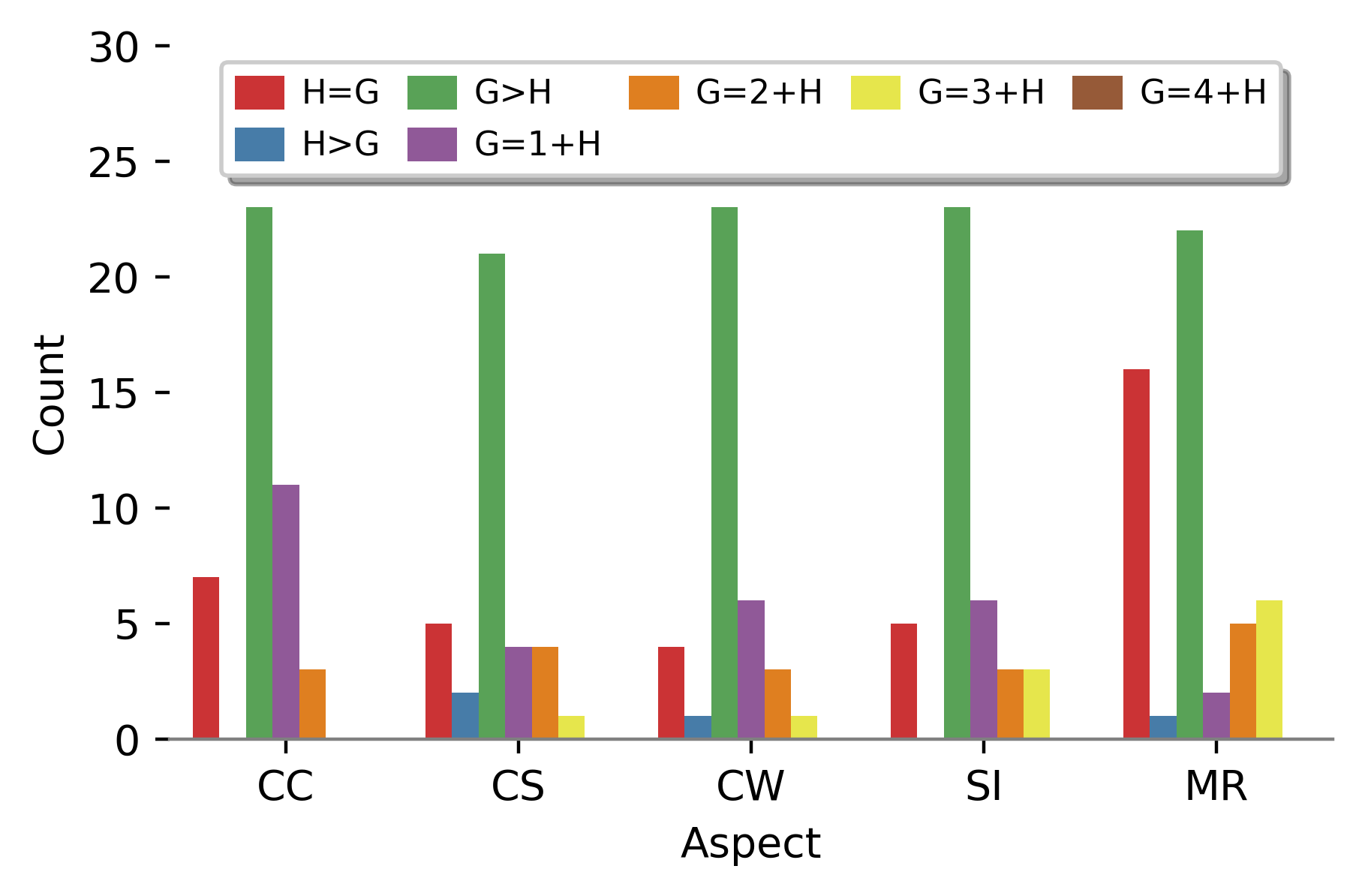}
            \centering            \captionsetup{justification=centering,margin=0cm}
            \vspace{-4mm}
            \caption{TELeR Prompt Level 3}
            \label{subfig:gpt_p3}            
        \end{subfigure}
        \begin{subfigure}{0.48\textwidth}
            \includegraphics[width = \textwidth]{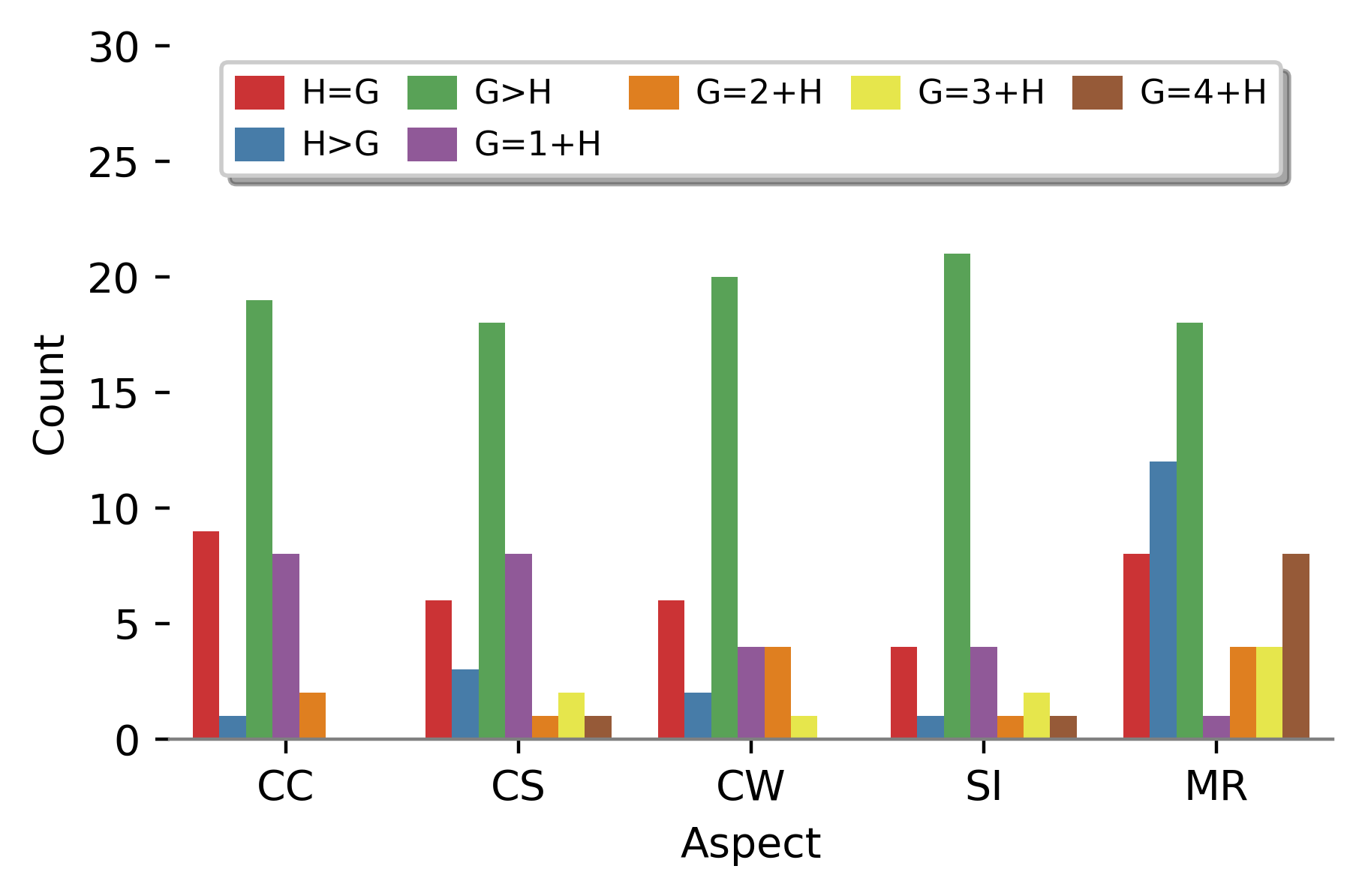}
            \centering            \captionsetup{justification=centering,margin=0cm}
            \vspace{-4mm}
             \caption{TELeR Prompt Level 4}
            \label{subfig:gpt_p4}            
        \end{subfigure}
        ~
        \vspace{-2mm}
        \caption{Count Distribution of Human (H) vs. GPT-4 (G) evaluation scores on GPT 3.5 generated MPS for higher prompt level (level 3 and 4.). Here, \textit{`HG='} indicates the human and GPT-4 give the same scores, \textit{`G$>$H'} indicates  GPT provides a higher score than humans, and  \textit{`G$=$x+H'} indicates  GPT provides x points more score than human.}
        \label{fig:gpt_high_rate} 
        \vspace{-2mm}
    \end{figure*}
\noindent
Human evaluators again rated their agreements with the Macro evaluation statements on the same 5-point scale, i.e., \{SA=Strongly Agree, A=Agree, N=Neutral, D=Disagree, SD=Strongly Disagree\}. Next, a unified quality score was computed by assigning the following scores to ratings, i.e., SA = 4, A = 3, N = 2, D = 1, and SD = 0, and, thereafter, adding the scores of 4 prompt styles and dividing the sum by the maximum possible score to get a normalized score in the range [0,1] (Table~\ref{tbl_case_study_part2}). Surprisingly, Table~\ref{tbl_case_study_part2} shows that GPT-3.5 is the worst performer, whereas PaLM2 is the best in all three aspects at the Macro Level. We were baffled by this outcome and decided to interview the annotators further. Based on the follow-up discussions, annotators generally suggested that micro-evaluation should be more reliable as micro-evaluation judgments are much more fine-grained and focused than macro ones. However, it remains an open question as to why GPT-3.5 was generally rated high in micro-evaluation but not in macro-evaluation.


\subsection{Automatic Evaluation with GPT-4o}
In addition to qualitative human evaluation, we also automatically evaluated the LLM-generated MPSs. While prior work~\cite{zeng2023scientific} employed GPT-based evaluators~\cite{liu2023g} to judge meta-reviews overall quality, we employ them for more fine-grained evaluation. To be more specific, we designed a structured, detailed instruction (Appendix~\S\ref{app:prompt_design}) and prompted the GPT-4o model to asses the five aspects of MPSs using the Likert scale for each LLM.


\smallskip

\noindent\textbf{Correlation in Micro-Evaluation:} Figure~\ref{fig:auto-eval} illustrates the Pearson correlation between human scores and GPT-4o scores for micro-evaluation. We report the mean of the precision and recall correlation scores for each of the five criteria. From Figure~\ref{subfig:corr-all-p}, we can see that the PaLM2 evaluation exhibits a better correlation between humans and GPT-4o across several aspects (i.e., core contributions, common strengths, common weaknesses), while GPT-3.5 demonstrates the weakest correlation when aggregating all the prompt levels. Figure~\ref{subfig:corr-gpt} further reveals that at higher prompt levels (P3 and P4) for GPT-3.5 outputs, GPT-4o evaluations show near-zero correlation with human evaluations. This lack of correlation might be attributed to the phenomenon that GPT-4o almost always rated its predecessors' (i.e., GPT-3.5) generated summaries more favorably than humans. To validate whether this is indeed true, we plot the distribution of GPT-4o and Human evaluation scores on GPT-3.5  generated MPS for prompt levels 3 and 4 (shown in Figure~\ref{fig:gpt_high_rate}). As shown in Figure~\ref{subfig:gpt_p3} and Figure~\ref{subfig:gpt_p4}, GPT-4 unduly assigned a higher score for more than 50\% samples (more than 20 out of 40 papers) across most criteria. There were even instances where GPT scores exceeded human scores by $3$ (G$=$3+H) and $4$ (G$=$4+H) points, which led to a poor correlation with human evaluations.


\begin{table}[!htb]
\begin{adjustbox}{width=\linewidth,center}
\begin{tabular}{r|c|c|c} 
 \hline
  \textbf{Aspect} & \textbf{PaLM2} & \textbf{GPT-3.5} & \textbf{LLaMA2} \\ [0.5ex] 
 \hline\hline
 Adherence to instructions & 0.102 & NaN & \textbf{0.534}  \\ 
  \hline
 Meta-review usefulness & 0.356 & -0.212 & \textbf{0.442} \\  
 \hline
 Meta-review matching & -0.128 & 0.00 & 0.263 \\ 
 \hline
\end{tabular}
\end{adjustbox}
\vspace{-2mm}
\caption{\label{tbl_macro-corr}Correlation for Macro Evaluation}
\vspace{-2mm}
\end{table}

\noindent\textbf{Correlation in Macro-Evaluation:} Table~\ref{tbl_macro-corr} shows Pearson correlation scores for macro-evaluation. Interestingly, the evaluation by GPT-4o shows a better correlation with human evaluations for the LLaMA2-generated MPSs, particularly in adherence and usefulness criteria. In contrast, no significant correlation is observed for the GPT-3.5-generated MPSs. Specifically for the adherence aspect, the \textit{NaN} value is attributed to GPT-4 assigning the same Likert score across all the MPSs. Similarly, looking at the other aspects implies that GPT-4 evaluation over its predecessors did not accurately align with human judgments. In summary, automatic evaluation results reveal that GPT-4o is not a reliable judge when assessing multi-perspective summaries generated from different expert opinions on a scholarly work.

\section{Conclusion}\label{Conclusion}


In this paper, we conducted a case study with three LLMs and four different prompt levels by following the recently proposed TELeR taxonomy to study whether LLMs can be useful in mitigating these issues. To achieve this goal, first, we collected 40 different ICLR research manuscripts, and then, for each manuscript, we used three LLMs separately to compose holistic multi-perspective summaries (MPSs) from available multiple expert reviewers' opinions. Next, we performed rigorous human evaluations to assess the MPSs' reliability. More specifically, we engaged $10$ human evaluators to judge the LLM-generated MPSs on five important criteria, including the quality of capturing Core contributions, Common strengths, Common weaknesses, Common suggestions, and Literature Review. Further, we assess the reliability of GPT-4o as an automatic evaluator for this complex task.


Our case study showed that the reliability of LLMs as Meta-Reviewers' assistants widely varies across different criteria (e.g., core contributions, common strengths, etc.), along with the LLM and prompt type used. Our study also revealed that while summarizing multiple reviewers' comments, GPT-3.5 and PaLM2s were rated higher by humans than LLaMA2 in all five aspects. The performances of PaLM2 and GPT-3.5 were comparable on average, but PaLM2 yielded better Recall ratings in general, while GPT-3.5 yielded better Precision ratings. Although our micro-evaluation indicated PaLM2 and GPT-3.5 to be comparable, macro-evaluation results indicate that humans like PaLM2 generated MPSs more than the GPT-3.5 generated ones. Meanwhile, the automatic evaluation with GPT-4 shows a poor correlation with human evaluations, suggesting that GPT-4 is unreliable for evaluating this facet-aware complex MPS tasks.


\section{Limitations}

This work only analyses three LLMs and 40 research manuscripts due to time and resource constraints. More work will need to be done to increase the number of models tested and draw a more general conclusion on the performance of LLMs on the meta-review composition task. However, it should be noted that the estimated time required to annotate each paper is about 4 hours. Each annotator was assigned four papers for annotation. With ten annotators, a total of 160 hours of annotation effort was spent apart from initial exploration, preliminary studies, and data analysis.

Due to a lack of space, we could not discuss our study on the sensitivity of LLMs regarding prompt variations in detail in the main paper. However, all details have been provided in the appendix for the interested reviewer. To summarize, we found that in terms of the TELeR level of prompts, the performance of LLMs improves substantially from level 1 to level 2 prompts. However, we did not always find appreciable improvement in performance by going from level 2 to level 3 and level 4 prompts.

During annotation, human volunteers did not read the actual research papers; they only read the abstract and the associated reviews, comments, and meta-reviews for that paper. Therefore, the annotators assumed the reviewer's comments were correct and justified. Also, although the human meta-reviewers have access to the original paper in practice, our experiments with LLMs did not have that access.

The NeurIPS experiment of 2022 found various biases and inconsistencies when humans evaluate reviews\footnote{https://blog.ml.cmu.edu/2023/12/01/peer-reviews-of-peer-reviews-a-randomized-controlled-trial-and-other-experiments/}\cite{goldberg2023peer}. Although our annotators are all NLP researchers, it is still possible that those biases and inconsistencies also applies when evaluating meta-reviews.

Finally, when we began this work, we utilized the latest LLM models (i.e., PalM2, LLaMA2, and GPT-3.5) available at the time and conducted a human-based evaluation of the generated meta-reviews. However, as newer models were frequently released, given the high cost of human annotation, we shifted our focus to automatic evaluation using GPT-4. Unfortunately, we found that GPT-4's evaluations correlated poorly with human judgments, making it unreliable for assessing the performance of the latest LLM models on our meta-review generation task. Due to time and budget constraints, conducting another round of human evaluation on the latest LLM models is not feasible. Therefore, we need to consider alternative approaches for this evaluation task in the future.



\section{Ethics Statement}

Regarding ethical concerns for the potential implications of this line of research, we would like to highlight the following points.

The objective of the paper is NOT to have LLMs write the meta-review by itself; rather, the goal is to help meta-reviewers prepare better quality meta-reviews by providing them with a holistic summary of the expert peer reviewers' consensus on the intellectual merits of the paper. The human meta-reviewer should never blindly trust LLMs; rather, they should verify the summary claims first and then write the meta-review manually only after verification of the comments against the paper contents.
 
The paper clarifies in multiple places that the role of the meta-reviewer is not just limited to the summarization of the individual reviews, but in cases, it also involves resolving conflicting viewpoints, which may also include adding their own expert opinion. Another task is to identify low-quality reviews and down-weight them or invite new reviewers. This is why LLM-generated multi-perspective summaries should never be treated as a substitute for meta-review; rather, they should be viewed as additional assistance in better comprehending expert opinions.

\section{Acknowledgments}
This work has been partially supported by the National Science Foundation (NSF) Standard Grant Award \#2452028 and Air Force Office of Scientific Research Grant/Cooperative Agreement Award \#FA9550-23-1-0426. We would also like to thank the University of Central Florida CS Department and AI initiative for their continuous support through Student Fellowships and Graduate Assistantships.

\bibliography{References/Mine,References/mr}

\appendix




\section{Appendix}

\subsection{Background on TELeR Taxonomy}\label{sec:background}

\begin{figure*}[!htb]
  \centering    
  \includegraphics[width=0.9\textwidth]{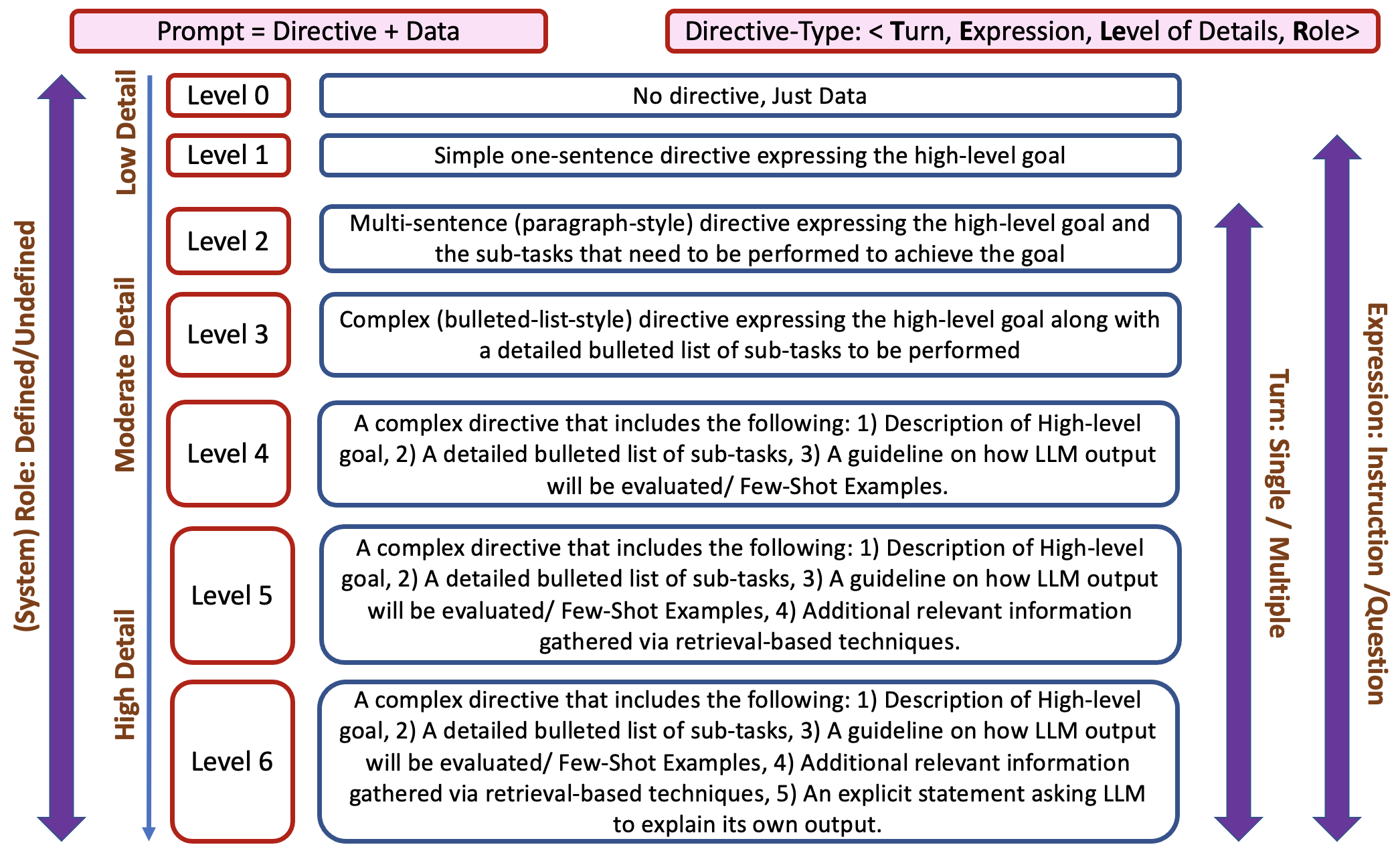}
  \caption{TELeR Taxonomy for prompting LLMs to perform complex tasks. For details, see \citet{santu2023teler}.}
  \label{TELeR_Fig}
  \vspace{-2mm}
\end{figure*}

As shown in Figure~\ref{TELeR_Fig}, the taxonomy introduced by \citet{santu2023teler} categorizes complex task prompts based on the following four criteria. 

\begin{enumerate}[leftmargin=*,itemsep=0.2ex,partopsep=0.2ex,parsep=0ex]
    
    \item \textbf{Turn}: This refers to the number of turns or shots used while prompting an LLM to accomplish a complex task. In general, prompts can be classified as either single or multi-turn.
    
    \item \textbf{Expression}: This refers to the style of expression for interacting with the LLM, such as questioning or instructing. 
    
    \item \textbf{Level of Details}: This dimension of prompt style deals with the granularity or depth of question or instruction. Prompts with higher levels of detail provide more granular instructions. For example, Level 0 has no directive, i.e., questions or instructions. On the other hand, Level 4 prompts have to specify the particular sub-tasks, few-shot examples, and/or the basis for evaluating the output generated by the LLM.
    
    \item \textbf{Role}: Some LLMs provide users with the option of specifying the system's role. The response of LLM can vary due to changes in role definitions, although the prompt content remains unchanged.
\end{enumerate}

\subsection{Detailed Plots for Five Criteria}

These figures (Figures~\ref{fig:cs-pr-p1234},~\ref{fig:cw-pr-p1234},~\ref{fig:csu-pr-p1234} and ~\ref{fig:mr-pr-p1234}) depict the comparative performances of three LLMs in terms of their (human-perceived) precision/recall distributions along four criteria: Common Strengths, Common Weaknesses, Common Suggestions, and Literature Review. Here, human-perceived precision and recall distributions are derived from their qualitative rating counts, i.e., \{SA=Strongly Agree, A=Agree, N=Neutral, D=Disagree, SD=Strongly Disagree\}, followed by a normalization term to convert them into a valid distribution (individual values ranging between [0-1]). Refer to Figures~\ref{fig:cs-pr-p1234},~\ref{fig:cw-pr-p1234},~\ref{fig:csu-pr-p1234} and ~\ref{fig:mr-pr-p1234} for these detailed results.

\begin{figure*}[!htb]
        \begin{subfigure}{0.5\textwidth}
            \includegraphics[width=0.95\textwidth]{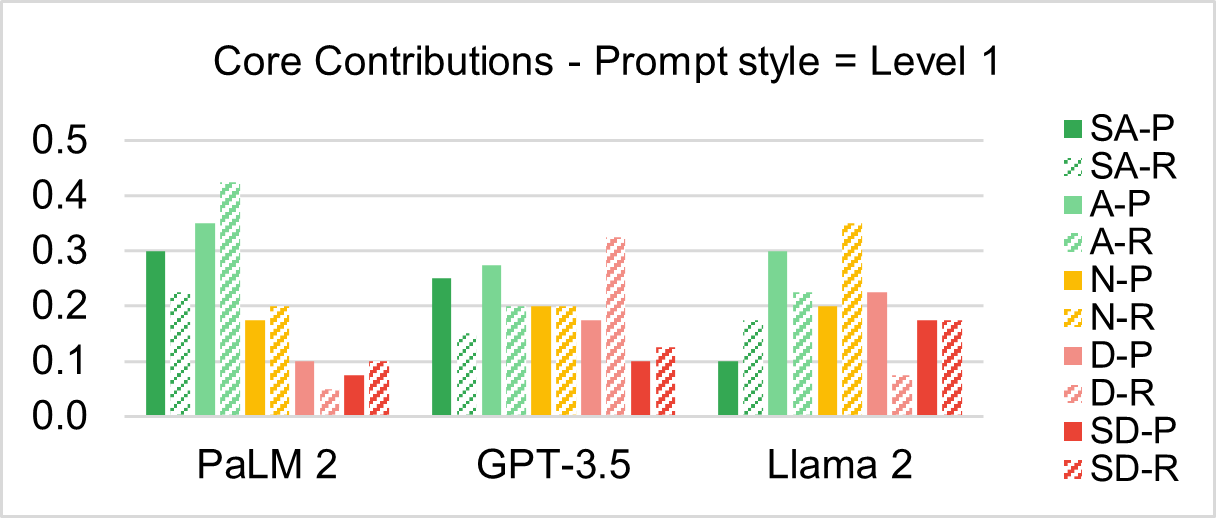}
            \centering            \captionsetup{justification=centering,margin=0cm}
            \caption{Core Contributions - Prompt Level 1} 
            \label{subfig:cc-pr-p1}    
        \end{subfigure}
        \begin{subfigure}{0.5\textwidth}
            \includegraphics[width=0.95\textwidth]{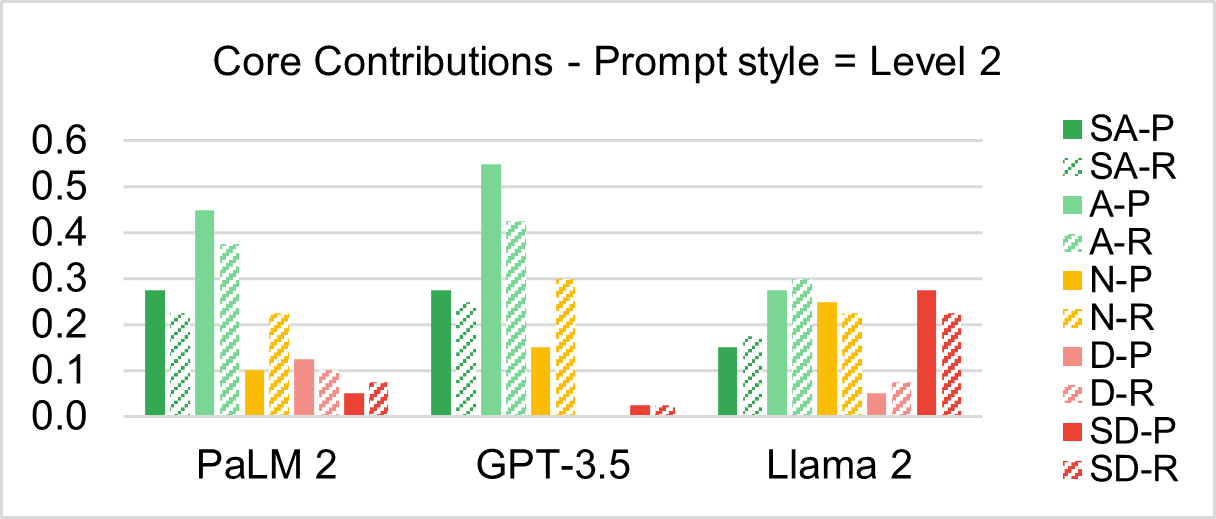}
            \centering            \captionsetup{justification=centering,margin=0cm}
            \caption{Core Contributions - Prompt TELeR Level 2}
            \label{subfig:cc-pr-p2} 
        \end{subfigure}
        ~
        \begin{subfigure}{0.5\textwidth}
            \includegraphics[width=0.95\textwidth]{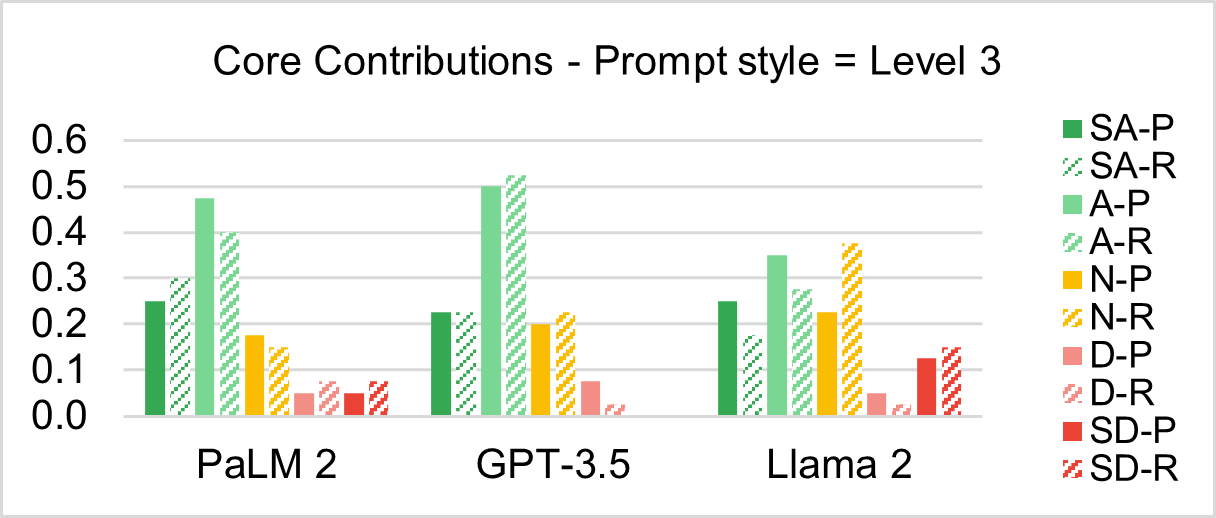}
            \centering            \captionsetup{justification=centering,margin=0cm}
            \caption{Core Contributions - Prompt TELeR Level 3}
            \label{subfig:cc-pr-p3} 
        \end{subfigure}
        \begin{subfigure}{0.5\textwidth}
            \includegraphics[width=0.95\textwidth]{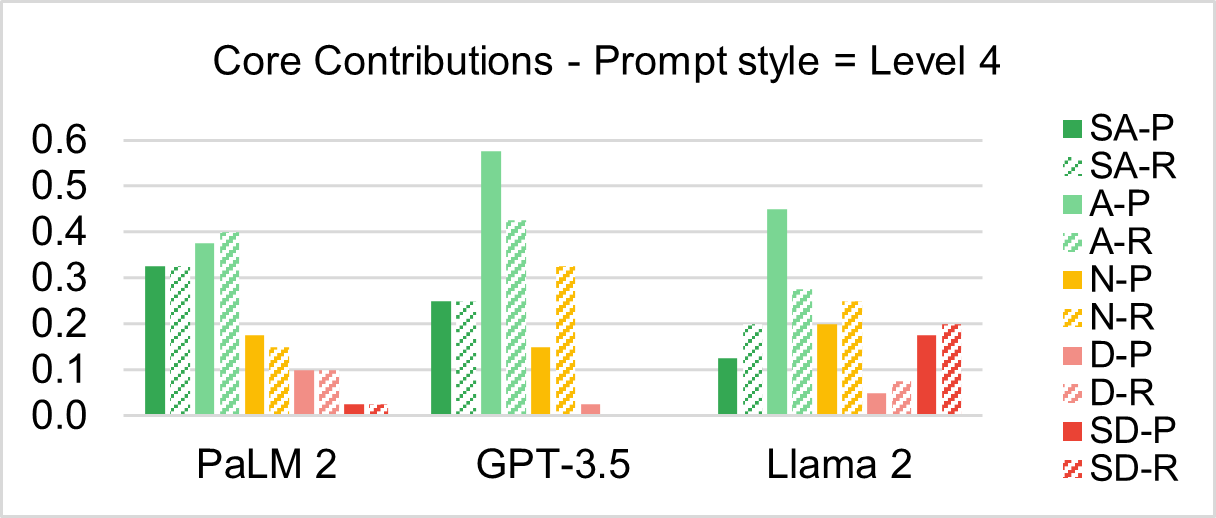}
            \centering            \captionsetup{justification=centering,margin=0cm}
            \caption{Core Contributions - Prompt TELeR Level 4}
            \label{subfig:cc-pr-p4}
        \end{subfigure}
        \caption{Core Contributions Ratings - Prompt TELeR Level 1-4.}
        \label{fig:cc-pr}
    \end{figure*}

\begin{figure*}[!htb]

        \begin{subfigure}{0.5\textwidth}
            \includegraphics[width=0.95\textwidth]{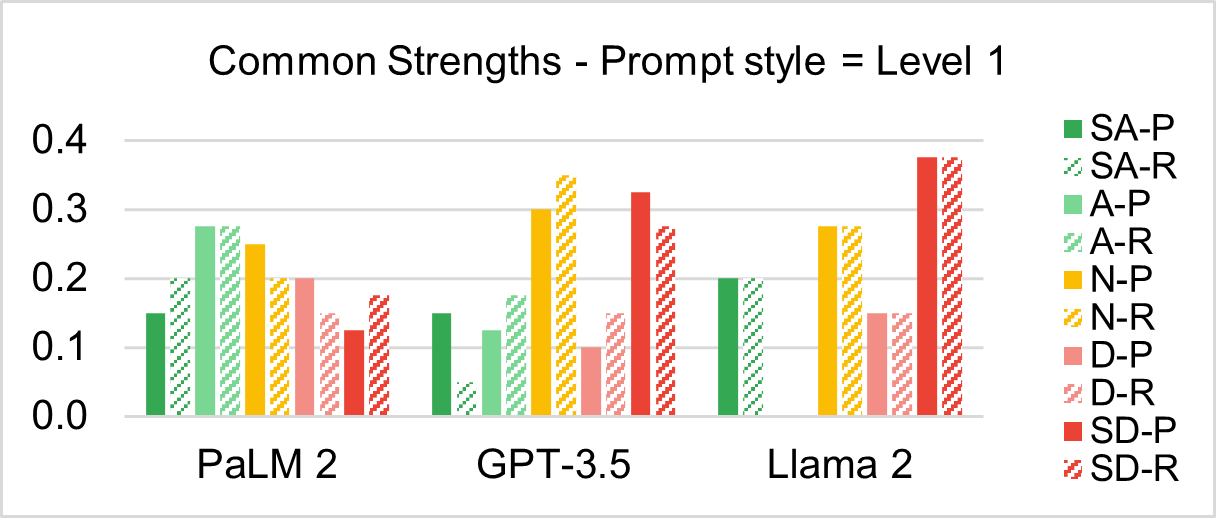}
            \centering            \captionsetup{justification=centering,margin=0cm}
            \caption{Common Strengths - Prompt TELeR Level 1}
            \label{fig:cs-pr-p1}            
        \end{subfigure}
        \begin{subfigure}{0.5\textwidth}
            \includegraphics[width=0.95\textwidth]{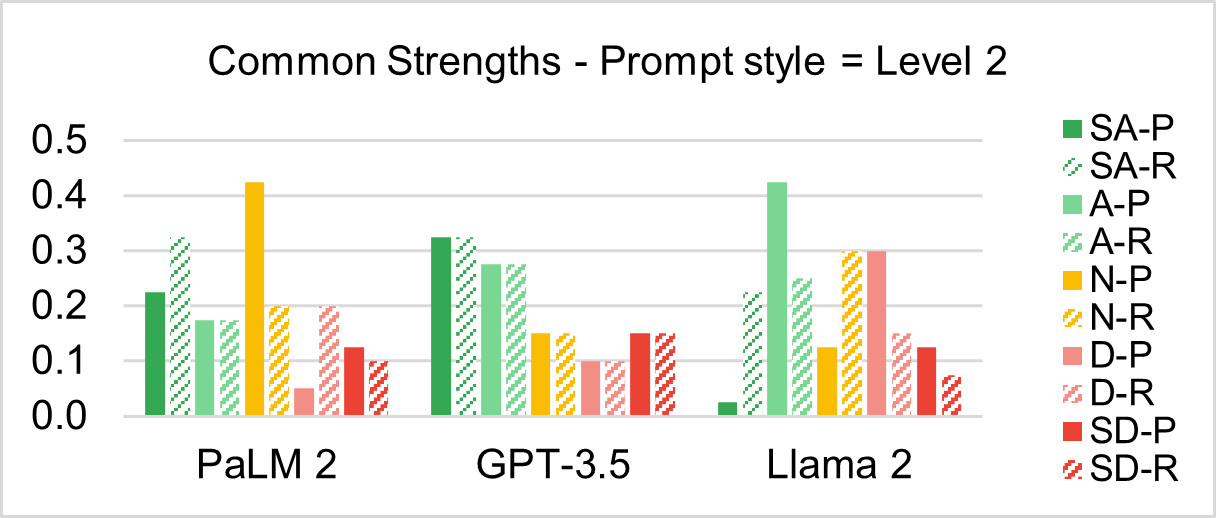}
            \centering            \captionsetup{justification=centering,margin=0cm}
            \caption{Common Strengths - Prompt TELeR Level 2}
            \label{fig:cs-pr-p2}            
        \end{subfigure}  
        ~
        \begin{subfigure}{0.5\textwidth}
            \includegraphics[width=0.95\textwidth]{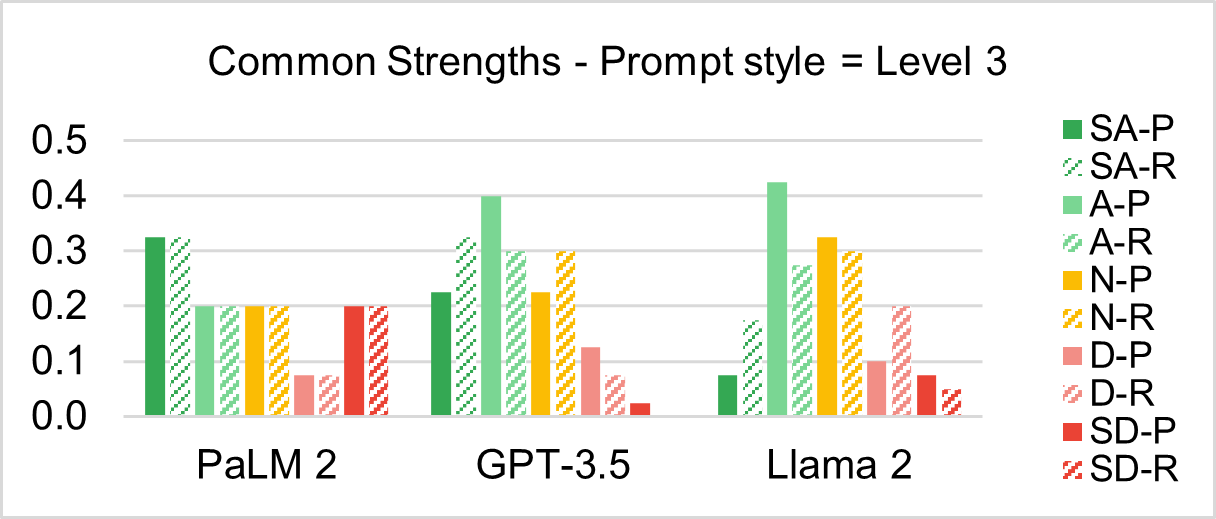}
            \centering            \captionsetup{justification=centering,margin=0cm}
            \caption{Common Strengths - Prompt TELeR Level 3}
            \label{fig:cs-pr-p3}            
        \end{subfigure}
        \begin{subfigure}{0.5\textwidth}
            \includegraphics[width=0.95\textwidth]{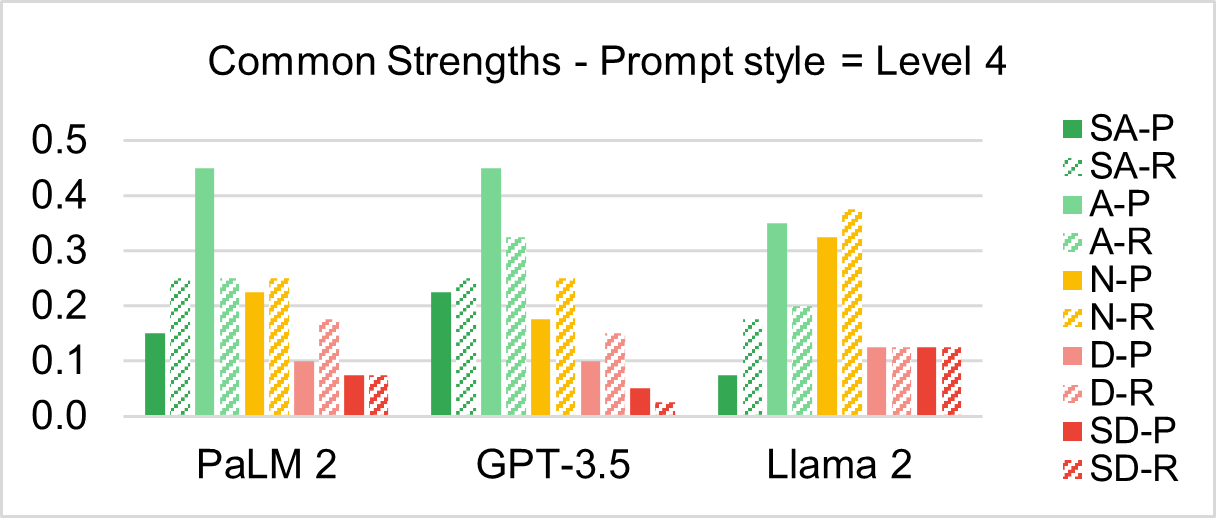}
            \centering            \captionsetup{justification=centering,margin=0cm}
            \caption{Common Strengths - Prompt TELeR Level 4}
            \label{fig:cs-pr-p4}            
        \end{subfigure}
        \caption{Common Strengths Ratings - Prompt TELeR Level 1-4.}
        \label{fig:cs-pr-p1234}
    \end{figure*}

    \begin{figure*}[!htb]
        \begin{subfigure}{0.5\textwidth}
            \includegraphics[width=0.95\textwidth]{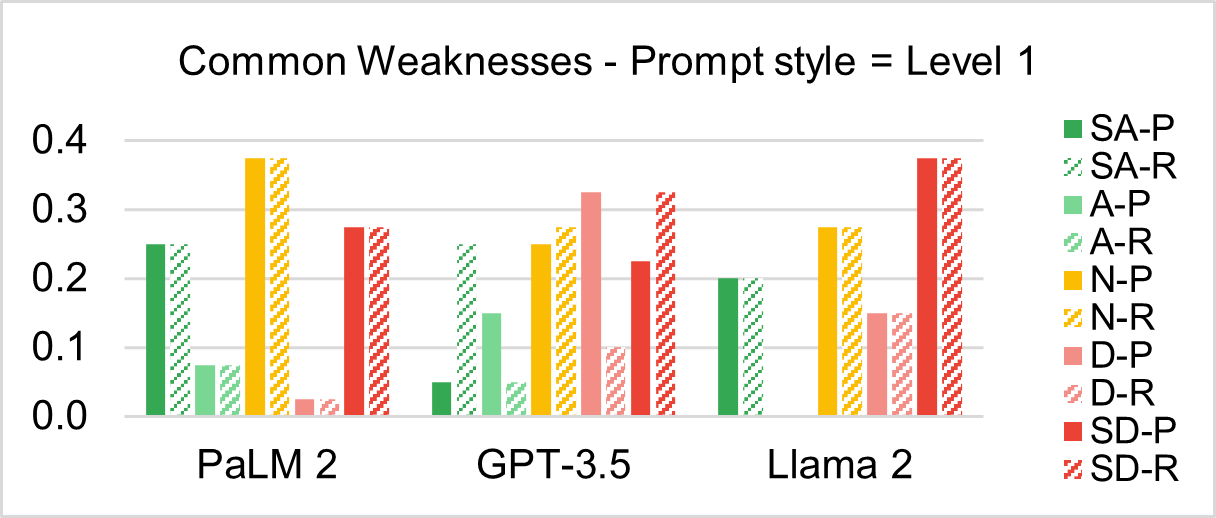}            \captionsetup{justification=centering,margin=0cm}
            \caption{Common Weaknesses - Prompt Level 1}
            \label{fig:cw-pr-p1}            
        \end{subfigure} 
        \begin{subfigure}{0.5\textwidth}
            \includegraphics[width=0.95\textwidth]{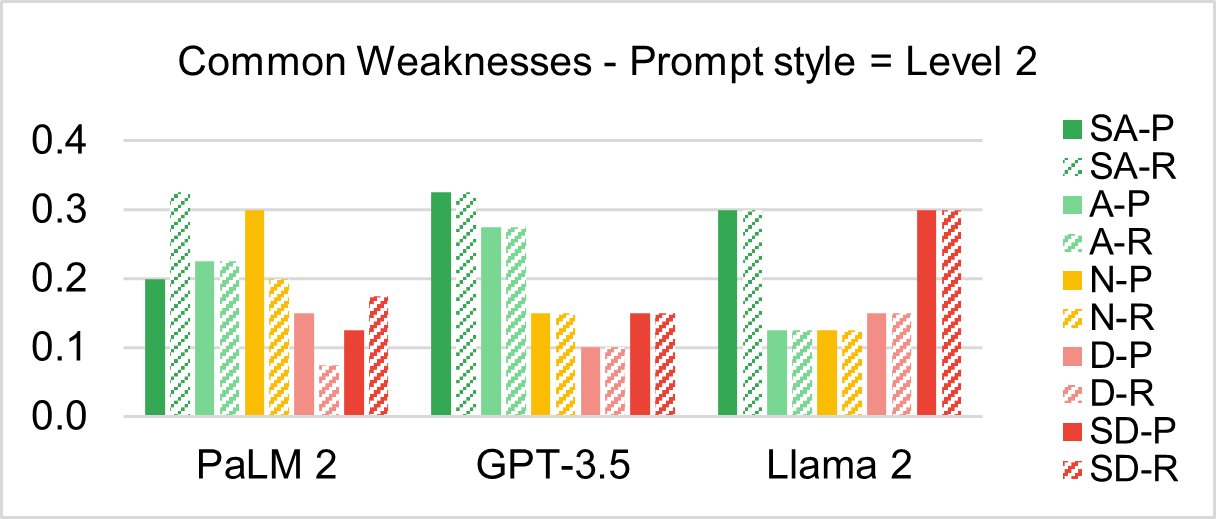}
            \centering            \captionsetup{justification=centering,margin=0cm}
            \caption{Common Weaknesses - Prompt Level 2}
            \label{fig:cw-pr-p2}            
        \end{subfigure}  
        ~
        \begin{subfigure}{0.5\textwidth}
            \includegraphics[width=0.95\textwidth]{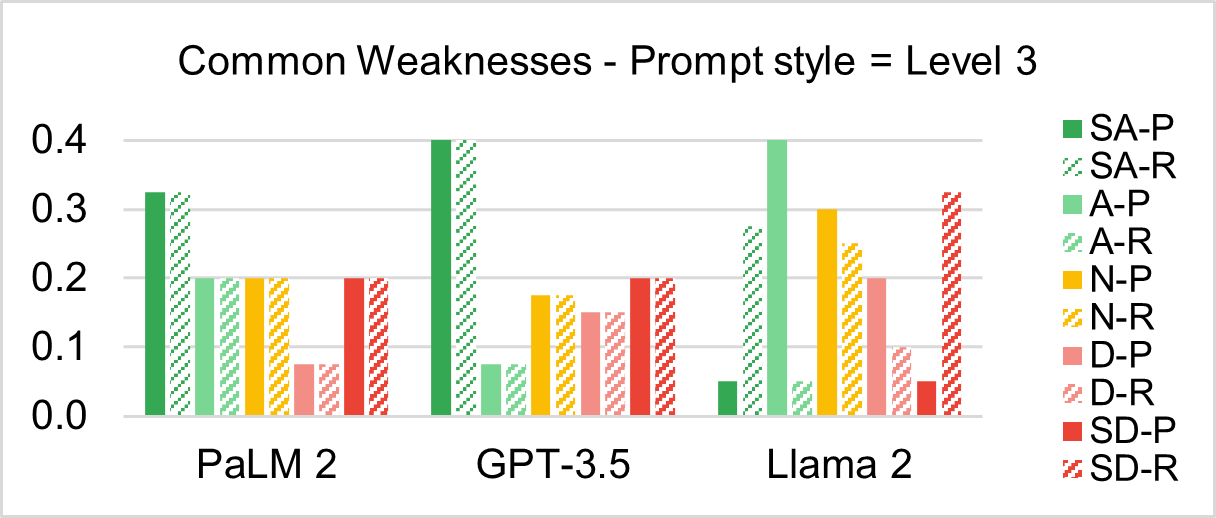}
            \centering            \captionsetup{justification=centering,margin=0cm}
            \caption{Common Weaknesses - Prompt Level 3}
            \label{fig:cw-pr-p3}            
        \end{subfigure} 
        \begin{subfigure}{0.5\textwidth}
            \includegraphics[width=0.95\textwidth]{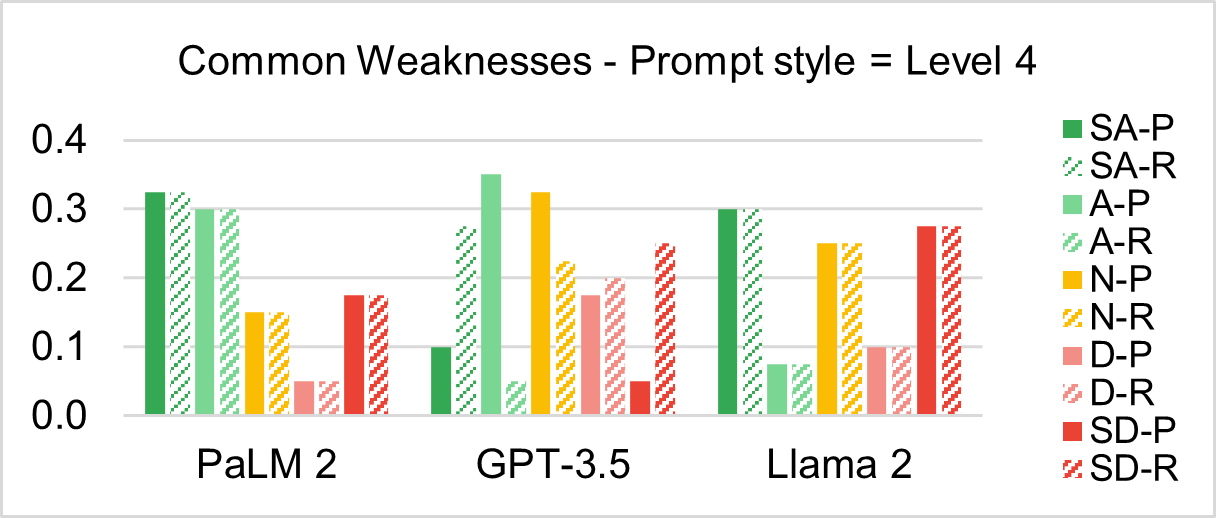}
            \centering            \captionsetup{justification=centering,margin=0cm}
            \caption{Common Weaknesses - Prompt Level 4}
            \label{fig:cw-pr-p4}            
        \end{subfigure}
        \caption{Core Weaknesses Ratings - Prompt TELeR Level 1-4.}
        \label{fig:cw-pr-p1234}
    \end{figure*}

    \begin{figure*}
        \begin{subfigure}{0.5\textwidth}
            \includegraphics[width=0.95\textwidth]{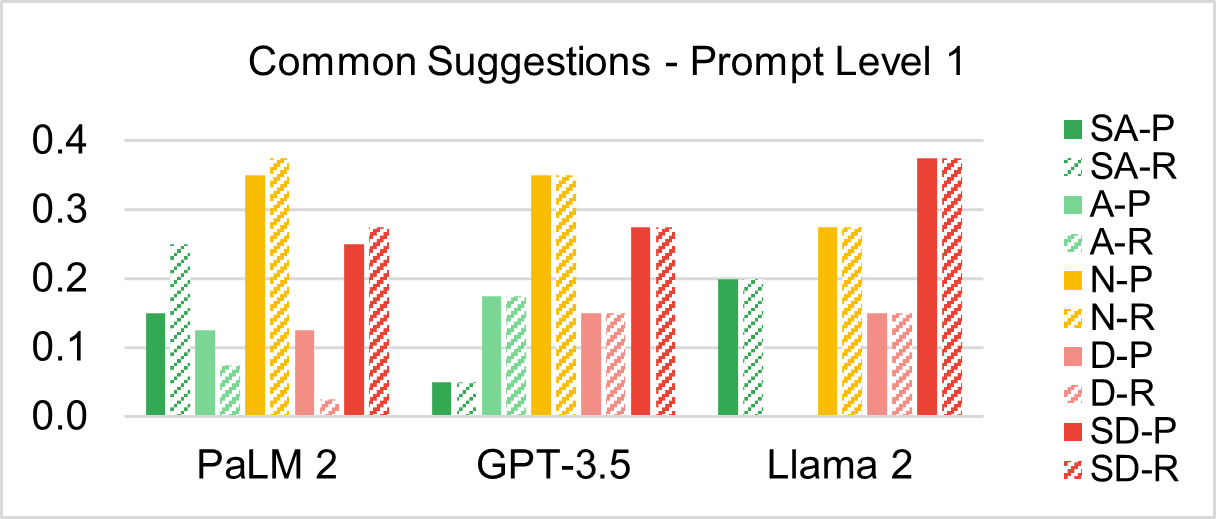}
            \centering            \captionsetup{justification=centering,margin=0cm}
            \caption{Common Suggestions - Prompt Level 1}
            \label{fig:csu-pr-p1}            
        \end{subfigure}
        \begin{subfigure}{0.5\textwidth}
            \includegraphics[width=0.95\textwidth]{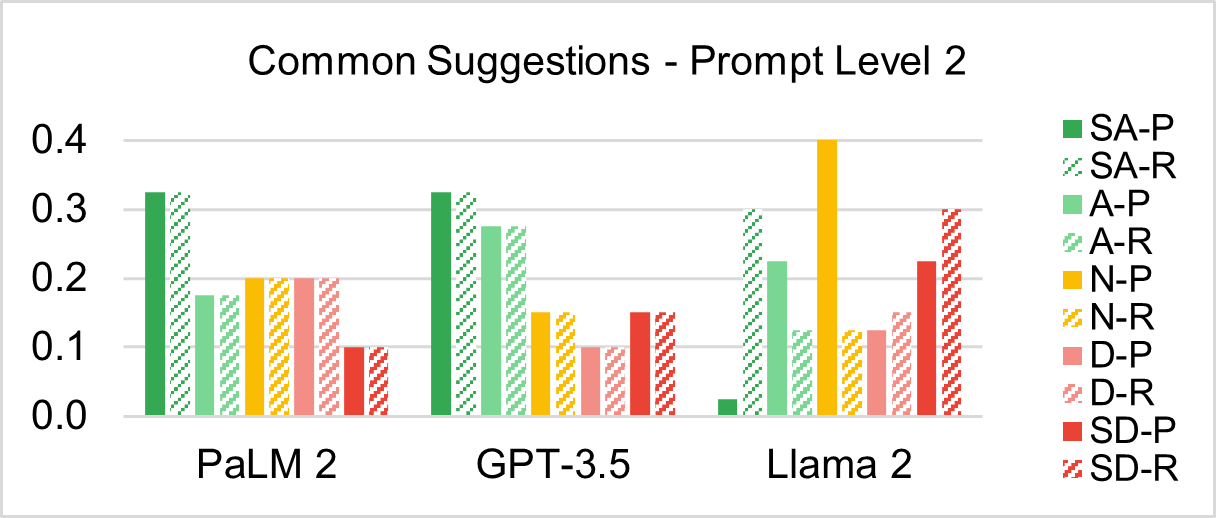}
            \centering            \captionsetup{justification=centering,margin=0cm}
            \caption{Common Suggestions - Prompt Level 2}
            \label{fig:csu-pr-p2}            
        \end{subfigure} 
        ~
        \begin{subfigure}{0.5\textwidth}
            \includegraphics[width=0.95\textwidth]{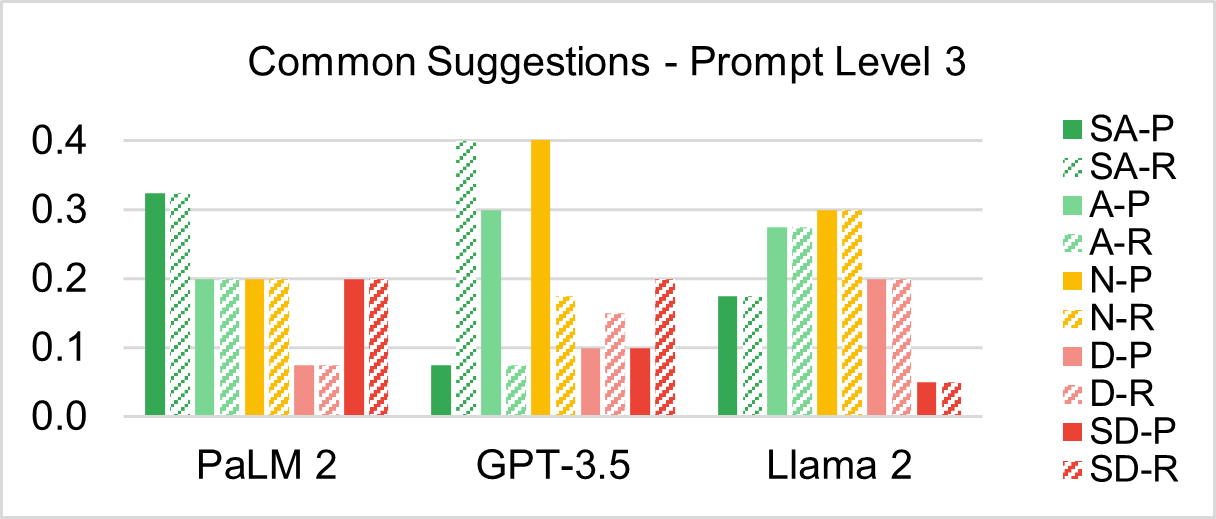}
            \centering            \captionsetup{justification=centering,margin=0cm}
            \caption{Common Suggestions - Prompt Level 3}
            \label{fig:csu-pr-p3}            
        \end{subfigure}
        \begin{subfigure}{0.5\textwidth}
            \includegraphics[width=0.95\textwidth]{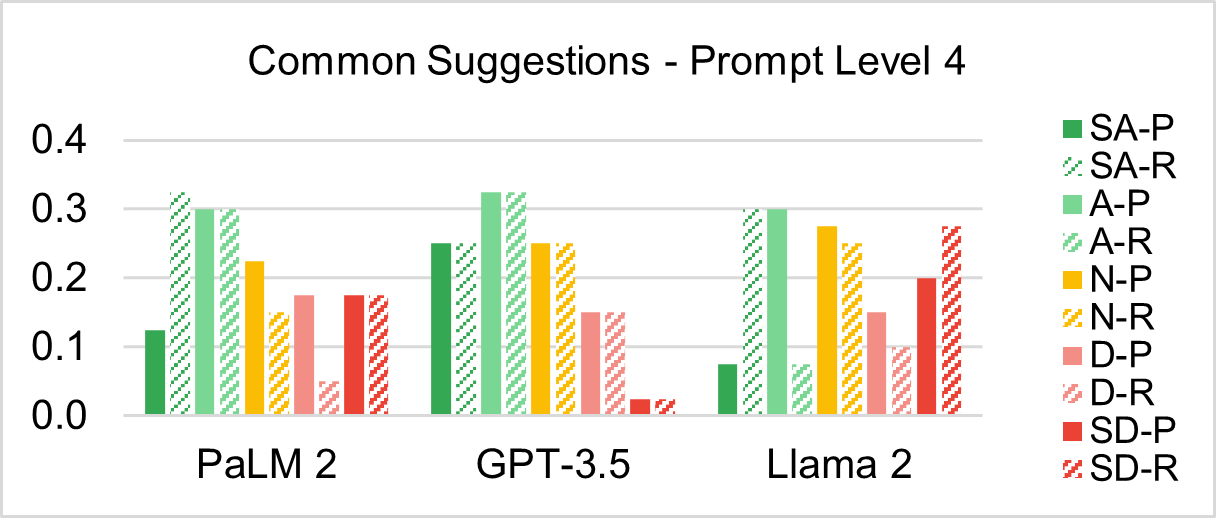}
            \centering            \captionsetup{justification=centering,margin=0cm}
            \caption{Common Suggestions - Prompt Level 4}
            \label{fig:csu-pr-p4}            
        \end{subfigure}
        \caption{Core Suggestions Ratings - Prompt TELeR Level 1-4.}
        \label{fig:csu-pr-p1234}
    \end{figure*}

    \begin{figure*}
        \begin{subfigure}{0.5\textwidth}
            \includegraphics[width=0.95\textwidth]{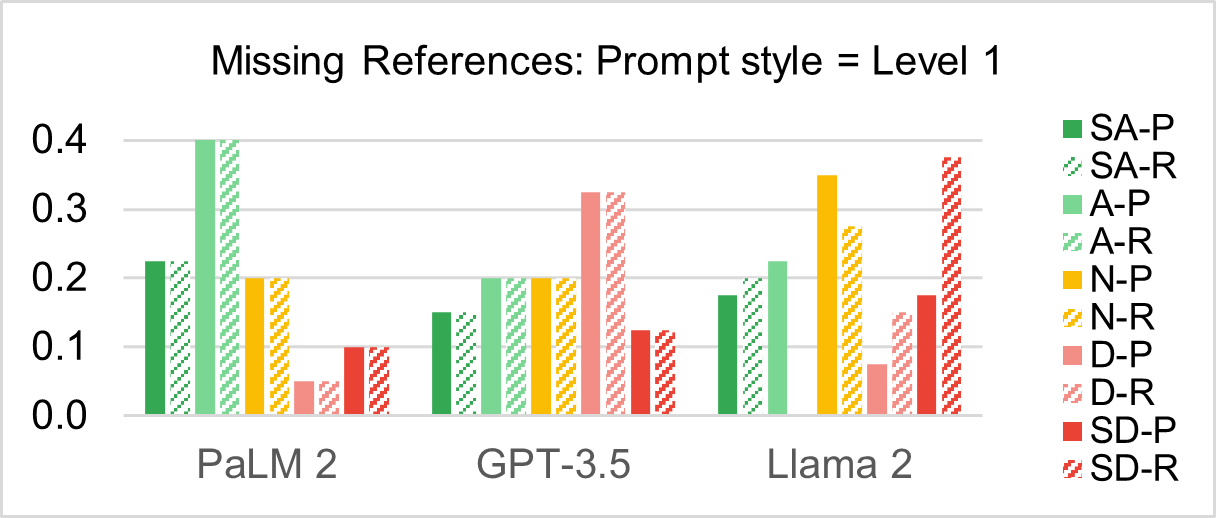}
            \centering            \captionsetup{justification=centering,margin=0cm}
            \caption{Literature Review - Prompt Level 1}
            \label{fig:mr-pr-p1}            
        \end{subfigure}
        \begin{subfigure}{0.5\textwidth}
            \includegraphics[width=0.95\textwidth]{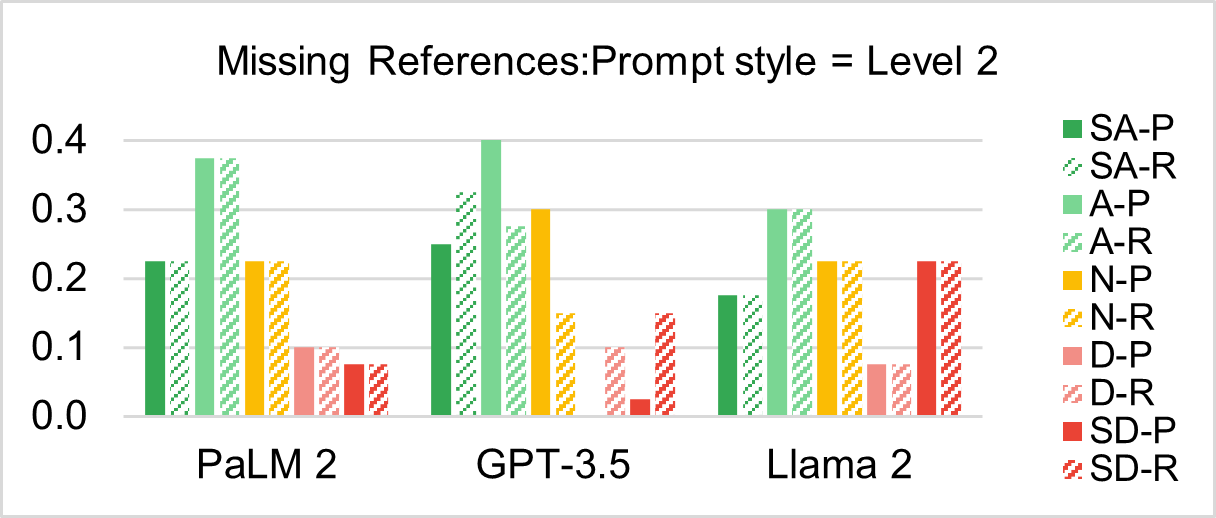}
            \centering            \captionsetup{justification=centering,margin=0cm}
            \caption{Literature Review  - Prompt Level 2}
            \label{fig:mr-pr-p2}            
        \end{subfigure}
        ~
        \begin{subfigure}{0.5\textwidth}
            \includegraphics[width=0.95\textwidth]{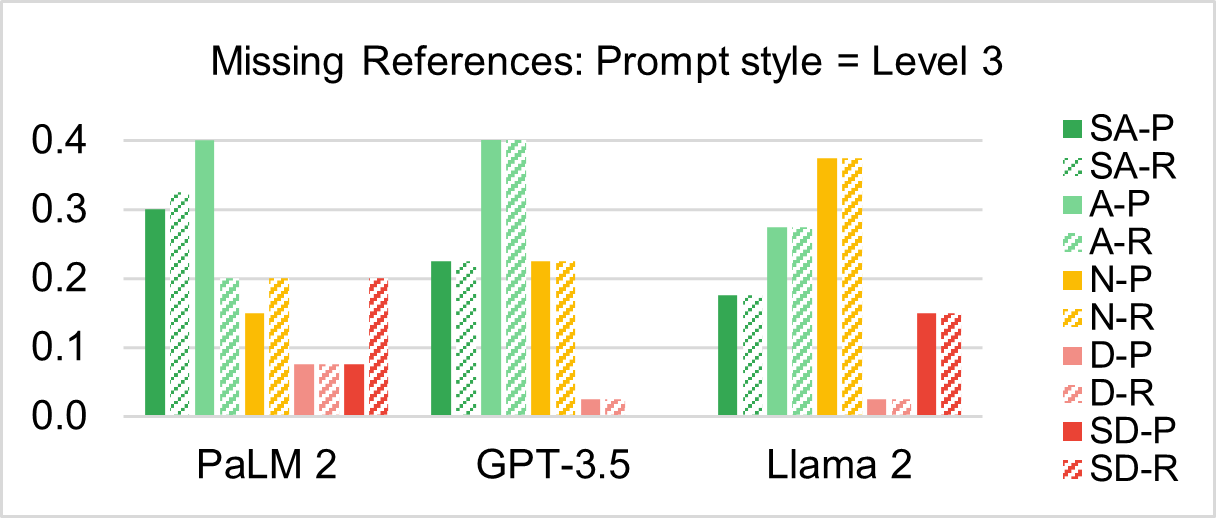}
            \centering            \captionsetup{justification=centering,margin=0cm}
            \caption{Literature Review - Prompt Level 3}
            \label{fig:mr-pr-p3}            
        \end{subfigure}
        \begin{subfigure}{0.5\textwidth}
            \includegraphics[width=0.95\textwidth]{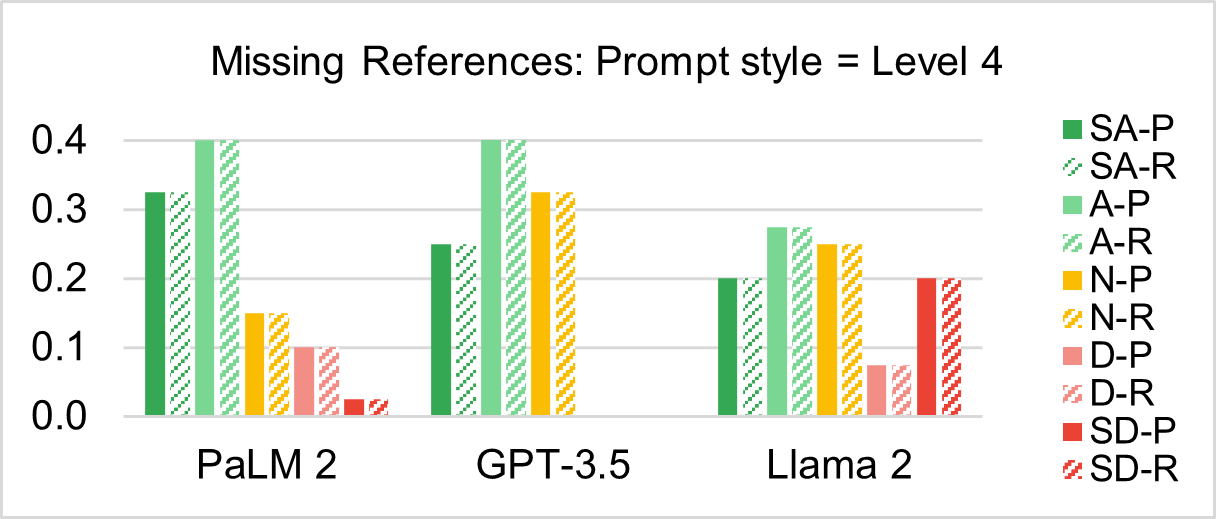}
            \centering            \captionsetup{justification=centering,margin=0cm}
            \caption{Literature Review - Prompt Level 4}
            \label{fig:mr-pr-p4}            
        \end{subfigure}
        \caption{Literature Review Ratings - Prompt TELeR Level 1-4.}
        \label{fig:mr-pr-p1234}
    \end{figure*}

\subsection{Prompt Design Details}
\label{app:prompt_design}
Prompt designs during our preliminary study and final case study are shown in Table~\ref{tbl_prompt_design}. Also, the prompt design for automatic evaluation is presented in Table~\ref{tbl_micro_eval_prompt} and Table~\ref{tbl_macro_eval_prompt}.

\begin{table*}[!htb]

\begin{tabular}{|p{0.7cm}|p{6.1cm}|p{8.0cm}|} 
 \hline
 Level & Prompt (Preliminary Study) & Prompt (Case Study)\\
 \hline

 \centering \textcolor{blue}{1} & \small{\textcolor{blue}{messages=[{“role": “system", “content": “As a meta-reviewer, answer the following: What would be a reasonable meta-review considering the given reviews?"},
{“role”: “user”, “content”: 1.$<R1>$ 2.$<R2>$ 3. $<R3>$}]}
} &
\small{\textcolor{blue}{Using 3 reviews given below draft a meta review:}

\textcolor{blue}{1.Review\#1{\textbackslash}n}\textcolor{blue}{$<R1>${\textbackslash}n}

\textcolor{blue}{2.Review\#2{\textbackslash}n}\textcolor{blue}{$<R2>${\textbackslash}n}

\textcolor{blue}{3.Review\#3{\textbackslash}n}\textcolor{blue}{$<R3>$}}
\\
 \hline
 \centering \textcolor{blue}{2} & \small{\textcolor{blue}{messages=[{“role": “system", “content": “You are a meta-review assistant. You should create a meta review by answering the following questions:"
    “(a) According to reviews, what are the core contributions?"
    “(b) What are the strengths as mentioned in the reviews?"
    “(c) What are the weaknesses as mentioned in the reviews?"
    “(d) What suggestions would you provide for improvement?"
    “(e) What are the missing references as described in the reviews?"}, 
{“role”: “user”, “content”: 1.$<R1>$ 2.$<R2>$ 3. $<R3>$}]}} & \small{\textcolor{blue}{Using the three reviews, generate a meta review by incorporating core contributions, common strengths, common  weaknesses, common suggestions for improvement, and missing references. Common strengths, Common weaknesses, and Common suggestions are strengths, weaknesses, and suggestion for improvement respectively that are common in at least 2 reviews.}
\textcolor{blue}{
Three reviews are as follows:{\textbackslash}n}

\textcolor{blue}{1.Review\#1{\textbackslash}n}\textcolor{blue}{$<R1>${\textbackslash}n}

\textcolor{blue}{2.Review\#2{\textbackslash}n}\textcolor{blue}{$<R2>${\textbackslash}n}

\textcolor{blue}{3.Review\#3{\textbackslash}n}\textcolor{blue}{$<R3>$}}\\
 \hline 
 \centering \textcolor{blue}{3} & \small{\textcolor{blue}{messages=[{“role": “system", “content": “As a meta-reviewer, draft a meta review as per the following directions:"
    “(a) State core contributions."
    “(b) Mention strengths."
    “(c) Mention weaknesses."
    “(d) Write suggestions for improvement."
    “(e) State missing references."},
{“role”: “user”, “content”: 1.$<R1>$ 2.$<R2>$ 3. $<R3>$}]}} & \small{\textcolor{blue}{You are a meta-review assistant. Using three reviews given as Review\#1, Review\#2, and Review\#3, give me a meta-review by answering:}}
\small{\textcolor{blue}{\begin{enumerate}[label=(\alph*)]
\item Mention core contributions with common contributions first.
\item A common strength is a strength that is mentioned in at least 2 reviews as a strength. Mention common strengths.
\item A common weakness is a weakness that is mentioned in at least 2 reviews as a weakness. Mention common weaknesses.
\item A common suggestion for improvement is a suggestion that is mentioned in at least 2 reviews as a suggestion for improvement. Mention common improvements suggested.
\item State whether reviews refer to missing references or not. A listing of missing references is not required.
\end{enumerate}}}

\textcolor{blue}{Three reviews are as follows:{\textbackslash}n}

\textcolor{blue}{1.Review\#1{\textbackslash}n}\textcolor{blue}{$<R1>${\textbackslash}n}

\textcolor{blue}{2.Review\#2{\textbackslash}n}\textcolor{blue}{$<R2>${\textbackslash}n}

\textcolor{blue}{3.Review\#3{\textbackslash}n}\textcolor{blue}{$<R3>$}\\

 \hline
 \centering \textcolor{blue}{4} & \small{\textcolor{blue}{messages=[
{“role": “system", “content": “As a meta-reviewer, draft a meta review by answering the following bullet points: "}}
\textcolor{blue}{
“- What is the summary of core contributions? Provide answer with supporting evidences."}

\textcolor{blue}{
“- Which common strengths are referred to in the reviews? Support your answer with explanations."}

\textcolor{blue}{
“- What common weaknesses are described in the reviews? Give evidences in support of the reply."}

\textcolor{blue}{
“- What suggestions for improvement have been provided by three reviews? Explain the basis for the answer."}

\textcolor{blue}{{
“- Which missing references are mentioned in the reviews? Answer with explanations will be desirable."},} 

\textcolor{blue}{
{”role”: ”user”, ”content”: 1.$<R1>$ 2.$<R2>$ 3. $<R3>$}]}} & \small{\textcolor{blue}{Using three reviews given as Review\#1, Review\#2, and Review\#3, as a meta-reviewer, your task is to draft a meta review by answering the following bulleted questions:}} 

\textcolor{blue}{- What is the summary of core contributions? Provide answer with supporting evidence.}

\textcolor{blue}{- Which common strengths are referred to in the reviews? A common strength is a strength that is mentioned in at least 2 reviews as a strength. Support your answer with explanation.}

\textcolor{blue}{- What common weaknesses are described in the reviews? A common weakness is a weakness that is mentioned in at least 2 reviews as a weakness.  Give evidence in support of the reply.}

\textcolor{blue}{- What suggestions for improvement have been provided by three reviews? A common suggestion for improvement is a suggestion that is mentioned in at least 2 reviews as a suggestion for improvement. Explain the basis for the answer.}

\textcolor{blue}{- Do the reviews mention about missing references? Answer with explanation is desirable but listing of missing references is not required.}

\textcolor{blue}{Reviews are as below:{\textbackslash}n}

\textcolor{blue}{1.Review\#1{\textbackslash}n}\textcolor{blue}{$<R1>${\textbackslash}n}

\textcolor{blue}{2.Review\#2{\textbackslash}n}\textcolor{blue}{$<R2>${\textbackslash}n}

\textcolor{blue}{3.Review\#3{\textbackslash}n}\textcolor{blue}{$<R3>$}\\
\hline
\centering  & \multicolumn{2}{|c|}{\small{\textit{\textcolor{blue}{In all the above prompts, R1, R2, and R3 are reviews provided by the reviewers.}}}} 
  \\
 \hline
\end{tabular}

\caption{\label{tbl_prompt_design}Prompt Design for different levels}

\end{table*}

\begin{table*}[!htb]

\begin{tabular}{|p{16cm}|} 
 \hline
  Evaluation Prompt  \\
 \hline

\small{
Imagine you are a human annotator evaluating the quality of meta-reviews written for a conference. You will evaluate five aspects of the meta-review by giving a value from 1 to 5, with $1$ being the worst and $5$ being the best, without providing any additional explanation. Please follow these steps:

1. Carefully read the reviews, noting the key information they contain.

2. Read the meta-review thoroughly.

3. Rate the meta-review on the following five aspects using Precision (P) and Recall (R): 
\begin{itemize}
    \item Core Contributions (CC)
    \item Common Strengths (CS)
    \item Common Weaknesses (CW)
    \item Suggestions for Improvement (SI)
    \item Missing References (MR)
\end{itemize}

4. Provide your ratings on a scale from 1 (worst) to 5 (best) for each aspect in both Precision and Recall.

Definitions are as follows:
\begin{itemize}
    \item Core Contributions (CC): Does the meta-review contain the contributions of the paper mentioned in multiple reviews?
    \item Common Strengths (CS): Does the meta-review include all the strengths that are common across reviews?
    \item Common Weaknesses (CW): Does the meta-review include all the weaknesses that are common in multiple reviews?
    \item Suggestions for Improvement (SI): Does the meta-review include all the common suggestions for improvement mentioned in multiple reviews?
    \item Missing References (MR): Does the meta-review include all the missing references mentioned in multiple reviews?
\end{itemize}

Specific Questions for Rating:

\begin{itemize}
    \item CC-P: Is the meta-review precise in capturing the contributions of the paper highlighted by at least two reviews?
    \item CC-R: Has the meta-review covered all the contributions of the paper, which are highlighted by at least two reviews?
    \item CS-P: Is the meta-review precise in capturing the common strengths highlighted by at least two reviews?
    \item CS-R: Has the meta-review covered all the common strengths highlighted by at least two reviews?
    \item CW-P: Is the meta-review precise in capturing the common weaknesses highlighted by at least two reviews?
    \item CW-R: Has the meta-review covered all the common weaknesses highlighted by at least two reviews?
    \item SI-P: Is the meta-review precise in capturing the suggestions highlighted by at least two reviews?
    \item SI-R: Has the meta-review covered all the suggestions highlighted by at least two reviews?
    \item MR-P: Is the meta-review precise in capturing the missing references highlighted by at least two reviews?
    \item MR-R: Has the meta-review covered all the missing references highlighted by at least two reviews?
\end{itemize}

Now complete the task for the given reviews and meta-review.

Input: 

{1.Review\#1{\textbackslash}n}$<R1>${\textbackslash}n

{2.Review\#2{\textbackslash}n}$<R2>${\textbackslash}n

{3.Review\#3{\textbackslash}n}{$<R3>$}

Output: 

Provide your evaluated scores in a list:
}               
\\
 \hline
\end{tabular}

\caption{\label{tbl_micro_eval_prompt}Prompt Design for automatic Micro evaluation}

\end{table*}

\begin{table*}[!htb]

\begin{tabular}{|p{16cm}|} 
 \hline
Evaluation Prompt  \\
 \hline
 \small{
Imagine you are a human annotator tasked with evaluating a model's performance in generating a meta-review based on three individual reviews. You are provided with the following:
\begin{itemize}
    \item The **prompt** that guided the generation of the meta-review.
    \item The **three individual reviews**, which the model used to generate the meta-review.
    \item The **generated meta-review**  created by the model.
    \item The **actual expert-written meta-review**  that serves as a reference for what the meta-review should ideally look like.
\end{itemize}

Please carefully examine this information and evaluate the model's performance based on three criteria. Provide a score on a Likert scale from 1 (poor) to 5 (excellent) for each aspect:

\begin{enumerate}
    \item **Adherence to instructions**: How well did the model follow the specific instructions given in the prompt? Consider how accurately the model addressed the tasks or questions posed in the prompt when generating the meta-review.
          - Score: [1-5]
    \item **Ability to create useful Meta-Reviews **: Evaluate the usefulness of the generated meta-review in terms of its practicality and effectiveness for someone preparing a comprehensive meta-review. Consider whether the generated content helps synthesize the individual reviews into a coherent and insightful summary.
          - Score: [1-5]
    \item **Matching against actual expert-written meta-reviews**: Assess the extent to which the model-generated meta-review aligns with the expert-written meta-review. This involves comparing the content, tone, insights, and overall quality.
          - Score: [1-5]
\end{enumerate}

Input:

**Prompt:** 

{Prompt{\textbackslash}n}{$<P>$}{\textbackslash}n

**Reviews:**

{1.Review\#1{\textbackslash}n}$<R1>${\textbackslash}n

{2.Review\#2{\textbackslash}n}$<R2>${\textbackslash}n

{3.Review\#3{\textbackslash}n}{$<R3>$}{\textbackslash}n
        
**Generated Meta-review:**
        
{Generated Meta Review{\textbackslash}n}{$<GMR>$}{\textbackslash}n

**Actual Meta-review:**

{Actual Meta Review{\textbackslash}n}{$<AMR>$}

Output:

Please provide your evaluation in numerical format in a list.
}\\
 \hline
\end{tabular}

\caption{\label{tbl_macro_eval_prompt}Prompt Design for automatic Macro evaluation}

\end{table*}

\subsection{Responsible NLP Checklist}

\begin{itemize}
    \item \textbf{Potential Risks:} This work has no potential risk. It is a purely curiosity-driven academic study.

    \item \textbf{Human Annotator Recruitment:} We recruited eight graduate students and two undergraduates for the annotation task. Undergrad human annotators were paid at the hourly rate of \$15/hour, while graduate student annotators were paid monthly stipends at the university-prescribed rate.

    \item \textbf{Annotation Guideline:} Detail instructions were given to human annotators for Micro-evaluation and Macro-Evaluation. Instructions in these evaluation forms are presented in Figures~\ref{fig:micro-ins},~\ref{fig:micro-form}, and~\ref{fig:macro-form}.

    \begin{figure*}[!htb]
        \centering    
            \includegraphics[width=\linewidth]{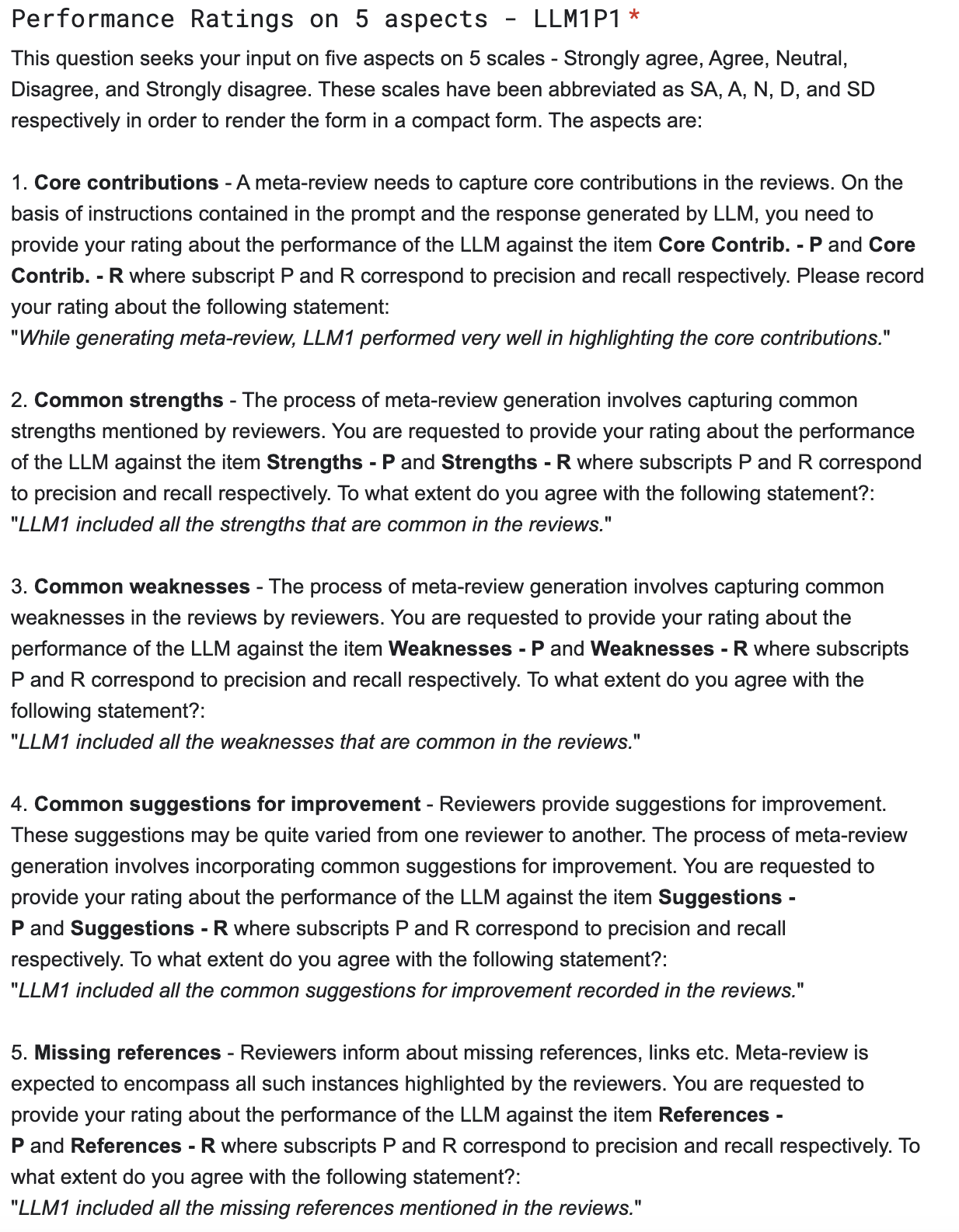}
            \caption{Micro Evaluation Instructions}
            \label{fig:micro-ins}   
            \vspace{-3mm}
    \end{figure*}

    \begin{figure*}[!htb]
            \centering
            \includegraphics[width=\linewidth]{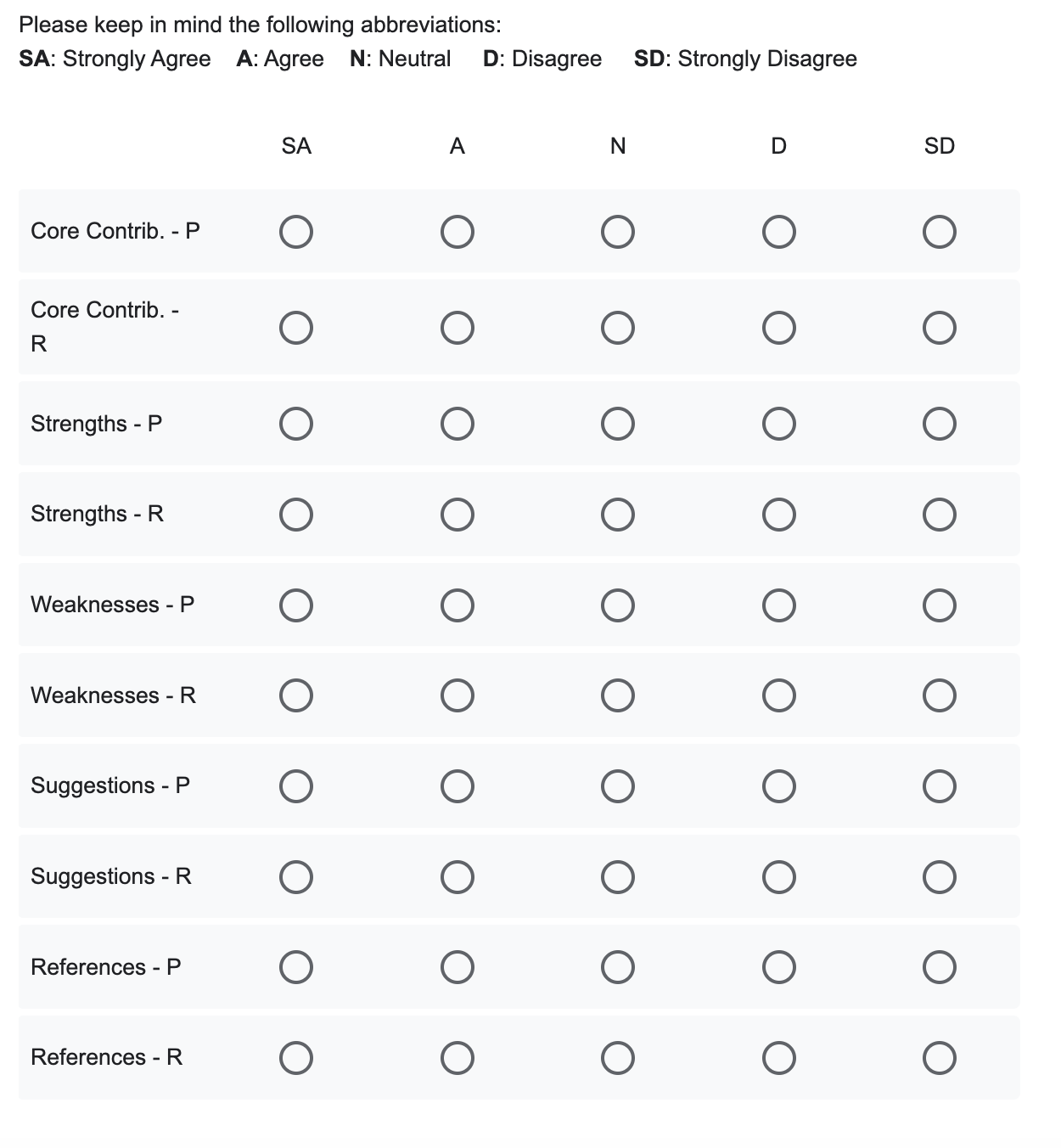}
            \caption{Micro Evaluation Form}
            \label{fig:micro-form}
    \end{figure*}
    
    \begin{figure*}[!htb]
            \centering
         \includegraphics[width=\linewidth]{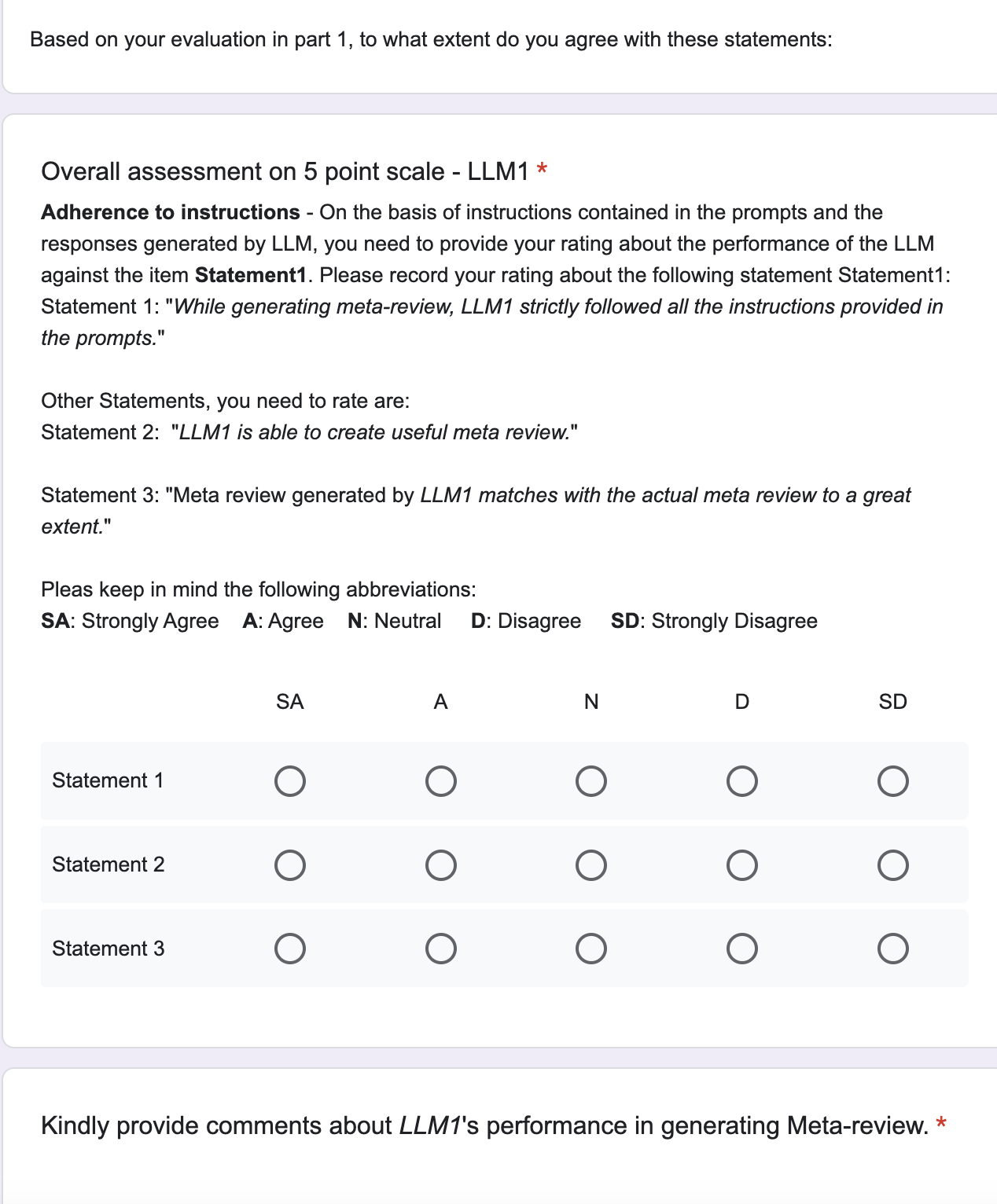}
            \captionsetup{justification=centering,margin=0cm}
            \caption{Macro Evaluation Form}
            \label{fig:macro-form}   
            \vspace{-3mm}
    \end{figure*}

    \item \textbf{Data Privacy and Anonymity Measures:} Any data relating to human subjects (e.g., evaluation survey/interview results) was reviewed to ensure confidentiality and privacy before being preparing the manuscript. We did not collect any user behavior data when annotators performed qualitative evaluations. Finally, we conducted a user survey and reported the summary data from the survey. The survey did not collect any user demographic and personal background information; no identifiable information was stored or reported during/after the lifecycle of this work.

    \item \textbf{Policies and Provisions for Re-use, Re-distribution, and the Production of Derivatives:}  We plan to use one of the Creative Commons copyright licenses to re-use and redistribute our research output. In particular, we will adopt Attribution CC BY – one of the Creative Commons licenses. This license lets others distribute, remix, tweak, and build upon our project as long as users credit the project funded by NSF for the original creation. We decided to use this license because Attribution CC BY is the most accommodating of licenses offered for maximum dissemination.  
\end{itemize}

\newpage
\newpage

\subsection{An Example Case}
Three reviews used for the Preliminary Study are as below:
The full text of these reviews follows -
\subsubsection{Review 1}
{\fontfamily{pcr}\selectfont
Review No.: 1

This is full review.
Review Date: 10/24/2020

Summary of the paper :

The authors propose to improve the sample quality of autoregressive models. The authors propose to (1) - smooth the input data distribution leveraging methods that have shown success in adversarial defense, (2) recover input distribution by learning to reverse the smoothing process. The authors first demonstrate the efficiency of their method on 1d toy-problem, and extend the demonstration to more complex datasets such as MNIST, CIFAR-10 and CelebA with application such as image generation, inpaintting and denoising.

Pros :

The idea to leverage a method previously used for adversarial defense to density estimation is interesting and novel.
The paper is well motivated through the manifold hypothesis approximation (which results in densities with high Lipschitz constants) and compounding errors.
The theory is strong

Cons :

The experiments on denoising and inpainting are only qualitative and suffer from a lack of quantitative evaluation.

Recommendation :

The article is clear, well motivated, and have a strong theoretical grounding. Therefore I would tend to accept the article.

Detailed comments :

The experiment on 2d synthetic datasets (especially the olympic dataset) should be discussed more thoroughly. First, it is not clear that the proposed model is generating better sample than the MADE baseline on this specific dataset. Second, the intersection between rings, in the olympic dataset, seems to be much poorly modeled with the proposed approach compared to the MADE baseline. What is the reason ?

In the section 3.2 the authors are introducing 2 different debiasing methods (either a denoising step or another autoregressive model). In the rest of the article it is not clear which of the two methods the authors are using. In addition, in the 2d toy-problem (i.e. ring and olympic) as the authors are choosing a gaussian smoothing both debiasing methods are usable. Therefore it would be interesting to show both methods and to describe thoroughly the differences (in addition, it might provide an answer to my previous point).

The authors should not mention denoising and inpainting applications if there is no quantitative assessment (at least in appendix)… For the inpainting part, the corrupted input are not even shown (which part of the image has been predicted). The denoising and inpainting experiments sounds like it’s been rushed…

Typos and suggestions to improve the paper :

Minor : Both theorems are provided with nice demonstrations, then the authors should refer to the demonstration in the core text of the article (e.g. see Appendix A).

Minor : Add small arrows in Table 2 to indicate that Inception score is better when lower, and opposite for FID

Typo : page 5, section 3.3, paragraph 2 : relative —> relatively

Figure3 : Right panel : What are the 3 shaded curves ? This should be shown in the legend or at least in the caption
Figure3 : Right panel: In the x-axis it should be specified ‘Variance of q($\tilde{x} \mid x$)

Page 7 : paragraph 1 : 'Thus, it is hard to conclusively determine what is the best way of choosing q(x|$\tilde{x}$).’ —> I think the authors actually give the key to properly choose the noise level (i.e. variance). It seems to depend on the task : if one wants to generate good samples, then the variance has to be set by heuristic. If one needs a good likelihood (e.g. for subsequent downstream tasks) then the variance could be optimized.

Figure 6 : On my understanding, the part ‘denoising’ is redundant with the section image generation. It is interesting to mention the denoising application, but I am not convinced of the utility of the figure 6.

Figure 7 : What is the corrupted input ? Which part of the input has been masked ??}

\subsubsection{Review 2}
Review No.: 2

This is full review.
Review Date: 10/25/2020

{\fontfamily{pcr}\selectfont
Summary. Autoregressive models have demonstrate their potential utility for modeling images and other types of complex data with high flexibility (particularly in density estimation). However, its sampling ability is not that good as explained in the paper. Authors show that one of the main weaknesses of autoregressive models comes from the propagation of mistakes due to the mismatch of conditionals. Inspired in the promising results of randomized smoothing in adversarial models (Cohen et al. 2019), authors propose a similar strategy. The addition of Gaussian noise and posterior modeling of the smoother data makes easier to the autoregressive density to capture the true data distribution. The benefits of this strategy are empirically proved and shown in the experiments.

Strengths. The quality of writing is high and the presentation of the paper facilitates the process of reading. I have to say that I enjoyed while reviewing it. The analysis and description of problems for sampling from autoregressive models is completely understandable to me and I agree with the manifold hypothesis held.

Results with the "sharp" multimodal data looks reliable to me and I believe that the smoother process can also reduce the lipschitz constant as stated in Theorem 1. Until pp. 5, nothing is said about the data denoising process, so one could initially think that there is no way to recover the target density without noise, but authors also did an effort on this. Good point. It is important to remark that the randomized smoothing process can be reverted once learning finishes.

Additionally, I particularly like how authors first present the idea on 1-d examples, later in the experiments, the method is validated with 2-d rings and finally, as stated in the introduction, with different image datasets.

Finally, I did not find any similar work that mixes the idea of smoothing for improving autoregressive modelling.

Weaknesses, Questions \& Recommendations. To me, there are 3 main points of weakness: [W1]. A lack of analysis about the optimal noise for randomized smoothing. [W2]. Why just Gaussian noise, what if data is discrete, could we do this with another type of noise? [W3]. Comments about denoising are included a bit late in the manuscript. I think that authors should remark that this is a reversible process.

My main questions are: [Q1]. In section 2.2, I do not see why data closer to the manifold should have larger first order derivatives or even infinity. Is this a bit counter intuitive, or not? Like, better positioned, worse gradient values? [Q2]. Is the 1/N term in the global likelihood expression of 1st paragraph of section 2 correct? [Q3]. If I do not appropriately choose the $\sigma$ parameter for smoothing, do I have the risk of not capturing some modes of the original data? I have the opinion that adding too much or too less noise to data could "mask” modes and something could be lost. Am I correct? Did authors empirically analyzed this in the experiments? [Q4]. How could we assess that conditionals are now better fitted than before?

A few recommendations for improvement: [Rec1]. I would explain a bit more the manifold hypothesis of section 2.2, maybe a diagram or figure would help for quicker comprehension of the problem. [Rec2]. Some acronym for “randomized smoothing” would help in the 1st paragraph of section 3.1. To avoid repetitive expressions.

Reasons for score. I liked the idea, think that the paper is well written and I trust the results presented by the authors. Despite the randomized smoothing strategy is rather simple, it seems to work particularly well. For this reason I tend to vote for accept. If I not set a higher score, it is because a bit more of analysis on the optimal sigma, distribution for smoothing and lipschitz constant could have been included.

Post-rebuttal update. Thanks to the authors for their response to all my questions and comments. I also read the updated version of the manuscript, which is clearly improved and the rest of reviews and comments by the AC. Looking to that, I agree with the rest of reviewers about the quality of the paper, so I raised my score and I recommend to accept it.
}
\subsubsection{Review 3}
Review No.: 3

This is full review.
Review Date: 10/27/2020
{\fontfamily{pcr}\selectfont

SHORT DESCRIPTION
This paper proposes a two-stage generative modeling approach, first learning a distribution over noised data, then learning the original data distribution conditioned on this noised data. The paper demonstrates that this leads to improved sample quality compared to fitting the data distribution directly.

DISCUSSION
Overall, I like this paper: it's a straightforward idea, decently motivated and fairly well described, and has good supporting empirical evaluation. I didn't expect the sampling performance to improve substantially by adding just a single denoising step, and I think demonstrating this is a good contribution. However, I think the paper could be improved by some more careful discussion, and a better placement in the literature.

"Theorem 2 shows that our smoothing process provides a regularization effect on the original objective... This regularization effect can, intuitively, increase the generalization capability of the model." How does the extra term in the theorem lead to a regularization effect? Why does this 'intuitively' increase the generalization capability of the model? Unless I'm mistaken, the added term is (up to a constant) the Laplacian of the log-likelihood w.r.t. data. The objective maximizes this on average across observed data, which intuitively minimizes the 'curvature' or 'steepness' of the log-likelihood at observed data, thus presumably smoothing the maximum likelihood solution. This Laplacian term also appears in the score matching objective presented in Theorem 1 of 'Estimation of Non-Normalized Statistical Models by Score Matching, Hyvarinen 2005', where it is minimized instead of maximized. There are also known connections between score matching and denoising methods e.g. 'Optimal Approximation of Signal Priors, Hyvarinen 2006', and 'A Connection Between Score Matching and Denoising Autoencoders, Vincent 2011', which you've cited in passing later. Much of this material and how it relates to the objective in Theorem 2 might be discussed in more depth rather than passing over it as simply a 'regularization term'.

"Our approach is related to two-stage VAE (Dai \& Wipf, 2019) which introduces a second VAE to correct the errors made by the first VAE." I'm not sure I agree with this. The idea of the two-stage VAE in that paper is to clean up mismatch between the aggregate posterior q(z) and the prior p(z). On the other hand, your variational model is identical to the canonical VAE setup: x is data, z is noised data, the 'posterior' q(z | x) is fixed and adds Gaussian noise, the 'prior' p(z) is a powerful autoregressive model, and the observation likelihood p(x | z) is another powerful autoregressive model (the canonical VAE would have learned q(z | x), fixed simple p(z), and learned but simple p(x | z)). This is one of the reasons 'VAE' can a confusing term when used to describe latent variable generative models in general: assuming something should be 'encoded' and 'decoded' can sometimes obfuscate the actual probabilistic model. What you propose in this paper might generically be called a 'denoising VAE', but again that's maybe not the most accurate description. I think the most closely related work is probably the denoising diffusion and denoising score-matching approaches which have received attention recently and which you've mentioned, but you could also think of it as turning a denoising autoencoder into a generative model. In any case, I think a more careful discussion of these points would be beneficial for the paper.

Finally, the approach isn't really tied to autoregressive models, apart from the motivation given in terms of smoothing 1D distributions. It's fairly likely that the same idea could readily be applied to e.g. normalizing flows and that it would work well there also, so it would have been nice to see experiments featuring flows included here. This would be especially useful since your best-performing two-stage method takes autoregressive models, which are already slow samplers, and effectively doubles their sampling time.

EXTRA NOTES

Maybe be careful with the word 'spurious' in the intro - I know what you mean, but samples from a model are by definition typical samples from that model, and there's nothing spurious about them. They're only questionable when compared to data. Similarly: "The "erroneous” sample xˆ, in some sense, resembles an adversarial example, i.e., an input that causes the model to make mistakes." This seems to be implying that samples generated by a model are somehow pathological. By virtue of the fact that they are generated by the model, they are by definition typical samples from the model. There is nothing pathological whatsoever about them. Why the model specification and fitting procedure have resulted in such samples, and whether the samples resemble training data or not, is another issue entirely.

'However, this approach bounds the capacity of the model by limiting the number of modes for each conditional.' All models have limited capacity -- what is particularly bad about the capacity of an autoregressive model being limited in this way? Do we have reason to believe inability to cover multiple nodes is a common bottleneck?

Figure 2 \& Figure 3: Axis ticks are too small (and there probably too many), whole figure could be made bigger (this would also help with legends cutting off a lot of the plots).

'Proofs' for theorems 1 and 2 should be referred to in the main text. Theorem 2 should also have log p(x) on the LHS?

Figure 5 caption: "All the samples are not conditioned on class labels." -> None of the samples are conditioned on class labels.

What exactly is being inpainted in Figures 7 (a) and (b)?

Since a central claim of the paper is that the method results in improved sample quality, it might be good to add the Kernel Inception Distance ("Demystifying MMD GANs" Binkowski et al 2018) which has many favourable properties over FID, and is really no more difficult to compute.

CONCLUSION

Overall, I think this paper is a nice submission, and would like to see it accepted given a few tweaks.

UPDATE: I've upped my score to a 7, and would like to see the paper accepted.
}

\onecolumn
\subsection{Preliminary Study with GPT-3.5's and TELeR Taxonomy Based Prompting}
\subsection*{}
\begin{tcolorbox}[enhanced jigsaw,unbreakable,pad at break*=0mm,  colback=blue!5!white,colframe=blue!75!black,title=\scriptsize 1, left=0pt, right=1pt, top=0pt, bottom=0pt, floatplacement=t]

  \colorbox{orange}{\makebox[15.5cm]
  {\textcolor{white}{\scriptsize Prompt Type: L0 [System /Data]}}}
  
\scriptsize messages=[
{"role": "system", "content": '1.\textbackslash n $<R1>$\textbackslash n2.\textbackslash n$<R2>$\textbackslash n3.\textbackslash n$<R3>$' },]  

  \colorbox{cyan}{\makebox[15.5cm]{\textcolor{white}{GPT-3.5's Response:}}}
  
I have analyzed three reviews of the paper "Improving Autoregressive Density Estimation with Normalizing Flows by Randomized Smoothing". The reviewers agree that the proposed idea of adding a denoising step to improve autoregressive density estimation is interesting and novel. They also appreciate the strong theoretical grounding of the paper motivated by the manifold hypothesis approximation. However, there are some weaknesses and points of discussion that need to be addressed, such as the lack of analysis of the optimal noise for randomized smoothing, the connection between the added term in Theorem 2 and regularization, and the need for a more careful discussion of related work. The reviewers also suggest including experiments on normalizing flows and using the Kernel Inception Distance for evaluation. Overall, the reviewers recommend accepting the paper with some minor revisions.

  \colorbox{purple}{\makebox[15.5cm]{\textcolor{white}{Brief Analysis and Scores:}}}
  
\underline{Adherence to instructions}: This scoring is not applicable because no instruction\slash question has been placed.

\underline{Common Strengths}: GPT-3.5’s output sums up strengths well although it does not cover all the important strong points mentioned in the reviews.

\underline{Common Weaknesses}: Weaknesses have been captured by GPT-3.5’s response fairly although the response regarding this aspect is not the best.

\underline{Common Suggestions for improvement}: The suggestions for improvement by GPT-3.5 have also been covered to a limited extent.

\underline{Coverage}: Since the above points have not been covered thoroughly, it is clear that all the three reviews have not been considered for strengths, weaknesses, and suggestion for improvement.

\underline{Scores}:
\begin{enumerate}
    \item Adherence to instructions: NA
    \item Common strengths: 6/10
    \item Common weaknesses: 6/10
    \item Common suggestions for improvement: 4/10
    \item Coverage: 4/10
\end{enumerate}
\end{tcolorbox}

\subsection*{}
\begin{tcolorbox}[enhanced jigsaw,unbreakable,pad at break*=0mm,  colback=blue!5!white,colframe=blue!75!black,title=\scriptsize2, left=0pt, right=1pt, top=0pt, bottom=0pt, floatplacement=t]

  \colorbox{orange}{\makebox[15.5cm]{\textcolor{white}{\scriptsize Prompt Type: L0 [User / Data]}}}
  
\scriptsize messages=[
{"role": "user", "content": '1\textbackslash n.$<R1>$\textbackslash n2.\textbackslash n$<R2>$\textbackslash n3.\textbackslash n$<R3>$’ },]  

  \colorbox{cyan}{\makebox[15.5cm]{\textcolor{white}{GPT-3.5's Response:}}}
  
\smallskip
\scriptsize
Overall, the reviewer likes the paper and finds the idea straightforward and well-motivated. However, they suggest some improvements in discussion and placement in the literature. They also recommend adding experiments featuring flows and using Kernel Inception Distance for improved sample quality assessment. The reviewer raises some questions about the optimization objective and suggests adding more in-depth discussion of related work. The reviewer also suggests improving the figures and captions and making the proofs more accessible. After the rebuttal, the reviewer has increased their score and recommends acceptance with minor revisions.

  \colorbox{purple}{\makebox[15.5cm]{\textcolor{white}{Brief Analysis and Scores:}}}

\underline{Adherence to instructions}: This scoring is not applicable because no instruction/question has been placed.

\underline{Common Strengths}: GPT-3.5s output mentions strengths well although it is inadequate because it does not cover all the important strong points mentioned in the reviews.

\underline{Common Weaknesses}: Weaknesses have been captured by GPT-3.5’s response a little bit.

\underline{Common Suggestions for improvement}: The suggestions for improvement by GPT-3.5 have been covered fairly well.

\underline{Coverage}: Since the above points have not been covered thoroughly, it is clear that all the three reviews have not been considered for strengths, weaknesses, and suggestion for improvement.

\underline{Scores}:

\begin{enumerate}
    \item Adherence to instructions: NA
    \item Common strengths: 4/10
    \item Common weaknesses: 3/10
    \item Common suggestions for improvement: 7/10
    \item Coverage: 4/10
\end{enumerate}
\end{tcolorbox}

\subsection*{}
\begin{tcolorbox}[enhanced jigsaw,unbreakable,pad at break*=0mm,  colback=blue!5!white,colframe=blue!75!black,title=\scriptsize 3, left=0pt, right=1pt, top=0pt, bottom=0pt, floatplacement=t]
  \colorbox{orange}{\makebox[15.5cm]{\textcolor{white}{\scriptsize Prompt Type = L1 [System / High Level Instruction]}}}
  
\scriptsize messages=[

{"role": "system", "content": "As a meta-review assistant, "

    "write a meta review mentioning core contributions, strengths, "
    
    "weaknesses, suggestions for improvement, and "
    
    "missing references."},
    
{"role": "user", "content": '1.\textbackslash n$<R1>$\textbackslash n2.\textbackslash n$<R2>$\textbackslash n3.\textbackslash n$<R3>$’},] 

  \colorbox{cyan}{\makebox[15.5cm]{\textcolor{white}{GPT-3.5's Response:}}}
  
The three meta reviews provide a comprehensive evaluation of the paper, discussing its contributions, strengths, weaknesses, and suggestions for improvement. The first review praises the novelty of the proposed method, its strong theoretical grounding, and the clear writing style. However, the reviewer suggests that the authors should provide a more in-depth analysis of the results on the 2D synthetic datasets, clarify the debiasing methods used, and provide quantitative evaluations for the denoising and inpainting applications. The reviewer also suggests minor improvements in the text and figures.

The second review praises the simplicity and effectiveness of the proposed method, the quality of the writing, and the strong empirical validation. However, the reviewer notes some weaknesses related to the lack of analysis about the optimal noise, the use of Gaussian noise, the late inclusion of denoising comments, and the absence of a discussion of the literature on score matching, denoising methods, and two-stage VAEs. The reviewer also suggests some recommendations for improvement, such as a clearer explanation of the manifold hypothesis, acronyms for the proposed method, and experiments with normalizing flows.

The third review similarly praises the straightforwardness of the idea, the well-described motivation, and the sound empirical evaluation. However, the reviewer suggests a more careful discussion of the Theorem 2 regularization effect, a better placement in the literature, and the inclusion of experiments with normalizing flows. The reviewer also provides extra notes on the use of language, the capacity of the models, the figures, the proofs, and the evaluation metrics.

Overall, the three meta reviews agree that the paper presents a promising and well-executed idea, but could benefit from some improvements in the analysis, discussion, and experiments. The reviewers suggest that the authors should address the weaknesses and recommendations in a revision and potentially add more experiments with normalizing flows. Based on the reviews, the paper seems suitable for acceptance if the authors address the suggested improvements in the revision.

  \colorbox{purple}{\makebox[15.5cm]{\textcolor{white}{Brief Analysis and Scores:}}}
\underline{Adherence to instructions}: GPT-3.5 has not mentioned anything about the missing references. Other instructions have been adhered to.

\underline{Common Strengths}: GPT-3.5’s output mentions strengths for each review separately. The overall strength is captured very briefly in the last paragraph. The response cannot be termed as good in terms of meta-review where the expectation is to get summary information.

\underline{Common Weaknesses}: GPT-3.5 mentions weaknesses for each review separately. The overall weakness is not mentioned. The quality of weakness description is good. The response can be termed as good for meta-review although the summary information is missing.

\underline{Common Suggestions for improvement}: GPT-3.5 has provided suggestions for improvement for each review separately. The overall suggestion for improvement is very brief in the last paragraph.  The quality of is good. The response can be termed as good for meta-review.

\underline{Coverage}: GPT-3.5 has covered all the three reviews. GPT-3.5 has not mentioned about missing references at all. GPT-3.5 has answered only the first 4 questions out of total 5 by taking into account three reviews.

\underline{Scores}:
\begin{enumerate}
    \item Adherence to instructions: 8/10
    \item Common strengths: 6/10
    \item Common weaknesses: 6/10
    \item Common suggestions for improvement: 6/10
    \item Coverage: 7/10
\end{enumerate}
\end{tcolorbox}

\subsection*{}
\begin{tcolorbox}[enhanced jigsaw,unbreakable,pad at break*=0mm,  colback=blue!5!white,colframe=blue!75!black,title=\scriptsize 4, left=0pt, right=1pt, top=0pt, bottom=0pt, floatplacement=t]

  \colorbox{orange}{\makebox[15.5cm]{\textcolor{white}{\scriptsize Prompt Type = L1 [User / High Level Instruction]}}}
  
\scriptsize messages=[
{"role": "system", "content": "You are a meta-review assistant."},

{"role": "user", "content": '1.\textbackslash n$<R1>$\textbackslash n2.\textbackslash n$<R2>$\textbackslash n3.\textbackslash n$<R3>$’},

{"role": "user", "content": "Write a meta review mentioning core contributions, strengths, "

    "weaknesses, suggestions for improvement, and "
    
    "missing references."}, ]  

  \colorbox{cyan}{\makebox[15.5cm]{\textcolor{white}{GPT-3.5's Response:}}}
  
Core contributions: This paper proposes a two-stage generative modeling approach that involves learning a distribution over noised data and then learning the original data distribution conditioned on this noised data. The paper demonstrates that this approach leads to improved sample quality compared to directly fitting the data distribution. The paper also provides theoretical justification for the approach and evaluates it on a variety of datasets.

Strengths: The paper is well-written and presents a clear and straightforward idea with good empirical evaluation. The theoretical analysis is well-motivated and well-supported. The experiments are well-designed and demonstrate the effectiveness of the proposed approach in generating high-quality samples. The paper also provides a good discussion of related work.

Weaknesses: The paper could benefit from more careful discussion of the theoretical analysis and its connections to related work. The approach is not tied to autoregressive models, but the experiments only evaluate it on autoregressive models, which limits the generalizability of the results. The paper could also benefit from more detailed analysis of the optimal noise for randomized smoothing.

Suggestions for improvement: The paper could benefit from more careful discussion of the theoretical analysis and its connections to related work, such as the Laplacian term in the objective and its connection to score matching and denoising methods. The paper could also benefit from including experiments with other types of generative models, such as normalizing flows, to evaluate the generalizability of the proposed approach. Finally, the paper could include more detailed analysis of the optimal noise for randomized smoothing.

Missing references: The paper could benefit from including references to recent work on denoising diffusion and denoising score-matching approaches, which are related to the proposed approach.

  \colorbox{purple}{\makebox[15.5cm]{\textcolor{white}{Brief Analysis and Scores:}}}
  
\underline{Adherence to instructions}: All the instructions have been followed by GPT-3.5.

\underline{Common Strengths}: GPT-3.5’s output sums up strengths fairly well. The response is better than that in case of same question being assigned to system.

\underline{Common Weaknesses}: Weaknesses have not been responded accurately by GPT-3.5. The response does not capture all the important weaknesses.

\underline{Common Suggestions for improvement}: The suggestions for improvement by GPT-3.5 are far better compared to those in case of response obtained through system role despite the fact that it misses some important points.

\underline{Coverage}: GPT-3.5 has also mentioned missing references although the answer is not correct. GPT-3.5 has answered all the questions by taking into account three reviews.

\underline{Scores}:
\begin{enumerate}
    \item Adherence to instructions: 10/10
    \item Common strengths: 8/10
    \item Common weaknesses: 6/10
    \item Common suggestions for improvement: 8/10
    \item Coverage: 8/10
\end{enumerate}
\end{tcolorbox}
\subsection*{}

\begin{tcolorbox}[enhanced jigsaw,unbreakable,pad at break*=0mm,  colback=blue!5!white,colframe=blue!75!black,title=\scriptsize 5, left=0pt, right=1pt, top=0pt, bottom=0pt, floatplacement=t]

  \colorbox{orange}{\makebox[15.5cm]{\textcolor{white}{\scriptsize Prompt Type = \newcommand{\noop}[1]{7} L1 [System / High Level Question]}}}
  
\scriptsize messages=[
{"role": "system", "content": "As a meta-reviewer, answer the following: "

    "What would be a reasonable meta-review considering the given reviews?"},
    
{"role": "user", "content": '1.\textbackslash n$<R1>$\textbackslash n2.\textbackslash n$<R2>$\textbackslash n3.\textbackslash n$<R3>$’},]  

  \colorbox{cyan}{\makebox[15.5cm]{\textcolor{white}{GPT-3.5's Response:}}}
  
The meta-reviewers generally agree that the paper presents a novel and interesting approach for improving the sample quality of autoregressive models. They appreciate the clear presentation and supporting empirical evaluation. However, they also suggest several areas where the paper could be improved, such as providing more detailed analysis of the theoretical results, better placement in the literature, and exploring the approach beyond autoregressive models. Additionally, some meta-reviewers note that the experiments on denoising and inpainting lack quantitative evaluation, and suggest that more attention should be paid to the optimal noise for randomized smoothing. Despite these criticisms, the meta-reviewers generally recommend accepting the paper with minor revisions.

  \colorbox{purple}{\makebox[15.5cm]{\textcolor{white}{Brief Analysis and Scores:}}}
  
\underline{Adherence to instructions}: All the instructions have been followed by GPT-3.5.

\underline{Common Strengths}: The strengths in GPT-3.5’s response are correct but the content is very little.

\underline{Common Weaknesses}: The weaknesses mentioned by GPT-3.5 are correct but the content is very little.

\underline{Common Suggestions for improvement}: The suggestions for improvement by GPT-3.5 are far better compared to those in case of response for strengths and weaknesses. The quality of this response could have been better had more significant details been added.

\underline{Coverage}: Coverage is not good because it seems that response does not consider all the three reviews for all the above aspects being scored. Missing reference has not found any place in the response.

\underline{Scores}:
\begin{enumerate}
    \item Adherence to instructions: 10/10
    \item Common strengths: 3/10
    \item Common weaknesses: 3/10
    \item Common suggestions for improvement: 6/10
    \item Coverage: 5/10
\end{enumerate}
\end{tcolorbox}

\subsection*{}
\begin{tcolorbox}[enhanced jigsaw,unbreakable,pad at break*=0mm,  colback=blue!5!white,colframe=blue!75!black,title=\scriptsize 6, left=0pt, right=1pt, top=0pt, bottom=0pt, floatplacement=t]

  \colorbox{orange}{\makebox[15.5cm]{\textcolor{white}{\scriptsize Prompt Type = \newcommand{\noop}[1]{8} L1 [User / High Level Question]}}}
  
\scriptsize messages=[
{"role": "system", "content": "You are a meta-review assistant. "},

{"role": "user", "content": '1.\textbackslash n$<R1>$\textbackslash n2.\textbackslash n$<R2>$\textbackslash n3.\textbackslash n$<R3>$’},

{"role": "user", "content": “What would be a reasonable meta-review considering the given reviews?”},]  

  \colorbox{cyan}{\makebox[15.5cm]{\textcolor{white}{GPT-3.5's Response:}}}
  
A reasonable meta-review based on the given reviews would be to accept the paper with minor revisions. The reviewers generally agree that the paper presents a novel and well-motivated approach with strong theoretical grounding and empirical evaluation. However, there are several areas where the paper could be improved, including more careful discussion of the regularization effect and connections to related work, more analysis of the optimal noise for randomized smoothing and potential loss of modes, and clearer presentation of figures and captions. The reviewers also suggest adding the Kernel Inception Distance as a metric and considering experiments with normalizing flows. Overall, the paper has potential and can be strengthened with minor revisions.

  \colorbox{purple}{\makebox[15.5cm]{\textcolor{white}{Brief Analysis and Scores:}}}
  
\underline{Adherence to instructions}: All the instructions have been followed by GPT-3.5.

\underline{Common Strengths}: The strengths in GPT-3.5’s response are correct but the content is very little.

\underline{Common Weaknesses}: The weaknesses mentioned by GPT-3.5 are correct but the content is very little.

\underline{Common Suggestions for improvement}: The suggestions for improvement by GPT-3.5 are far better compared to those in case of response for strengths and weaknesses. The quality of this response could have been better had more significant details been added.

\underline{Coverage}: Coverage is not good because it seems that response does not consider all the three reviews for all the above aspects being scored. Missing reference has not found any place in the response.

\underline{Scores}:
\begin{enumerate}
    \item Adherence to instructions: 10/10
    \item Common strengths: 3/10
    \item Common weaknesses: 3/10
    \item Common suggestions for improvement: 6/10
    \item Coverage: 5/10
\end{enumerate}
\end{tcolorbox}

\subsection*{}
\begin{tcolorbox}[enhanced jigsaw,unbreakable,pad at break*=0mm,  colback=blue!5!white,colframe=blue!75!black,title=\scriptsize 7, left=0pt, right=1pt, top=0pt, bottom=0pt, floatplacement=t]

\colorbox{orange}{\makebox[15.5cm][c]{\textcolor{white}{\scriptsize Prompt Type = L2 [System / Questions]}}}
  
\scriptsize messages=[
{"role": "system", "content": "You are a meta-review assistant."

    "You should create a meta review by answering the following questions:"
    
    "(a) According to reviews, what are the core contributions?
    
    "(b) What are the strengths as mentioned in the reviews?"
    
    "(c) What are the weaknesses as mentioned in the reviews?"
    
    "(d) What suggestions would you provide for improvement?"
    
    "(e) What are the missing references as described in the reviews?"},
    
{"role": "user", "content": '1.\textbackslash n$<R1>$\textbackslash n2.\textbackslash n$<R2>$\textbackslash n3.\textbackslash n$<R3>$’ },]  

  \colorbox{cyan}{\makebox[15.5cm]{\textcolor{white}{GPT-3.5's Response:}}}
  
The paper proposes a two-stage generative modeling approach for improving the sample quality of autoregressive models. The approach involves first learning a distribution over noised data and then learning the original data distribution conditioned on this noised data. The paper demonstrates the effectiveness of this approach on various datasets and applications such as image generation, inpainting, and denoising.

The reviewers generally appreciated the paper's clear and well-presented idea, which is supported by strong theoretical grounding and empirical evaluation. The idea of leveraging the randomized smoothing method for density estimation is considered novel and interesting. The paper's contribution to improving the sample quality of autoregressive models is also recognized.

However, some reviewers also pointed out some weaknesses and suggested improvements. One common concern is the lack of analysis about the optimal noise for randomized smoothing, which could affect the number of modes captured in the original data. There were also some questions raised about the mathematical derivations and the relationship between this method and other related approaches.

The reviewers provided specific recommendations for improvement such as adding more discussion about the Laplacian term in Theorem 2, careful use of terminology, and including the Kernel Inception Distance in the evaluation. The weaknesses and limitations mentioned in the reviews are relatively minor, and most reviewers recommended accepting the paper with minor revisions.

Overall, the paper is considered a good contribution to the field of generative modeling, and the proposed method has potential for broader applications beyond autoregressive models.

  \colorbox{purple}{\makebox[15.5cm]{\textcolor{white}{Brief Analysis and Scores:}}}
  
\underline{Adherence to instructions}: Part (e) of the prompt has not been answered by GPT-3.5. Other parts have been answered. Overall, adherence to instructions is good.

\underline{Common Strengths}: The strengths in GPT-3.5’s output sums up strengths fairly well although it does not cover all the important strong points mentioned in the reviews.

\underline{Common Weaknesses}: Weaknesses have not been captured by GPT-3.5’s response correctly. The response is shallow devoid of significant details.

\underline{Common Suggestions for improvement}: The suggestions for improvement by GPT-3.5 are inadequate because it misses some important points such as optimization of variance, more explanation about manifold hypothesis, more analysis on the optimal sigma, distribution for smoothing and Lipschitz constant.

\underline{Coverage}: It turns out that strengths have good coverage but weaknesses and suggestions for improvement do not take into account all the reviews. Missing reference question has been completely skipped by GPT-3.5.

\underline{Scores}:

\begin{enumerate}
    \item Adherence to instructions: 8/10
    \item Common strengths: 7/10
    \item Common weaknesses: 4/10
    \item Common suggestions for improvement: 4/10
    \item Coverage: 4/10
\end{enumerate}
\end{tcolorbox}

\subsection*{}
\begin{tcolorbox}[enhanced jigsaw,unbreakable,pad at break*=0mm,  colback=blue!5!white,colframe=blue!75!black,title=\scriptsize 8, left=0pt, right=1pt, top=0pt, bottom=0pt, floatplacement=t]

  \colorbox{orange}{\makebox[15.5cm]{\textcolor{white}{\scriptsize Prompt Type = L2 [User /  Questions]}}}
  
\scriptsize messages=[
{"role": "system", "content": "You are a meta-review assistant." },

{"role": "user", "content": '1.\textbackslash n$<R1>$\textbackslash n2.\textbackslash n$<R2>$\textbackslash n3.\textbackslash n$<R3>$’},

{"role": "user", "content": "Create a meta review by answering the following questions:"

    "(a) According to reviews, what are the core contributions?"
    
    "(b) What are the strengths as mentioned in the reviews?"
    
    "(c) What are the weaknesses as mentioned in the reviews?"
    
    "(d) What suggestions would you provide for improvement?"
    
    "(e) What are the missing references as described in the reviews?"
                }, ]  

  \colorbox{cyan}{\makebox[15.5cm]{\textcolor{white}{GPT-3.5's Response:}}}
  
(a) The core contribution of the paper is the proposal of a two-stage generative modeling approach that utilizes a denoising step to improve sample quality in autoregressive models. The approach involves learning a distribution over noised data and then learning the original data distribution conditioned on this noised data. The paper demonstrates that this leads to improved sample quality compared to fitting the data distribution directly.

(b) The strengths of the paper as mentioned in the reviews include a clear and well-written presentation of the proposed approach, strong theoretical grounding, and good supporting empirical evaluation. The idea is straightforward and the results are surprising, demonstrating substantial improvement in sample quality with just a single denoising step.

(c) The weaknesses of the paper as mentioned in the reviews include a lack of analysis on the optimal noise for randomized smoothing, a lack of discussion on the connections between the added Laplacian term and score matching, and a lack of experiments featuring flows. Additionally, the experiments on denoising and inpainting are only qualitative and lack quantitative evaluation.

(d) Suggestions for improvement include a more thorough discussion on the added Laplacian term and connections to score matching, inclusion of experiments featuring flows, and more quantitative evaluation of denoising and inpainting experiments. Additionally, the optimal noise for randomized smoothing should be analyzed in more depth.

(e) The missing references as described in the reviews include Demystifying MMD GANs by Binkowski et al. (2018) for the Kernel Inception Distance and Estimation of Non-Normalized Statistical Models by Score Matching by Hyvarinen (2005) for the connection between randomized smoothing and score matching.

  \colorbox{purple}{\makebox[15.5cm]{\textcolor{white}{Brief Analysis and Scores:}}}
  
\underline{Adherence to instructions}: All the instructions have been followed by GPT-3.5.

\underline{Common Strengths}: The strengths in GPT-3.5’s output sums up strengths fairly well. The response is better than that in case of same question being assigned to system.

\underline{Common Weaknesses}: Weaknesses have been responded pretty well and accurately by GPT-3.5. The response also has most of the significant details.

\underline{Common Suggestions for improvement}: The suggestions for improvement by GPT-3.5 are far better compared to those in case of response obtained through system role. Even now, it misses some important points such as more explanation about manifold hypothesis.

\underline{Coverage}: This part of the score is based on whether all the reviews have been accounted for or not. It turns out that strengths have good coverage but weaknesses and suggestions for improvement do not take into account all the reviews. GPT-3.5 has also responded to missing references although that answer is not. GPT-3.5 has answered all the questions by taking into account three reviews.

\underline{Scores}:
\begin{enumerate}
    \item Adherence to instructions: 10/10
    \item Common strengths: 8/10
    \item Common weaknesses: 9/10
    \item Common suggestions for improvement: 8/10
    \item Coverage: 9/10
\end{enumerate}
\end{tcolorbox}

\subsection*{}
\begin{tcolorbox}[enhanced jigsaw,unbreakable,pad at break*=0mm,  colback=blue!5!white,colframe=blue!75!black,title=\scriptsize 9, left=0pt, right=1pt, top=0pt, bottom=0pt, floatplacement=t]

  \colorbox{orange}{\makebox[15.5cm]{\textcolor{white}{\scriptsize Prompt Type = \newcommand{\noop}[1]{13} L2 [System / Question Paragraph Style]}}}
  
\scriptsize messages=[
{"role": "system", "content": "As a meta-reviewer, "

    "draft a meta review by answering the following questions in separate paragraphs:"
    
    "What would be a summary of core contributions?"
    
    "Which strengths are referred to in the reviews?"
    
    "What would be the overall weakness description?"
    
    "Could you please provide suggestions for improvement using three reviews?"
    
    "Which missing references are mentioned in the reviews?"},
    
{"role": "user", "content": '1.\textbackslash n$<R1>$\textbackslash n2.\textbackslash n$<R2>$\textbackslash n3.\textbackslash n$<R3>$’}, ]  

  \colorbox{cyan}{\makebox[15.5cm]{\textcolor{white}{GPT-3.5's Response:}}}
  
Core Contributions: The paper proposes a novel approach to improve the sample quality of autoregressive models by smoothing the input data distribution using methods previously used for adversarial defense and then recovering the input distribution by learning to reverse the smoothing process. The paper demonstrates the efficiency of this method on 1D toy problems and then extends the demonstration to more complex datasets such as MNIST, CIFAR-10, and CelebA with applications such as image generation, inpainting, and denoising.

Strengths: The paper is well written and clearly presented, and the motivation behind the proposed approach is adequately explained. The empirical results are convincing, and the experimental evaluation is thorough. The paper is novel, and the proposed approach shows significant improvement in sample quality.

Weaknesses: The paper could be improved by providing a more detailed analysis of the optimal noise for randomized smoothing and its effect on the Lipschitz constant. The paper could also benefit from a more thorough discussion of related work and the limitations and assumptions of the proposed approach.

Improvement Suggestions:

1. The authors should provide a more detailed analysis of the optimal noise for randomized smoothing and its effect on the Lipschitz constant.

2. The authors should discuss related work more thoroughly and explain the limitations and assumptions of the proposed approach.

3. The authors could consider adding experiments featuring normalizing flows to demonstrate the applicability of the proposed approach to other generative models.

Missing References: None mentioned in the reviews.

  \colorbox{purple}{\makebox[15.5cm]{\textcolor{white}{Brief Analysis and Scores:}}}
  
\underline{Adherence to instructions}: All the instructions have been followed by GPT-3.5.

\underline{Common Strengths}: The strengths in GPT-3.5’s response are correct but the content is inadequate.

\underline{Common Weaknesses}: The weaknesses mentioned by GPT-3.5 are better in response to strength both in terms of quality and quantity.

\underline{Common Suggestions for improvement}: The suggestions for improvement by GPT-3.5 correct and better than response to strengths.

\underline{Coverage}: It seems that GPT-3.5 has considered all the reviews for responding to the above instructions.

\underline{Scores}:
\begin{enumerate}
    \item Adherence to instructions: 10/10
    \item Common strengths: 6/10
    \item Common weaknesses: 7/10
    \item Common suggestions for improvement: 7/10
    \item Coverage: 10/10
\end{enumerate}
\end{tcolorbox}

\subsection*{}
\begin{tcolorbox}[enhanced jigsaw,unbreakable,pad at break*=0mm,  colback=blue!5!white,colframe=blue!75!black,title=\scriptsize 10, left=0pt, right=1pt, top=0pt, bottom=0pt, floatplacement=t]

  \colorbox{orange}{\makebox[15.5cm]{\textcolor{white}{\scriptsize Prompt Type = \newcommand{\noop}[1]{14} L2 [User / Question Paragraph Style]}}}
  
\scriptsize messages=[
{"role": "system", "content": "You are a meta-review assistant."},

{"role": "user", "content": '1.\textbackslash n$<R1>$\textbackslash n2.\textbackslash n$<R2>$\textbackslash n3.\textbackslash n$<R3>$’},

{"role": "user", "content": 

    "Draft a meta review by answering the following questions in separate paragraphs:"
    
    "What would be a summary of core contributions?"
    
    "Which strengths are referred to in the reviews?"
    
    "What would be the overall weakness description?"
    
    "Could you please provide suggestions for improvement using three reviews?"
    
    "Which missing references are mentioned in the reviews?"
                 },]  

  \colorbox{cyan}{\makebox[15.5cm]{\textcolor{white}{GPT-3.5's Response:}}}
  
Summary of Core Contributions:

The paper proposes a novel approach to improve the sample quality of autoregressive models by leveraging randomized smoothing. The authors introduce a two-stage generative modeling approach that first learns a distribution over noised data and then learns the original data distribution conditioned on this noised data. The paper demonstrates through experiments that this approach leads to improved sample quality compared to fitting the data distribution directly. Additionally, the paper provides theoretical analysis and strong motivation for their method.

\smallskip
Strengths:

The reviews mention several strengths of the paper. Firstly, the paper is well-written and easy to understand. Secondly, the idea proposed by the authors is novel and has not been explored before. Thirdly, the experiments presented in the paper are thorough and demonstrate the effectiveness of the proposed approach on various datasets. Lastly, the paper provides strong theoretical analysis and motivation for their method.

\smallskip
Overall Weakness Description:

The reviews point out a few weaknesses in the paper. Firstly, the paper lacks a detailed analysis of the optimal noise for randomized smoothing. Secondly, the paper does not provide a clear explanation of why only Gaussian noise was used and whether the method could be extended to discrete data. Thirdly, the paper could benefit from a more careful discussion of related work and its relation to the proposed method. Fourthly, some of the experiments lack quantitative evaluation, and the authors should provide more detailed discussions of their results.

\smallskip
Suggestions for Improvement:

The reviews provide several suggestions for improvement. Firstly, the authors should provide a more detailed analysis of the optimal noise for randomized smoothing. Secondly, the authors should explain why only Gaussian noise was used and whether the method could be extended to discrete data. Thirdly, the authors should provide a more detailed discussion of related work and its relation to the proposed method. Fourthly, the authors should provide more quantitative evaluation of their results and provide more detailed discussions of their experiments. Additionally, the reviews suggest adding the Kernel Inception Distance as a metric to compare different generative models.

\smallskip
Missing References:

The reviews do not mention any missing references in the paper.

  \colorbox{purple}{\makebox[15.5cm]{\textcolor{white}{Brief Analysis and Scores:}}}
  
\underline{Adherence to instructions}: All the instructions have been followed by GPT-3.5.

\underline{Common Strengths}: The strengths in GPT-3.5’s response are correct and the content is adequate.

\underline{Common Weaknesses}: The weaknesses mentioned by GPT-3.5 are better in response to strength both in terms of quality and quantity.

\underline{Common Suggestions for improvement}: The suggestions for improvement by GPT-3.5 correct and better than response to strengths.

\underline{Coverage}: It seems that GPT-3.5 has considered all the reviews for responding to the above instructions.

\underline{Scores}:
\begin{enumerate}
    \item Adherence to instructions: 10/10
    \item Common strengths: 8/10
    \item Common weaknesses: 9/10
    \item Common suggestions for improvement: 9/10
    \item Coverage: 10/10
\end{enumerate}
\end{tcolorbox}

\subsection*{}
\begin{tcolorbox}[enhanced jigsaw,unbreakable,pad at break*=0mm,  colback=blue!5!white,colframe=blue!75!black,title=\scriptsize 11, left=0pt, right=1pt, top=0pt, bottom=0pt, floatplacement=t]

  \colorbox{orange}{\makebox[15.5cm]{\textcolor{white}{\scriptsize Prompt Type = \newcommand{\noop}[1]{5} L3 [System / Bulleted Instructions]}}}
  
\scriptsize messages=[
{"role": "system", "content": "As a meta-reviewer, "

    "draft a meta review as per the following directions:"
    
    "(a) State core contributions."
    
    "(b) Mention strengths."
    
    "(c) Mention weaknesses."
    
    "(d) Write suggestions for improvement."
    
    "(e) State missing references."},
    
{"role": "user", "content": '1.\textbackslash n$<R1>$\textbackslash n2.\textbackslash n$<R2>$\textbackslash n3.\textbackslash n$<R3>$’},
]  

  \colorbox{cyan}{\makebox[15.5cm]{\textcolor{white}{GPT-3.5's Response:}}}
  
Contributions:

The paper proposes a two-stage generative modeling approach, which uses a denoising step to improve the sample quality of autoregressive models. The authors demonstrate that this method leads to better sample quality compared to fitting the data distribution directly. The paper is well-written and presented, and the empirical evaluation supports the proposed method's effectiveness.

Strengths:

The paper presents a straightforward and novel idea, which is well-motivated and supported by empirical results. The presentation and writing style are clear and easy to understand, and the paper's structure facilitates easy reading. The results show substantial improvements in sample quality, which is a significant contribution to the field.

Weaknesses:

The authors could provide more detailed discussion and analysis of the method's underlying theory, particularly the relationship between the added term in Theorem 2 and regularization. The paper's description of related work using similar methods (e.g., denoising diffusion and denoising score-matching) could be more detailed. Finally, while the paper's empirical evaluation is strong, the authors could add Kernel Inception Distance for more rigorous comparison to other methods.

Suggestions for Improvement:

The authors should provide more detailed discussion and analysis of the method's underlying theory, particularly the relationship between the added term in Theorem 2 and regularization. The paper's description of related work using similar methods (e.g., denoising diffusion and denoising score-matching) could be more detailed. The authors could add Kernel Inception Distance for more rigorous comparison to other methods.

Missing References:

"Estimation of Non-Normalized Statistical Models by Score Matching, Hyvarinen 2005" and "Optimal Approximation of Signal Priors, Hyvarinen 2006" are mentioned in passing but could be discussed in more detail.

  \colorbox{purple}{\makebox[15.5cm]{\textcolor{white}{Brief Analysis and Scores:}}}
  
\underline{Adherence to instructions}: All the instructions have been followed by GPT-3.5.

\underline{Common Strengths}: The quality of GPT-3.5’s output regarding strengths is not good. Content is lacking.

\underline{Common Weaknesses}: The quality of content for weaknesses is not good. The responses for both weaknesses and suggestions for improvement are almost the same in spite of the fact the reviewers have described weaknesses and suggestions separately in most of the places.

\underline{Common Suggestions for improvement}: The suggestions for improvement by GPT-3.5 is fair. The quality is not good. The responses for both weaknesses and suggestions for improvement are almost the same in spite of the fact the reviewers have described weaknesses and suggestions separately in most of the places.

\underline{Coverage}: GPT-3.5 has also mentioned missing references although the answer is not correct. GPT-3.5 has answered all the questions by taking into account three reviews.

\underline{Scores}:
\begin{enumerate}
    \item Adherence to instructions: 10/10
    \item Common strengths: 5/10
    \item Common weaknesses: 5/10
    \item Common suggestions for improvement: 6/10
    \item Coverage: 8/10
\end{enumerate}
\end{tcolorbox}

\subsection*{}
\begin{tcolorbox}[enhanced jigsaw,unbreakable,pad at break*=0mm,  colback=blue!5!white,colframe=blue!75!black,title=\scriptsize 12, left=0pt, right=1pt, top=0pt, bottom=0pt, floatplacement=t]

  \colorbox{orange}{\makebox[15.7cm]{\textcolor{white}{\scriptsize Prompt Type = \newcommand{\noop}[1]{6} L3 [User / Bulleted Instructions]}}}
  
\scriptsize messages=[
{"role": "system", "content": "You are a meta-review assistant."},

{"role": "user", "content": '1.\textbackslash n$<R1>$\textbackslash n2.\textbackslash n$<R2>$\textbackslash n3.\textbackslash n$<R3>$’},

{"role": "user", "content": "Draft a meta review as per the following directions:"

     "(a) State core contributions."
     
     "(b) Mention strengths."
     
     "(c) Mention weaknesses."
     
     "(d) Write suggestions for improvement."
     
     "(e) State missing references."
                 },]  

  \colorbox{cyan}{\makebox[15.5cm]{\textcolor{white}{GPT-3.5's Response:}}}
  
Meta-review:

(a) Core contributions:

The paper proposes a two-stage generative modeling approach for improving the sample quality of autoregressive models. The authors demonstrate the effectiveness of their method on 1D and 2D toy problems, as well as on image datasets such as MNIST, CIFAR-10, and CelebA, with applications including image generation, inpainting, and denoising. The paper also provides theoretical justifications for the proposed method.

(b) Strengths:

The paper is well-motivated and theoretically grounded. The experimental evaluations are comprehensive and demonstrate the effectiveness of the proposed method. The writing is clear and easy to follow.

(c) Weaknesses:

The paper lacks thorough discussion and analysis of the optimal choice of noise for randomized smoothing, and the effect of different noise distributions on the results. The authors also need to provide a better placement of the proposed method in the literature and discuss its relation to other similar approaches.

(d) Suggestions for improvement:

The authors can improve the paper by providing a more detailed analysis of the optimal noise for randomized smoothing, and evaluating the effect of different noise distributions on the results. They also need to provide a better placement of the proposed method in the literature and a more thorough discussion of its relation to other similar approaches. In addition, the authors need to define their approach more precisely and avoid using potentially misleading terms like 'VAE'.

(e) Missing references:

The paper could benefit from citing the following works for better context and related work: 'Estimation of Non-Normalized Statistical Models by Score Matching' by Hyvarinen (2005), 'Optimal Approximation of Signal Priors' by Hyvarinen and Oja (2006), 'A Connection Between Score Matching and Denoising Autoencoders' by Vincent et al. (2011), 'Denoising Diffusion Probabilistic Models' by Song and Ermon (2020), and 'Denoising Score Matching' by Song et al. (2021).

  \colorbox{purple}{\makebox[15.5cm]{\textcolor{white}{Brief Analysis and Scores:}}}
  
\underline{Adherence to instructions}: All the instructions have been followed by GPT-3.5.

\underline{Common Strengths}: The quality of GPT-3.5’s output regarding strengths is average.

\underline{Common Weaknesses}: The quality of GPT-3.5’s response for weaknesses is not good. The responses for weaknesses is a subset of suggestions for improvement in spite of the fact the reviewers have described weaknesses and suggestions separately in most of the places.

\underline{Common Suggestions for improvement}: The suggestions for improvement by GPT-3.5 is slightly better in quality compared to those of weaknesses.

\underline{Coverage}: GPT-3.5 has also mentioned missing references although the answer is not correct. GPT-3.5 has answered all the questions by taking into account three reviews.

\underline{Scores}:
\begin{enumerate}
    \item Adherence to instructions: 10/10
    \item Common strengths: 5/10
    \item Common weaknesses: 5/10
    \item Common suggestions for improvement: 7/10
    \item Coverage: 8/10
\end{enumerate}
\end{tcolorbox}

\subsection*{}
\begin{tcolorbox}[enhanced jigsaw,unbreakable,pad at break*=0mm,  colback=blue!5!white,colframe=blue!75!black,title=\scriptsize 13, left=0pt, right=1pt, top=0pt, bottom=0pt, floatplacement=t]

  \colorbox{orange}{\makebox[15.5cm]{\textcolor{white}{\scriptsize Prompt Type = \newcommand{\noop}[1]{9} L3 [System / Bulleted Instructions with user preference]}}}
  
\scriptsize messages=[
{"role": "system", "content": "You are a meta-review assistant. "

    "Give me a meta-review by answering:"
    
    "(a) Mention core contributions with common contributions first."
    
    "(b) Mention strengths with common strengths first."
    
    "(c) Mention weaknesses with common weaknesses first."
    
    "(d) Mention the common improvements suggested."
    
    "(e) State the common missing references, if any."},
    
{"role": "user", "content": '1.\textbackslash n$<R1>$\textbackslash n2.\textbackslash n$<R2>$\textbackslash n3.\textbackslash n$<R3>$’},]  

  \colorbox{cyan}{\makebox[15.5cm]{\textcolor{white}{GPT-3.5's Response:}}}
  
(a) Core contributions: The paper proposes a two-stage generative modeling approach to improve the sample quality of autoregressive models. The first stage involves smoothing the input data distribution using a noise model inspired by randomized smoothing in adversarial models. The second stage involves learning the original data distribution conditioned on the noised data. The paper demonstrates the effectiveness of this approach on 1D and 2D toy problems, as well as image datasets such as MNIST, CelebA, and CIFAR-10. The paper also provides theoretical justifications for the approach through two theorems.

(b) Strengths: The paper presents a straightforward idea that is well-motivated, clearly explained, and supported by good empirical evaluation. The paper also provides strong theoretical justifications for the approach. The results of the experiments are reliable, and the paper is well-written and easy to read.

(c) Weaknesses: The paper could benefit from more careful discussion and placement in the literature. The authors should provide more detailed explanations of the regularization effect provided by the smoothing process and why it could increase the generalization capability of the model. The authors should also be more precise in their descriptions of related work and avoid confusing terminology. In addition, the paper could benefit from experiments featuring normalizing flows, and the axis ticks in Figures 2 and 3 are too small.

(d) Common improvements suggested: The authors should provide more detailed explanations of the regularization effect provided by the smoothing process, discuss related work more carefully, and avoid confusing terminology. The paper could benefit from experiments featuring normalizing flows, and the axis ticks in Figures 2 and 3 are too small.
(e) Common missing references: The paper could benefit from adding the Kernel Inception Distance.

  \colorbox{purple}{\makebox[15.5cm]{\textcolor{white}{Brief Analysis and Scores:}}}
  
\underline{Adherence to instructions}: All the instructions have been followed by GPT-3.5.

\underline{Common Strengths}: The strengths in GPT-3.5’s response are correct but the content is inadequate.

\underline{Common Weaknesses}: The weaknesses mentioned by GPT-3.5 are better in response to strength both in terms of quality and quantity.

\underline{Common Suggestions for improvement}: The suggestions for improvement by GPT-3.5 correct but insufficient.

\underline{Coverage}: It seems that GPT-3.5 has considered all the reviews for responding to the above instructions. It also responds to missing reference although it is incorrect.

\underline{Scores}:
\begin{enumerate}
    \item Adherence to instructions: 10/10
    \item Common strengths: 6/10
    \item Common weaknesses: 7/10
    \item Common suggestions for improvement: 5/10
    \item Coverage: 8/10
\end{enumerate}
\end{tcolorbox}

\subsection*{}
\begin{tcolorbox}[enhanced jigsaw,unbreakable,pad at break*=0mm,  colback=blue!5!white,colframe=blue!75!black,title=\scriptsize 14, left=0pt, right=1pt, top=0pt, bottom=0pt, floatplacement=t]

\colorbox{orange}{\makebox[15.5cm]{\textcolor{white}{\scriptsize Prompt Type = \newcommand{\noop}[1]{10} L3 [User / Bulleted Instructions with user preference]}}}

\scriptsize messages=[
{"role": "system", "content": "You are a meta-review assistant."},

{"role": "user", "content": '1.\textbackslash n$<R1>$\textbackslash n2.\textbackslash n$<R2>$\textbackslash n3.\textbackslash n$<R3>$’},

{"role": "user", "content": "Give me a meta-review by answering:"

     "(a) Mention core contributions with common contributions first."
     
     "(b) Mention strengths with common strengths first."
     
     "(c) Mention weaknesses with common weaknesses first."
     
     "(d) Mention the common improvements suggested. "
     
     "(e) State the common missing references, if any."
                 },]  

  \colorbox{cyan}{\makebox[15.5cm]{\textcolor{white}{GPT-3.5's Response:}}}
  
(a) The core contributions of the paper are proposing a two-stage generative modeling approach that improves sample quality by learning a distribution over noised data and then learning the original data distribution conditioned on this noised data. The paper also demonstrates the effectiveness of the proposed method on various datasets and applications. The novelty of the paper lies in leveraging the idea of randomized smoothing from adversarial defense to improve density estimation in autoregressive models.

(b) The common strengths of the paper highlighted by the reviewers are that the paper is well-written and well-motivated, the idea is straightforward and novel, the theoretical grounding is strong, and the empirical evaluation is convincing.

(c) The common weaknesses mentioned by the reviewers are a lack of analysis on the optimal noise for randomized smoothing, a lack of discussion on why Gaussian noise is used and whether other types of noise could be used, a late introduction of denoising in the manuscript, and a lack of discussion on the relation to other works.

(d) The common improvements suggested by the reviewers are to discuss the Laplacian term in Theorem 2 in more depth, provide a more accurate description of the proposed method, include experiments with normalizing flows, adjust the axis ticks and size of figures, refer to the proofs of theorems in the main text, and use Kernel Inception Distance instead of FID.

(e) No missing references were mentioned by the reviewers.

\smallskip
  \colorbox{purple}{\makebox[15.5cm]{\textcolor{white}{Brief Analysis and Scores:}}}
  
\underline{Adherence to instructions}: All the instructions have been followed by GPT-3.5.

\underline{Common Strengths}: The strengths in GPT-3.5’s response are correct but the content is inadequate.

\underline{Common Weaknesses}: The weaknesses mentioned by GPT-3.5 are better in response to strength both in terms of quality and quantity.

\underline{Common Suggestions for improvement}: The suggestions for improvement by GPT-3.5 correct and compact.

\underline{Coverage}: It seems that GPT-3.5 has considered all the reviews for responding to the above instructions.

\underline{Scores}:
\begin{enumerate}
    \item Adherence to instructions: 10/10
    \item Common strengths: 6/10
    \item Common weaknesses: 7/10
    \item Common suggestions for improvement: 8/10
    \item Coverage: 10/10
\end{enumerate}
\end{tcolorbox}

\subsection*{}
\begin{tcolorbox}[enhanced jigsaw,unbreakable,pad at break*=0mm,  colback=blue!5!white,colframe=blue!75!black,title=\scriptsize 15, left=0pt, right=1pt, top=0pt, bottom=0pt, floatplacement=t]

  \colorbox{orange}{\makebox[15.5cm]{\textcolor{white}{\scriptsize Prompt Type = \newcommand{\noop}[1]{11} L3 [System / Bulleted Instructions with user preference]}}}
  
\scriptsize messages=[
{"role": "system", "content": "You are a meta-review assistant. "

    "Give me a meta-review by answering following bullets:"
    
    "- Mention core contributions with common contributions first."
    
    "- Mention strengths with common strengths first."
    
    "- Mention weaknesses with common weaknesses first."
    
    "- Mention the common improvements suggested."
    
    "- State the common missing references, if any."},
    
{"role": "user", "content": '1.\textbackslash n$<R1>$\textbackslash n2.\textbackslash n$<R2>$\textbackslash n3.\textbackslash n$<R3>$’},]  

  \colorbox{cyan}{\makebox[15.5cm]{\textcolor{white}{GPT-3.5's Response:}}}
  
Core contributions:

- The authors propose a two-stage generative modeling approach that improves the sample quality of autoregressive models.

- The proposed method smooths the input data distribution using a method previously used for adversarial defense, and then recovers the input distribution by learning to reverse the smoothing process.

- The proposed method is demonstrated on 1D toy problems and extended to more complex datasets such as MNIST, CIFAR-10, and CelebA with applications such as image generation, inpainting, and denoising.

\smallskip
Strengths:

- The paper is well-written and easy to understand with good supporting empirical evaluation.

- The proposed method is an interesting and novel approach to improving the sample quality of autoregressive models.

- The empirical results show that the proposed method outperforms the baseline methods in terms of sample quality.

- The paper includes a strong theoretical grounding and presents two theorems with demonstrations.

\smallskip
Weaknesses:

- The paper lacks analysis of the optimal noise for randomized smoothing, and why only Gaussian noise is used.

- The discussion of the regularization effect that the smoothing process provides could be improved.

- The paper would benefit from a better placement in the literature and a more careful discussion of related works.

- The approach is not tied to autoregressive models, but the experiments focus only on autoregressive models.

\smallskip
Improvements suggested:

- The authors could analyze the optimal noise for randomized smoothing and why only Gaussian noise is used.

- The discussion of the regularization effect could be improved by providing a more detailed explanation of the added term in Theorem 2.

- The paper would benefit from a more careful discussion of related works, and a better placement in the literature.

- The authors could include experiments with other types of models, such as normalizing flows.

- The authors should refer to the proofs for Theorems 1 and 2 in the main text.

- The authors could add the Kernel Inception Distance to the evaluation metrics.

\smallskip
Missing references:

- None mentioned.

  \colorbox{purple}{\makebox[15.5cm]{\textcolor{white}{Brief Analysis and Scores:}}}
  
\underline{Adherence to instructions}: All the instructions have been followed by GPT-3.5.

\underline{Common Strengths}: The strengths in GPT-3.5’s response are correct but the content is inadequate.

\underline{Common Weaknesses}: The weaknesses mentioned by GPT-3.5 are better in response to strength both in terms of quality and quantity.

\underline{Common Suggestions for improvement}: The suggestions for improvement by GPT-3.5 correct and better than responses to strengths and weaknesses.

\underline{Coverage}: It seems that GPT-3.5 has considered all the reviews for responding to the above instructions.

\underline{Scores}:

\begin{enumerate}
    \item Adherence to instructions: 10/10
    \item Common strengths: 6/10
    \item Common weaknesses: 7/10
    \item Common suggestions for improvement: 8/10
    \item Coverage: 10/10
\end{enumerate}
\end{tcolorbox}

\subsection*{}
\begin{tcolorbox}[enhanced jigsaw,unbreakable,pad at break*=0mm,  colback=blue!5!white,colframe=blue!75!black,title=\scriptsize 16, left=0pt, right=1pt, top=0pt, bottom=0pt, floatplacement=t]

  \colorbox{orange}{\makebox[15.5cm]{\textcolor{white}{\scriptsize Prompt Type = \newcommand{\noop}[1]{12} L3 [User / Bulleted Instruction with user preference]}}}
  
\scriptsize messages=[
{"role": "system", "content": "You are a meta-review assistant. "},

{"role": "user", "content": '1.\textbackslash n$<R1>$\textbackslash n2.\textbackslash n$<R2>$\textbackslash n3.\textbackslash n$<R3>$’},

 {"role": "user", "content": "Give me a meta-review by answering following bullets:"
 
    "- Mention core contributions with common contributions first."
    
    "- Mention strengths with common strengths first."
    
    "- Mention weaknesses with common weaknesses first."
    
    "- Mention the common improvements suggested."
    
    "- State the common missing references, if any."},]  

  \colorbox{cyan}{\makebox[15.5cm]{\textcolor{white}{GPT-3.5's Response:}}}
  
Core contributions:

- The paper proposes a two-stage generative modeling approach for improving the sample quality of autoregressive models.

- The authors demonstrate that adding Gaussian noise and posterior modeling of the smoothed data leads to improved sample quality compared to directly fitting the data distribution.

- The paper includes strong theoretical grounding and empirical evaluation on several datasets.

\smallskip
Strengths:

- The paper is well-written and easy to follow.

- The idea of leveraging randomized smoothing for density estimation is interesting and novel.

- The theoretical analysis is sound and provides a strong foundation for the proposed method.

- The empirical evaluation shows significant improvements in sample quality on several datasets.

\smallskip
Weaknesses:

- The optimal noise level for randomized smoothing is not thoroughly analyzed.

- The paper focuses only on Gaussian noise and does not explore other types of noise.

- The discussion of related work could be improved, particularly in terms of the relationship to denoising diffusion and score-matching approaches.

- The paper lacks a discussion on how to assess whether conditionals are better fitted than before.

\smallskip
Common improvements suggested:

- Provide a more in-depth discussion of the regularization effect in Theorem 2 and how it increases the generalization capability of the model.

- Carefully discuss the relationship to denoising diffusion and score-matching approaches in the related work section.

- Include experiments with normalizing flows to demonstrate the method's potential applicability beyond autoregressive models.

- Add an assessment of whether conditionals are better fitted than before.

\smallskip
Common missing references:

- None mentioned.

  \colorbox{purple}{\makebox[15.5cm]{\textcolor{white}{Brief Analysis and Scores:}}}
  
\underline{Adherence to instructions}: All the instructions have been followed by GPT-3.5.

\underline{Common Strengths}: The strengths in GPT-3.5’s response are correct but the content is inadequate.

\underline{Common Weaknesses}: The weaknesses mentioned by GPT-3.5 are better in response to strength both in terms of quality and quantity.

\underline{Common Suggestions for improvement}: The suggestions for improvement by GPT-3.5 correct and better than responses to strengths and weaknesses.

\underline{Coverage}: It seems that GPT-3.5 has considered all the reviews for responding to the above instructions.

\underline{Scores}:
\begin{enumerate}
    \item Adherence to instructions: 10/10
    \item Common strengths: 5/10
    \item Common weaknesses: 7/10
    \item Common suggestions for improvement: 8/10
    \item Coverage: 10/10
\end{enumerate}
\end{tcolorbox}

\subsection*{}
\begin{tcolorbox}[enhanced jigsaw,unbreakable,pad at break*=0mm,  colback=blue!5!white,colframe=blue!75!black,title=\scriptsize 17, left=0pt, right=1pt, top=0pt, bottom=0pt, floatplacement=t]

  \colorbox{orange}{\makebox[15.5cm]{\textcolor{white}{\scriptsize Prompt Type = \newcommand{\noop}[1]{17} L4 [System / Bulleted points with explanation]}}}
  
\scriptsize messages=[
{"role": "system", "content": "As a meta-reviewer," 

"draft a meta review by answering the following bullet points: "

"- What is the summary of core contributions? Provide answer with supporting evidence."

"- Which common strengths are referred to in the reviews? Support your answer with explanation."

"- What common weaknesses are described in the reviews? Give evidence in support of the reply"

"- What suggestions for improvement have been provided by three reviews? Explain the basis for the answer."

"- Which missing references are mentioned in the reviews? Answer with explanation will be desirable."},

{"role": "user", "content": '1.\textbackslash n$<R1>$\textbackslash n2.\textbackslash n$<R2>$\textbackslash n3.\textbackslash n$<R3>$’},]  

  \colorbox{cyan}{\makebox[15.5cm]{\textcolor{white}{GPT-3.5's Response:}}}
  
Summary of core contributions:
The paper proposes a two-stage generative modeling approach that improves the sample quality of autoregressive models. The first stage involves smoothing the input data distribution by adding Gaussian noise and learning a distribution over the noised data. The second stage involves learning the original data distribution conditioned on this noised data. The authors demonstrate the effectiveness of this approach on various datasets such as MNIST, CIFAR-10, CelebA, and synthetic datasets.

Common strengths referred to in the reviews:
All reviewers agree that the paper proposes an interesting and novel idea. The paper is well-written, and the presentation is easy to understand. The authors provide a good motivation for the problem, and the theory is well-founded. The empirical evaluation is comprehensive and convincing, and the results show that the proposed method improves the sample quality of autoregressive models.

Common weaknesses described in the reviews:
The main weakness of the paper is the lack of thorough analysis of the optimal noise for randomized smoothing. The authors only use Gaussian noise, and it is not clear whether this approach is effective for discrete data. The paper could also be improved by discussing the Laplacian term in the objective of Theorem 2 in more depth, and by placing the paper in the literature more accurately. Some reviewers also noted that the experiments on denoising and inpainting are only qualitative and lack quantitative evaluation.

Suggestions for improvement provided by three reviews:
The reviewers suggest the following improvements:

- Provide a more thorough analysis of the optimal noise for randomized smoothing, and consider using other types of noise for discrete data.

- Discuss the Laplacian term in the objective of Theorem 2 in more depth, and provide a more accurate placement of the paper in the literature.

- Include more quantitative evaluation for the experiments on denoising and inpainting.

- Consider adding experiments featuring flows to test whether the same idea could be applied to them.

Missing references mentioned in the reviews:
One reviewer suggests using Kernel Inception Distance (KID) as an additional evaluation metric to supplement FID.

  \colorbox{purple}{\makebox[15.5cm]{\textcolor{white}{Brief Analysis and Scores:}}}
  
\underline{Adherence to instructions}: Almost all the instructions have been followed by GPT-3.5 except the explanation part. The core contributions have not focused only on contributions but also on the process. There is a slight departure from the instructions on this aspect. In all the parts, instruction was to provide explanation. No explicit mention of support, explanation or evidence is given in any of the bullet points.

\underline{Common Strengths}: The strengths in GPT-3.5’s response are correct and the content is quite good.

\underline{Common Weaknesses}: The weaknesses mentioned by GPT-3.5 are better in response to strength in terms of quality.

\underline{Common Suggestions for improvement}: The suggestions for improvement by GPT-3.5 correct and better than response to common strengths.

\underline{Coverage}: It seems that GPT-3.5 has considered all the reviews for responding to the above questions.

\underline{Scores}:
\begin{enumerate}
    \item Adherence to instructions: 8/10
    \item Common strengths: 10/10
    \item Common weaknesses: 10/10
    \item Common suggestions for improvement: 10/10
    \item Coverage: 10/10
\end{enumerate}
\end{tcolorbox}

\subsection*{}
\begin{tcolorbox}[enhanced jigsaw,unbreakable,pad at break*=0mm,  colback=blue!5!white,colframe=blue!75!black,title=\scriptsize 18, left=0pt, right=1pt, top=0pt, bottom=0pt, floatplacement=t]

  \colorbox{orange}{\makebox[15.5cm]{\textcolor{white}{\scriptsize Prompt Type = \newcommand{\noop}[1]{12} L4 [User / Bulleted points with explanation]}}}
  
\scriptsize messages=[
{"role": "system", "content": "You are a meta-review assistant. "},

{"role": "user", "content": '1.\textbackslash n$<R1>$\textbackslash n2.\textbackslash n$<R2>$\textbackslash n3.\textbackslash n$<R3>$’},

 {"role": "user", "content": "Draft a meta review by answering the following bullet points:"
 
"- What is the summary of core contributions? Provide answer with supporting evidence."

"- Which common strengths are referred to in the reviews? Support your answer with explanation."

"- What common weaknesses are described in the reviews? Give evidence in support of the reply"

"- What suggestions for improvement have been provided by three reviews? Explain the basis for the answer."

"- Which missing references are mentioned in the reviews? Answer with explanation will be desirable."},]  

  \colorbox{cyan}{\makebox[15.5cm]{\textcolor{white}{GPT-3.5's Response:}}}
  
Summary of core contributions:
The paper proposes a two-stage generative modeling approach that involves learning a distribution over noised data and then learning the original data distribution conditioned on this noised data. The authors demonstrate that this method leads to improved sample quality compared to fitting the data distribution directly. The paper also provides theoretical support for the approach and validates it through experiments on various datasets.

Common strengths referred to in the reviews:
The paper is well written, and the presentation is clear and easy to understand. The core contributions of the paper are interesting and novel, and the theoretical support provided by the authors is strong. The empirical evaluation of the proposed method is also thorough.

Common weaknesses described in the reviews:
Some weaknesses highlighted by the reviewers include a lack of analysis on the optimal noise for randomized smoothing, why only Gaussian noise is used, and the lack of discussion about the properties of the Laplacian term in the objective function. Additionally, some reviewers have suggested that the experiments on denoising and inpainting suffer from a lack of quantitative evaluation. 

Suggestions for improvement provided by the reviews:
The reviewers suggest that a more in-depth discussion of the properties of the Laplacian term in the objective function and its connection to score matching and denoising methods would be beneficial. Additionally, some reviewers suggest that the paper could be improved by analyzing the optimal noise for randomized smoothing and experimenting with other types of noise. The lack of quantitative evaluation in the experiments on denoising and inpainting is also pointed out, and the reviewers suggest that this could be addressed in future work.

Missing references mentioned in the reviews:
One reviewer suggests that references to recent work on denoising diffusion and denoising score-matching approaches should be included. Additionally, the Kernel Inception Distance is suggested as a metric that could be used to further validate the improved sample quality claimed by the authors.

  \colorbox{purple}{\makebox[15.5cm]{\textcolor{white}{Brief Analysis and Scores:}}}
  
\underline{Adherence to instructions}: Almost all the instructions have been followed by GPT-3.5. The only part that has not been adhered to is the explanation that has not been mentioned explicitly in any of the answers.

\underline{Common Strengths}: The strengths in GPT-3.5’s response are correct but the content is not as good as that in case of questions provided as system content.

\underline{Common Weaknesses}: The weaknesses mentioned by GPT-3.5 are not very good. This has also taken the point from suggestions for improvement.

\underline{Common Suggestions for improvement}: The suggestions for improvement by GPT-3.5 are correct and better than response to strengths. These are not as good as those in case the instruction/question is placed as system content.

\underline{Coverage}: It seems that GPT-3.5 has considered all the reviews for responding to the above questions.

\underline{Scores}:
\begin{enumerate}
    \item Adherence to instructions: 8/10
    \item Common strengths: 9/10
    \item Common weaknesses: 6/10
    \item Common suggestions for improvement: 9/10
    \item Coverage: 10/10
\end{enumerate}
\end{tcolorbox}

\subsection*{}
\begin{tcolorbox}[enhanced jigsaw,unbreakable,pad at break*=0mm,  colback=blue!5!white,colframe=blue!75!black,title=\scriptsize 19, left=0pt, right=1pt, top=0pt, bottom=0pt, floatplacement=t]

  \colorbox{orange}{\makebox[15.5cm]{\textcolor{white}{\scriptsize Prompt Type = \newcommand{\noop}[1]{19} L4 [System / Bulleted points with explanation]}}}
  
\scriptsize messages=[
{"role": "system", "content": "Considering yourself to be a meta-reviewer, "

"draft a meta review by answering the following points: "

"(a) What would you consider to be the most balanced opinion about core contributions and why?"

"(b) Which strengths should be considered at your level and why?"

"(c) What are the weaknesses that deserve attention at your level and why?"

"(d) What suggestions for improvement are worth considering? Provide explanation."

"(e) Which missing references do you consider appropriate for meta review? Support with explanation."
},

{"role": "user", "content": '1.\textbackslash n$<R1>$\textbackslash n2.\textbackslash n$<R2>$\textbackslash n3.\textbackslash n$<R3>$’},]  

  \colorbox{cyan}{\makebox[15.5cm]{\textcolor{white}{GPT-3.5's Response:}}}
  
(a) The most balanced opinion about the core contributions of the paper is that it presents a simple and effective approach to improving the sample quality of autoregressive models by smoothing the input data distribution and then recovering the original distribution. The paper is well-motivated, has strong theoretical grounding and provides good empirical validation of the proposed method. However, there are some weaknesses that require attention, such as a lack of analysis on the optimal noise for randomized smoothing and why Gaussian noise is chosen compared to other types of noise.

(b) The strengths of the paper at the meta level include a clear and well-motivated presentation, good empirical validation, and a strong theoretical foundation. The paper also provides a novel approach to improving the sample quality of autoregressive models, which has potential applications in various fields.

(c) The weaknesses that deserve attention at the meta level include a lack of analysis on the optimal noise for randomized smoothing, a lack of discussion on why Gaussian noise is chosen over other types of noise, and a need for a more in-depth placement in the literature. Additionally, the approach isn't really tied to autoregressive models, so it would have been useful to see experiments featuring other models, such as normalizing flows.

(d) One suggestion for improvement is to provide more analysis on the optimal noise for randomized smoothing and why Gaussian noise is chosen over other types of noise. Another suggestion is to more carefully discuss the connections between the added Laplacian term in Theorem 2 and the score matching objective presented in Hyvarinen (2005) and other related works. Additionally, it would be helpful to include experiments featuring other models, such as normalizing flows, to further demonstrate the effectiveness of the proposed method.

(e) One missing reference that could be appropriate for the meta-review is "Denoising Diffusion Probabilistic Models" by Song and Ermon, which presents a related approach for improving the quality of samples from generative models using denoising methods. The paper also discusses connections between denoising methods and score matching objectives, which could be helpful for the discussion of Theorem 2 in the paper.

  \colorbox{purple}{\makebox[15.5cm]{\textcolor{white}{Brief Analysis and Scores:}}}
  
\underline{Adherence to instructions}: Almost all the instructions have been followed by GPT-3.5. The only part that has not been adhered to is the explanation that has not been mentioned explicitly in any of the answers.

\underline{Common Strengths}: The strengths in GPT-3.5’s response are correct and the quality is good.

\underline{Common Weaknesses}: The weaknesses mentioned by GPT-3.5 are better in response to strength.

\underline{Common Suggestions for improvement}: The suggestions for improvement by GPT-3.5 are better than response to strengths.

\underline{Coverage}: It seems that GPT-3.5 has considered all the reviews for responding to the above questions.

\underline{Scores}:
\begin{enumerate}
    \item Adherence to instructions: 8/10
    \item Common strengths: 8/10
    \item Common weaknesses: 9/10
    \item Common suggestions for improvement: 9/10
    \item Coverage: 10/10
\end{enumerate}
\end{tcolorbox}

\subsection*{}
\begin{tcolorbox}[enhanced jigsaw,unbreakable,pad at break*=0mm,  colback=blue!5!white,colframe=blue!75!black,title=\scriptsize 20, left=0pt, right=1pt, top=0pt, bottom=0pt, floatplacement=t]

  \colorbox{orange}{\makebox[15.5cm]{\textcolor{white}{\scriptsize Prompt Type = \newcommand{\noop}[1]{20} L4 [User / Bullet ed points with explanation]}}}
  
\scriptsize messages=[
{"role": "system", "content": "You are a meta-review assistant. "},

{"role": "user", "content": '1.\textbackslash n$<R1>$\textbackslash n2.\textbackslash n$<R2>$\textbackslash n3.\textbackslash n$<R3>$’},

 {"role": "user", "content": "Draft a meta review by answering the following points:"
 
"(a) What would you consider to be the most balanced opinion about core contributions and why?"

"(b) Which strengths should be considered at your level and why?"

"(c) What are the weaknesses that deserve attention at your level and why?"

“(d) What suggestions for improvement are worth considering? Provide explanation."

"(e) Which missing references do you consider appropriate for meta review? Support with explanation."},]  

  \colorbox{cyan}{\makebox[15.5cm]{\textcolor{white}{GPT-3.5's Response:}}}
  
(a) The most balanced opinion about the core contributions of the paper would be that the proposed two-stage generative modeling approach, which includes a denoising step and learning the original data distribution conditioned on the noised data, leads to improved sample quality compared to fitting the data distribution directly. This approach is motivated by the problem of mismatched conditionals in autoregressive models and the manifold hypothesis. The paper provides strong theoretical grounding and empirical evaluation to support the proposed method.

(b) At my level, the strengths of the paper that should be considered include the clear writing and presentation, the well-motivated problem, the novel and straightforward approach, and the strong theoretical grounding of the proposed method.

(c) The weaknesses that deserve attention at my level include a lack of analysis on the optimal amount of noise for randomized smoothing, the limited discussion of the Laplacian term in Theorem 2, and a lack of inclusion of experiments featuring normalizing flows. Additionally, there is some confusion around the terminology used to describe the proposed method, and a more thorough discussion of related work could be beneficial.

(d) Suggestions for improvement that are worth considering include providing more analysis on the optimal amount of noise for randomized smoothing, discussing the Laplacian term in Theorem 2 in more depth, and including experiments featuring normalizing flows. Additionally, clarifying the terminology used to describe the proposed method and providing a more thorough discussion of related work could also be beneficial.

(e) One missing reference that could be appropriate for meta review is "Denoising Diffusion Probabilistic Models" by Song et al. (2020), which proposes a denoising diffusion probability model and demonstrates its effectiveness in generative modeling tasks. This reference could be relevant to the discussion of denoising methods and their applicability to generative modeling.

  \colorbox{purple}{\makebox[15.5cm]{\textcolor{white}{Brief Analysis and Scores:}}}
  
\underline{Adherence to instructions}: Almost all the instructions have been followed by GPT-3.5. The only part that has not been adhered to is the explanation that has not been mentioned explicitly in any of the answers.

\underline{Common Strengths}: The strengths in GPT-3.5’s response are correct and the content is quite good.

\underline{Common Weaknesses}: The weaknesses mentioned by GPT-3.5 are not as good as in the case of the questions placed as system content. Moreover, there is an overlap between common weaknesses and common suggestions for improvement.

\underline{Common Suggestions for improvement}: The suggestions for improvement by GPT-3.5 are correct and quite good in quality.

\underline{Coverage}: It seems that GPT-3.5 has considered all the reviews for responding to the above questions.

\underline{Scores}:
\begin{enumerate}
    \item Adherence to instructions: 8/10
    \item Common strengths: 10/10
    \item Common weaknesses: 8/10
    \item Common suggestions for improvement: 10/10
    \item Coverage: 10/10
\end{enumerate}
\end{tcolorbox}

\end{document}